\documentclass[oneside,11pt]{article}

\usepackage{times,url,mathrsfs}
\usepackage{amsmath,wrapfig,color,bigfoot}
\usepackage{amsfonts}
\usepackage{graphicx}
\usepackage{subfigure}

\begin{document}

\title{\Huge Sign Stable Random Projections \\ for Large-Scale Learning }
\author{ \textbf{\Large Ping Li} \vspace{0.05in}\\
         Department of Statistics and Biostatistics\\
         Department of Computer Science\\
       Rutgers University\\
          Piscataway, NJ 08854, USA\\
       \texttt{pingli@stat.rutgers.edu}
}

\date{}
\maketitle

\begin{abstract}

\noindent In this paper, we study the use of ``sign $\alpha$-stable random projections'' (where $0<\alpha\leq 2$) for building basic data processing tools in the context of large-scale machine learning applications (e.g., classification, regression,  clustering, and near-neighbor search). After the processing by sign stable random projections, the inner products of the processed data approximate various types of nonlinear kernels depending on the value of $\alpha$. Thus, this approach provides an effective strategy for approximating nonlinear learning algorithms essentially at the cost of linear learning. When $\alpha =2$,  it is known that the corresponding nonlinear kernel is the arc-cosine kernel. When $\alpha=1$, the procedure approximates the arc-cos-$\chi^2$ kernel (under certain condition). When $\alpha\rightarrow0+$, it corresponds to the resemblance kernel, which provides the  exciting connection between two popular randomized algorithms: (i) stable random projections (ii) $b$-bit minwise hashing.  No theoretical results are known so far for other $\alpha$ values except for $\alpha=2$, 1, or $0+$.\\

\noindent From practitioners' perspective, the method of sign $\alpha$-stable random projections is  ready  to be tested for large-scale learning applications, where $\alpha$ can be simply viewed as a tuning parameter. What is missing in the literature is an extensive empirical study to show the effectiveness of sign stable random projections, especially for $\alpha\neq 2$ or 1. The paper supplies such a study on a wide variety of classification datasets. In particular, we compare shoulder-by-shoulder sign stable random projections with the recently proposed ``0-bit consistent weighted sampling (CWS)''~\cite{Report:Li_CWS15} (which is only for nonnegative data). We provide the detailed comparisons  on all the 34 datasets used by~\cite{Report:Li_CWS15}. In addition, we present the comparison on a larger dataset with 350,000 examples. For all datasets, we experiment with $\alpha \in \{0.1, 0.25, 0.5, 0.75, 1, 1.25, 1.5, 1.75,2\}$. For most datasets, sign stable random projections can approach (or in some cases even slightly exceed) the performance of 0-bit CWS, given enough projections. Typically, to reach the same accuracy, sign stable random projections would require significantly more projections than the number of samples needed by 0-bit CWS. There are also datasets for which sign stable random projections could not achieve the same accuracy as 0-bit CWS regardless of $\alpha$. \\

\noindent While the comparison results seem to favor 0-bit consistent weighted sampling (which is only for nonnegative data), the distinct advantage of sign stable random projections is that the method is applicable to general data types, not only for nonnegative data. It is also an interesting  research problem to combine 0-bit CWS with sign stable random projections, for example, a strategy similar to ``CoRE kernels''~\cite{Proc:Li_UAI14}.

\end{abstract}

\newpage

\section{Introduction}\label{sec:intro}

In this paper, we focus on the idea of ``sign $\alpha$-stable random projections'' and the applications in  machine learning with massive (and possibly streaming~\cite{Article:Muthukrishnan_05}) data. Consider two data vectors $u,\ v\in\mathbb{R}^D$ from a data matrix, the central idea is to multiply them with a random projection matrix $\{s_{ij}\}, i = 1, ..., D$, $j=1, ..., k$, whose entries, $s_{ij}$, are sampled  i.i.d. from an $\alpha$-stable distribution, denoted by $S(\alpha,1)$.  That is,
\begin{align}
x_j = \sum_{i=1}^D u_i s_{ij},\hspace{0.2in} y_j = \sum_{i=1}^D v_i s_{ij},\hspace{0.2in} s_{ij}\sim S(\alpha,1),\ \ i.i.d. \hspace{0.2in} j = 1, 2, ..., k
\end{align}

The use of $\alpha$-stable distributions was studied in the context of estimating frequency moments of data streams~\cite{Article:Indyk_JACM06,Proc:Li_SODA08} and in the recent work on  ``one scan 1-bit compressed sensing''~\cite{Report:Li_1bitCS15}. Here, we adopt the parameterization~\cite{Book:Zolotarev_86,Book:Samorodnitsky_94} such that, if $s \sim S(\alpha,d)$, then the characteristic function is $E\left(e^{\sqrt{-1}st}\right) = e^{-d|t|^\alpha} $. When $\alpha=2$, $S(2,d)$ is equivalent to a Gaussian distribution $N(0,\sigma^2=2d)$.  When $\alpha=1$, $S(1,1)$ is the standard Cauchy distribution. Although in general no closed-form density functions of $\alpha$-stable distributions are available, one can  easily sample from an $\alpha$-stable distribution  by (e.g.,) the classical CMS~\cite{Article:Chambers_JASA76} method.

Stable distributions with $\alpha<2$ are also known to be ``heavy-tailed'' distributions because if $s \sim S(\alpha,1)$,  then unless $\alpha=2$,  we always have $E(|s|^\lambda)=\infty$ if $\lambda\geq \alpha$. This is probably the reason why stable distributions were rarely used in machine learning and data mining applications.

\subsection{Sign Stable Random Projections}

By property of stable distributions, we have $x_j \sim S\left(\alpha,\sum_{i=1}^D |u_i|^\alpha\right)$ and $y_j \sim S\left(\alpha,\sum_{i=1}^D |v_i|^\alpha\right)$, $j = 1, 2, ..., k$. Unless $\alpha =2$, it might be difficult to imagine how one can make use of these (manually generated) heavy-tailed data for of machine learning applications. Indeed, we do not directly use the projected data. Instead, in this paper, we only utilize the projected data through their signs, i.e., $sign(x_j)$ and $sign(y_j)$, which are well-behaved and can be used for building tools for large-scale machine learning.

If $x_j\leq 0$, we can code $x_j$ as a two-dimensional vector $[0\ \ 1]$. If $x_j>0$, then we code it as $[1\ \ 0]$. Then we concatenate $k$ such two-dimensional vectors to form a vector of length $2k$ (with $k$ 1's). We apply the same coding scheme to $y_j$ (and all the projected data).  The signs, $sign(x_j)$ and $sign(y_j)$, are statistically dependent and it is interesting (and in general challenging) to find out how the signs are related. \\

When $\alpha=2$, the relationship between $sign(x_j)$ and $sign(y_j)$ is well-known~\cite{Article:Goemans_JACM95,Proc:Charikar,Proc:Li_ICML14}
\begin{align}
\alpha = 2:\hspace{0.2in} \mathbf{Pr}\left(sign(x_j) = sign(y_j)\right) = 1-\frac{1}{\pi}\cos^{-1} \rho_2,\hspace{0.2in} \rho_2= \frac{\sum_{i=1}^D u_i v_i}{\sqrt{\sum_{i=1}^D |u_i|^2}\sqrt{\sum_{i=1}^D |v_i|^2}}
\end{align}
Thus, the ``collision probability'' is monotonic in $\rho_2$,  which is  the correlation coefficient.  Although $\cos^{-1}\rho_2$ is nonlinear, the estimator of the probability, i.e.,  $\frac{1}{k}\sum_{j=1}^k 1\{x_j = y_j\}$ can  be viewed as an inner product once we expand a sign as either $[0\ \ 1]$ or $[1\ \ 0]$.  In other words, we only need to pay the cost of linear learning to approximately train a classifier originally based on nonlinear kernels.

It is not so straightforward to calculate  the collision probability once $\alpha<2$. A recent work~\cite{Proc:Li_NIPS13} focused on $\alpha=1$ and showed that, when $u_i\geq 0, v_i\geq 0, \sum_{i=1}^D u_i=\sum_{i=1}^D v_i=1$, we have
\begin{align}
\alpha = 1:\hspace{0.2in} \mathbf{Pr}\left(sign(x_j) = sign(y_j)\right) \approx 1-\frac{1}{\pi}\cos^{-1} \rho_{\chi^2},\hspace{0.2in} \rho_{\chi^2}= \sum_{i=1}^D \frac{2u_i v_i}{u_i+v_i}
\end{align}
Note that the so-called $\chi^2$-kernel, $\rho_{\chi^2}$, is popular in computer vision, for data generated from histograms.

When $\alpha\rightarrow0+$, \cite{Proc:Li_NIPS13} mentioned in the ``future work'' that the collision probability is related to the ``resemblance''  when the data are nonnegative:
\begin{align}
\alpha = 0+:\hspace{0.2in} \mathbf{Pr}\left(sign(x_j) = sign(y_j)\right) = \frac{1}{2}+\frac{1}{2} R,\hspace{0.2in} R =\frac{  \sum_{i=1}^D 1\{u_i>0 \text{ and } v_i >0\} }{ \sum_{i=1}^D 1\{u_i>0 \text{ or } v_i >0\} }
\end{align}
Interestingly, this collision probability is essentially the same as the collision probability of ``1-bit minwise hashing''~\cite{Article:Li_Konig_CACM11}.\\

For other $\alpha$ values, at this moment we can not  relate the collision probabilities to any known similarity measures. On the other hand, the estimator $\frac{1}{k}\sum_{j=1}^k 1\{x_j = y_j\}$ (which is an inner product) is of course still a valid positive definite kernel for any $\alpha$. Thus, we can anyway use sign $\alpha$-stable random projections for building large-scale learning algorithms, where $\alpha$ can be viewed as an important tuning parameter. What is missing in the literature is an extensive empirical study and our paper supplies such a study.

\subsection{Resemblance, Min-Max Kernel, and 0-Bit Consistent Weighted Sampling (CWS)}

As mentioned above, the collision probability of sign stable random projections at $\alpha=0+$ is related to the resemblance  $R$ when the data (e.g., $u$ and $v$) are nonnegative. From the definition
\begin{align}
 R =R(u,v) = \frac{  \sum_{i=1}^D 1\{u_i>0 \text{ and } v_i >0\} }{ \sum_{i=1}^D 1\{u_i>0 \text{ or } v_i >0\} }, \hspace{0.2in} u_i\geq 0,\ v_i\geq 0
\end{align}
we can see that $R$ only makes sense when the data are sparse (i.e., most entries are zero). When the data are fully dense, we have $R=1$ always.  This may seriously limit the use of resemblance when the data are not sparse. This issue can be largely fixed by the introduction of the min-max kernel which is defined as
\begin{align}\label{eqn_MM}
K_{MM}(u,v) = \frac{\sum_{i=1}^D \min\{u_i,\ v_i\}}{\sum_{i=1}^D \max\{u_i,\ v_i\}}, \hspace{0.2in} u_i\geq 0,\ v_i\geq 0
\end{align}
The recent work~\cite{Report:Li_CWS15} also provides a variant,   called  the ``normalized min-max kernel'':
\begin{align}\label{eqn_NMM}
K_{NMM}(u,v) = \frac{\sum_{i=1}^D \min\{u_i,\ v_i\}}{\sum_{i=1}^D \max\{u_i,\ v_i\}},\hspace{0.2in} \sum_{i=1}^D u_i = 1,\ \ \ \sum_{i=1}^D v_i = 1
\end{align}

The resemblance is a popular measure of similarity for binary data and can be sampled efficiently by minwise hashing~\cite{Proc:Broder_WWW97,Article:Li_Konig_CACM11}.  The min-max kernels can also be
sampled using the technique called consistent weighted sampling (CWS)~\cite{Report:Manasse_CWS10,Proc:Ioffe_ICDM10}. Traditionally, each sample of CWS consists of two values, one of which is unbounded. The so-called ''0-bit'' CWS~\cite{Report:Li_CWS15} simply discarded the unbounded value to make CWS much more convenient for  large-scale  machine learning tasks.

Because~\cite{Report:Li_CWS15} experimented with a large collection of  datasets, we hope to compare, shoulder-by-shoulder, sign stable random projections with 0-bit CWS, although we should reiterate that 0-bit CWS is only designed for nonnegative data and is hence not as general as sign stable random projections.

\section{Experiments}

\subsection{Datasets and Summary of Results}

We have experimented all the 34 datasets used in the recent paper for ''0-bit CWS''~\cite{Report:Li_CWS15} to provide a shoulder-by-shoulder comparison. The results are summarized in Table~\ref{tab_data}. The results show that, given enough projections, sign $\alpha$-stable random projections can often achieve good accuracies (and better than linear). The value of $\alpha$ is an important parameter which needs to be individually tuned for each dataset.

\begin{table*}[t]
\caption{Datasets and classification accuracies (in \%). We use all the datasets in the recent work on ``0-bit'' CWS~\cite{Report:Li_CWS15}. We report the results of linear kernels, min-max kernels (\ref{eqn_MM}), normalize min-max kernels (\ref{eqn_NMM}) and sign $\alpha$-stable random projections with $\alpha\in\{0.1, 0.25, 0.5, 0.75, 1, 1.25, 1.5, 1.75,2\}$ and $k=8192$. The values for the linear kernel, min-max kernels, and n-min-max (or n-m-m) kernels are directly quoted from~\cite{Report:Li_CWS15}.  For the min-max  (and n-m-m) kernels,  the accuracies were computed on the original data using LIBSVM ``pre-computed'' kernel functionality and $l_2$-regularized kernel SVM (which has a tuning parameter $C$). The reported test classification accuracies  are the best accuracies from a wide range of $C$ values. The reported accuracies of sign $\alpha$-stable random projections (i.e., the last 9 columns) and linear kernels  $l_2$-regularized linear SVM  were computed by LIBLINEAR~\cite{Article:Fan_JMLR08}. We {highlight}  (in {\bf bold}) the highest accuracies among all methods as well as the highest accuracies of sign $\alpha$-stable random projections among  9 $\alpha$ values.
}
\begin{center}{\scriptsize
{\begin{tabular}{l r r c c c c c c c c c  c  c c}
\hline \hline
Dataset     &\# train  &\# test    &linear  &min-max  &n-m-m&0.1 &0.25& 0.5 &0.75 &1 &1.25 &1.5  &1.75 &2 \\
\hline
Covertype10k   &10,000 &50,000 &70.9 &\textbf{80.4}  &80.2 &74.5  &76.7 &77.9 & 78.3 &78.4 &\textbf{78.5} &78.4&78.3 &78.2\\
Covertype20k   &20,000 &50,000 &71.1  &\textbf{83.3}  &83.1 &76.5  &78.4  &79.8   &80.3   &80.4   &80.4   &\textbf{80.7}   &80.5   &80.3\\
IJCNN5k &5,000 &91,701 &   91.6  &94.4 &95.3 &91.0  &92.8  &93.7   &94.5   &95.2  &94.7 &\textbf{95.4}   &95.3   &\textbf{95.4}\\
IJCNN10k &10,000 &91,701 & 91.6  &95.7 &\textbf{96.0} &91.2   &93.3   &94.2   &95.4 &95.7  &95.9  &95.7  &95.9   &\textbf{96.0}\\
Isolet &6,238 &1,559 &95.4   &96.4 &\textbf{96.6} &90.9   &93.7  &94.9  &95.3   &95.7  &95.6   &\textbf{95.8}   &\textbf{95.8}   &95.6\\
Letter &16,000 &4,000 &62.4   &\textbf{96.2}  &95.0   &88.0 &92.2 &94.1 &94.8 &95.3 &95.3 &95.4 &\textbf{95.6}&\textbf{95.6}\\
Letter4k &4,000 &16,000 &61.2 &91.4 &90.2 & 84.9   &88.1   &90.1   &91.1   &91.5   &91.9   &\textbf{92.1}   &92.0   &91.7\\
M-Basic   &12,000 &50,000 &90.0 &\textbf{96.2} &96.0   &95.9   &\textbf{96.0}   &\textbf{96.0}   &95.9   &95.7  &95.5   &95.4   &95.2   &95.0\\
M-Image &12,000 &50,000 &70.7  &\textbf{80.8} &77.0 &55.6   &64.1   &67.9   &69.9   &70.9   &71.4   &71.9   &\textbf{72.1}   &72.0\\
MNIST10k &10,000 &60,000 &90.0  &\textbf{95.7} &95.4  &95.6   &\textbf{95.7}  &95.6  &95.5  &95.3   &95.2 &95.0  &94.8   &94.7\\
M-Noise1 &10,000 &4,000 &60.3  &\textbf{71.4} &68.5 & 47.0   &53.2   &56.8  &58.2   &58.9   &59.7   &60.4  &60.4   &\textbf{60.9}\\
M-Noise2 &10,000 &4,000 &62.1  &\textbf{72.4} &70.7 &46.4   &54.6   &57.5   &59.4   &60.6   &61.5   &61.9  &61.5   &\textbf{61.7}\\
M-Noise3 &10,000 &4,000 &65.2  &\textbf{73.6} &71.9 &50.1   &57.1   &60.6   &62.3   &63.1   &64.0   &64.4  &64.7   &\textbf{64.8}\\
M-Noise4 &10,000 &4,000 & 68.4  &\textbf{76.1}  &75.2 & 53.0 &59.2  &62.9   &65.2  &66.0  &66.7  &67.2   &67.5  &\textbf{67.8}\\
M-Noise5 &10,000 &4,000 &72.3   &\textbf{79.0}   &78.4   &55.4  &62.4 &66.4  &68.6  &68.9 &70.2  &70.4 &70.7 &\textbf{71.5}\\
M-Noise6 &10,000 &4,000 &78.7   &84.2   &\textbf{84.3}  &59.9  &68.4 &72.6 &74.2 &75.5 &76.1 &76.5 &76.6 &\textbf{77.3}\\
M-Rand &12,000 &50,000 &78.9   &{\bf84.2}   &84.1   &60.2 &69.1  &72.5 &74.2 &75.2 &76.1 &76.5  &76.8 &{\bf77.1}\\
M-Rotate &12,000 &50,000    &48.0   &{\bf84.8}   &83.9   &82.6  &{\bf83.0} &82.5  &81.6 &80.9 &80.2 &79.5 &78.8 &78.2 \\
M-RotImg &12,000 &50,000 &31.4   &{\bf41.0}   &38.5   & 24.1 &26.8  &29.3 &30.6 &32.0 &32.7 &33.4 &33.7 &{\bf34.1}\\
Optdigits &3,823 &1,797 &95.3   &{97.7}  &97.4  &95.7 &96.4 &96.7 &97.3 &97.4 &97.5 &{\bf97.8} &{\bf97.8} &97.7\\
Pendigits &7,494 &3,498 &87.6   &97.9   &{98.0}  & 96.6  &97.0 &97.5   &97.7  &97.9  &97.9 &98.0  &{\bf98.1}  &{\bf98.1}\\
Phoneme &3,340 &1,169 &91.4   &{\bf92.5}   &92.0   & 88.0 &90.4 &91.3 &91.5 &91.7 &91.6 &91.5 &{\bf91.9} &91.6\\
Protein &17,766 &6,621 & 69.1   &{\bf72.4}  &70.7   & 69.0 &69.9 &70.6 &{\bf70.7} &70.5 &70.3 &69.7 &69.4 & 68.8\\
RCV1 &20,242 &60,000 &96.3   &{\bf96.9}   &{\bf96.9}   &94.8 &{\bf94.9} &{\bf94.9} &{\bf94.9}&{\bf94.9} &94.8 &94.7&94.6 &94.4\\
Satimage &4,435 &2,000 & 78.5   &{\bf90.5}   &87.8   &84.3 &86.1 & 87.1 &87.1 &87.3 &87.7 &{\bf88.0} &87.8 &87.7\\
Segment &1,155 &1,155 &92.6   &{\bf98.1}  &97.5  &96.1&97.0 &{\bf97.4} &97.2 & 97.3 &97.2 &97.2 &96.9&96.9\\
SensIT20k &20,000 &19,705 &80.5   &86.9   &{\bf87.0}   &85.5 &86.2 &86.6 &{\bf86.7} &{\bf86.7} &86.3 &86.0 &85.3 &84.7\\
Shuttle1k &1,000 &14,500 &90.9   &{\bf99.7}   &99.6   &99.2 &99.2 &99.4 &{\bf99.6} &99.5 &{\bf99.6} &99.5 &{\bf99.6} &{\bf99.6}\\
Spam  &3,065 &1,536 & 92.6   &{\bf95.0} & 94.7   & {\bf95.0} &{\bf95.0} &94.9 &94.7 &94.7 &94.4 &94.4 &94.2 &94.0\\
Splice &1,000 &2,175 &85.1   &{\bf95.2}   &94.9  &87.4 &90.7 &{\bf91.7} &91.6 & 91.0 &90.7 &89.6 &88.9 &87.3\\
USPS   &7,291 &2,007     &91.7  &{95.3}   &{95.3} & 94.6 &95.3 &{\bf95.5} &95.4 &95.3 &95.3 &95.1 &95.1 &95.1\\
Vowel &528 &462 &40.9   &{\bf59.1}  &53.5   &41.2 &41.3 &43.8 &46.1 &47.2 &49.3 &51.2 &52.7 &{\bf52.9}\\
WebspamN1-20k  &20,000 &60,000 &93.0  &{\bf97.9}   &97.8   & 96.9 &97.3 &{\bf97.5} &{\bf97.5} &97.5 &97.4 &97.3 &97.2  &97.0\\
YoutubeVision &11,736 &10,000 &63.3 &{\bf72.4}   &{\bf72.4}   &59.7 &65.0 &68.4 &{\bf69.4} &69.2 &68.9 &67.9 &66.2 &64.8\\
\hline\hline
\end{tabular}}
}
\end{center}\label{tab_data}

\end{table*}

\newpage\clearpage

\subsection{Detailed Results of Sign $\alpha$-Stable Random Projections}

\begin{figure}[h!]
\begin{center}

\mbox{
\includegraphics[width=2.3in]{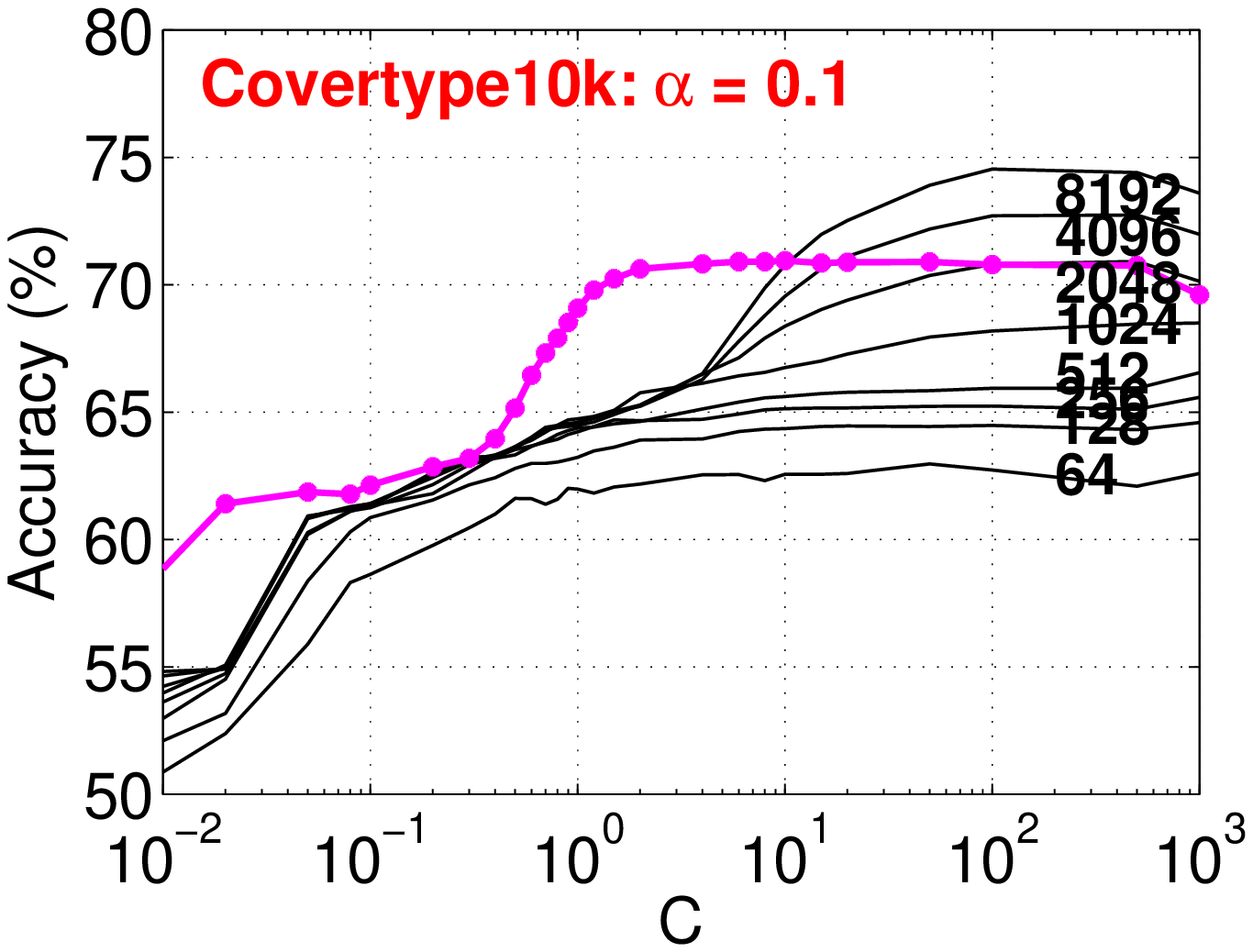}\hspace{-0.15in}
\includegraphics[width=2.3in]{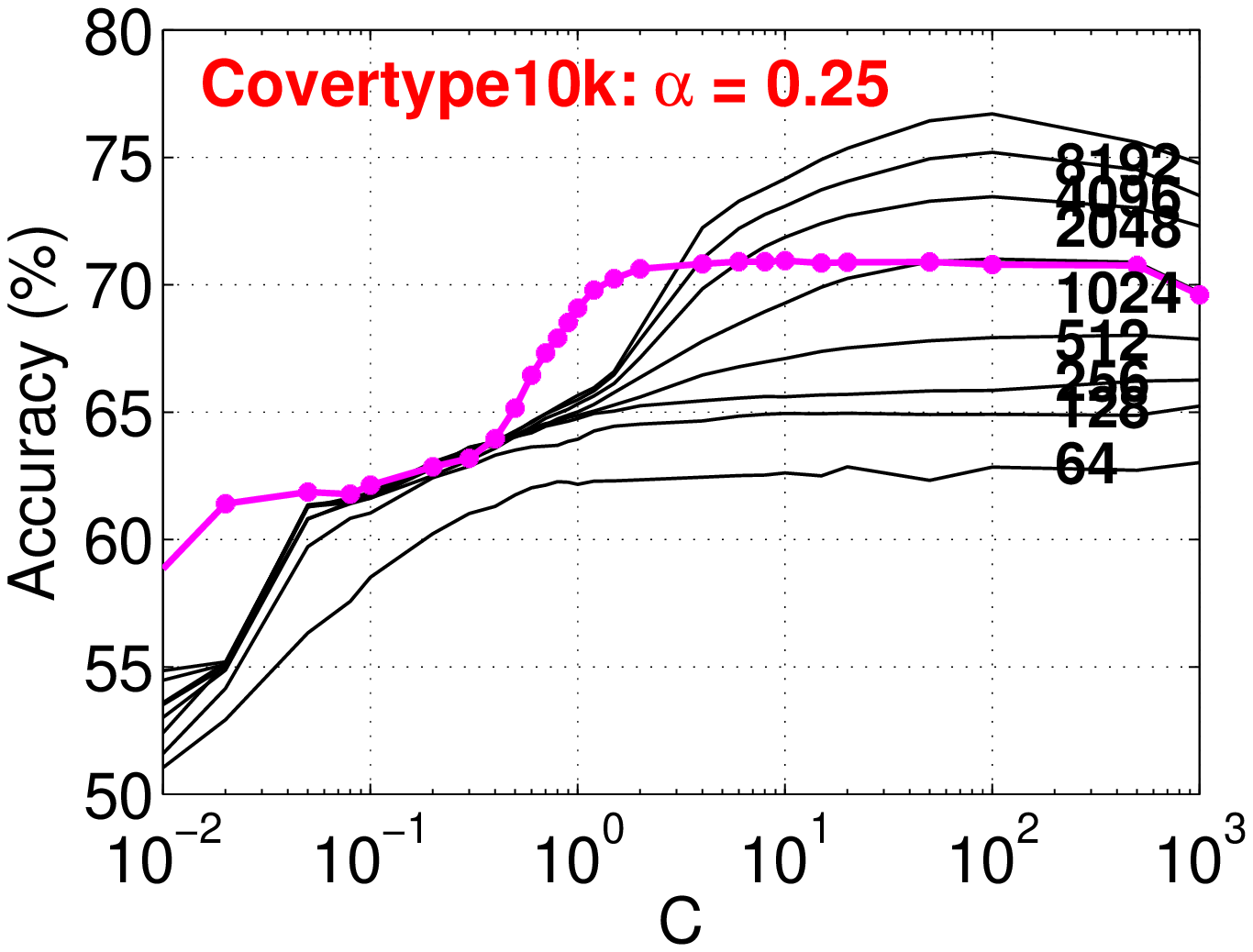}\hspace{-0.15in}
\includegraphics[width=2.3in]{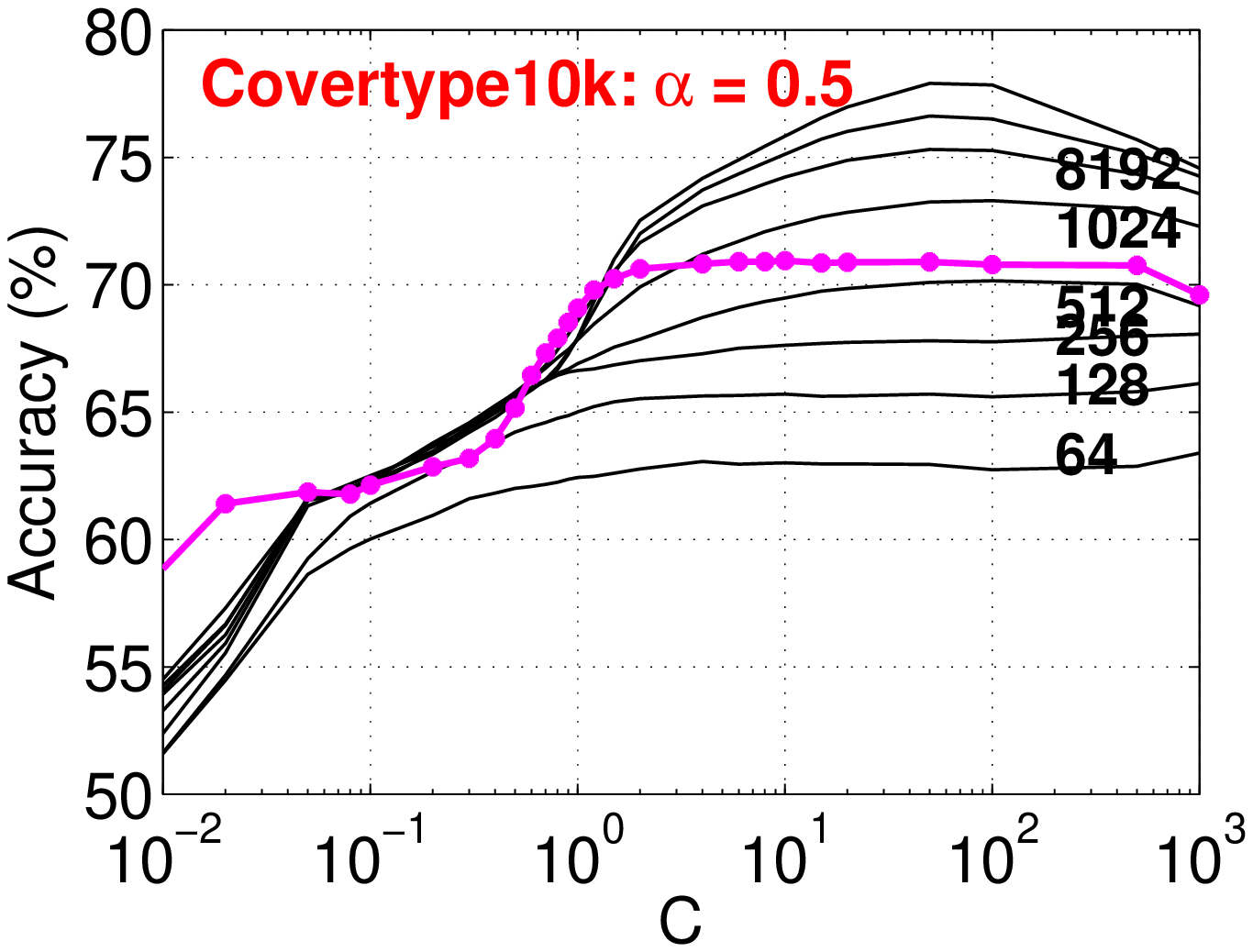}
}
\mbox{
\includegraphics[width=2.3in]{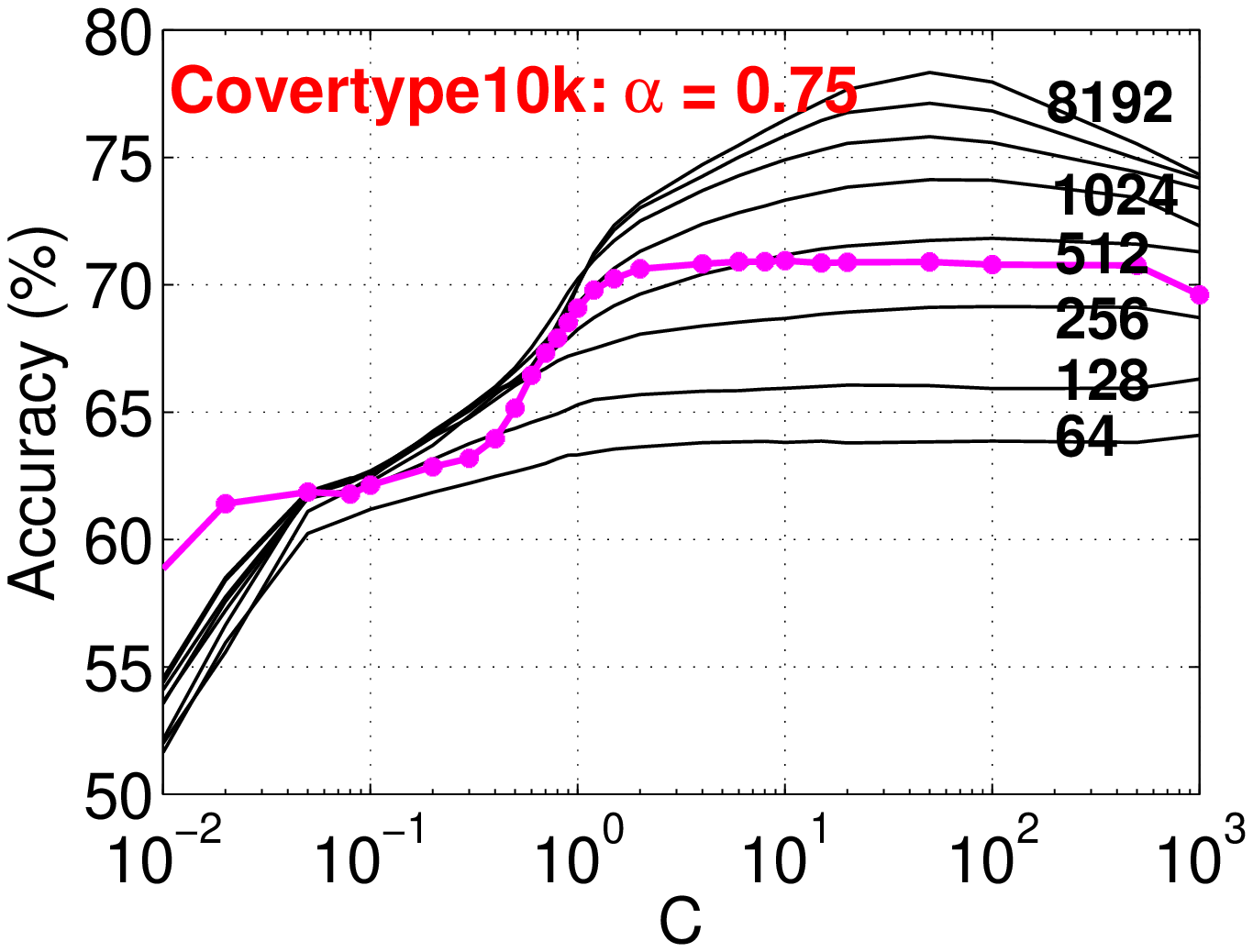}\hspace{-0.15in}
\includegraphics[width=2.3in]{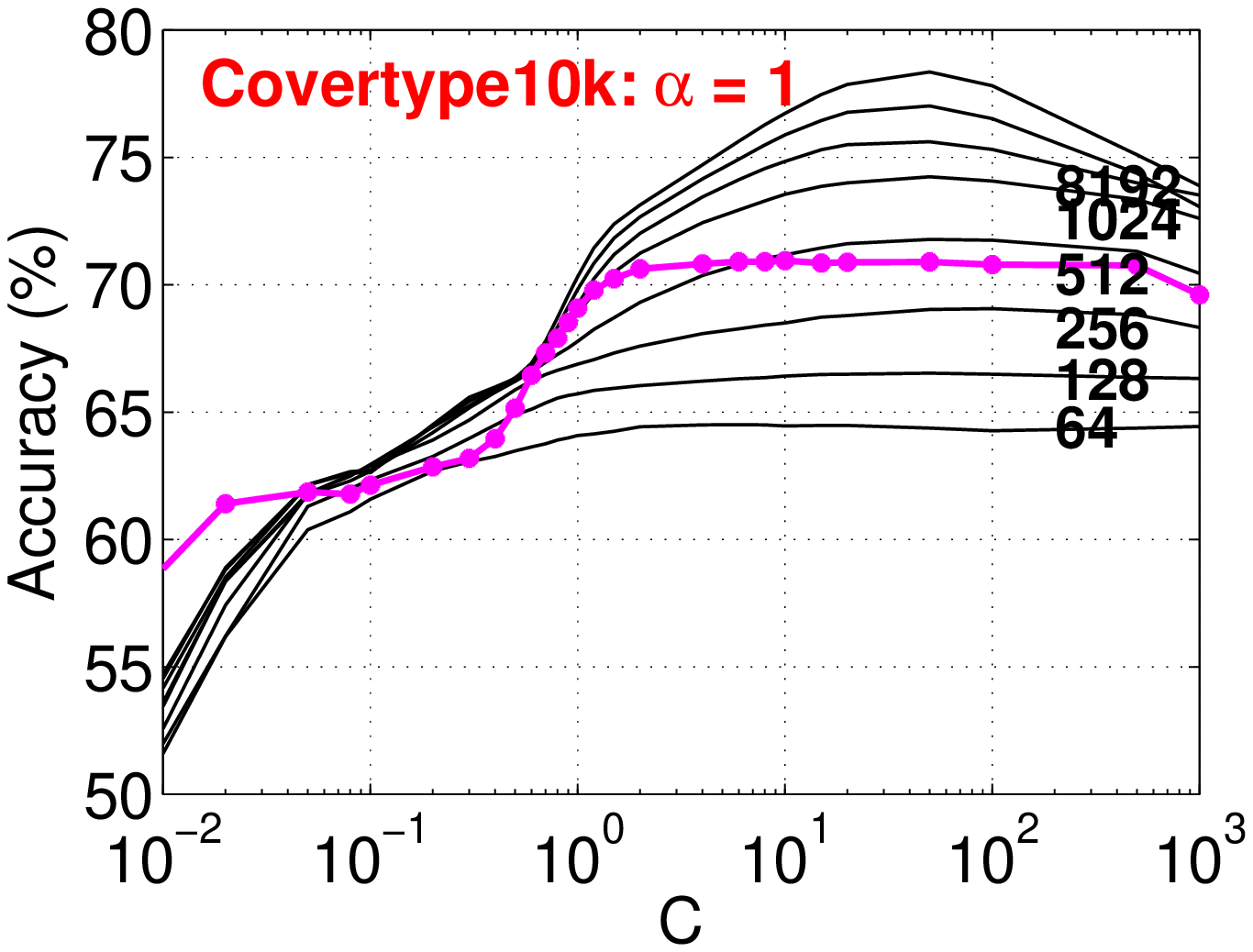}\hspace{-0.15in}
\includegraphics[width=2.3in]{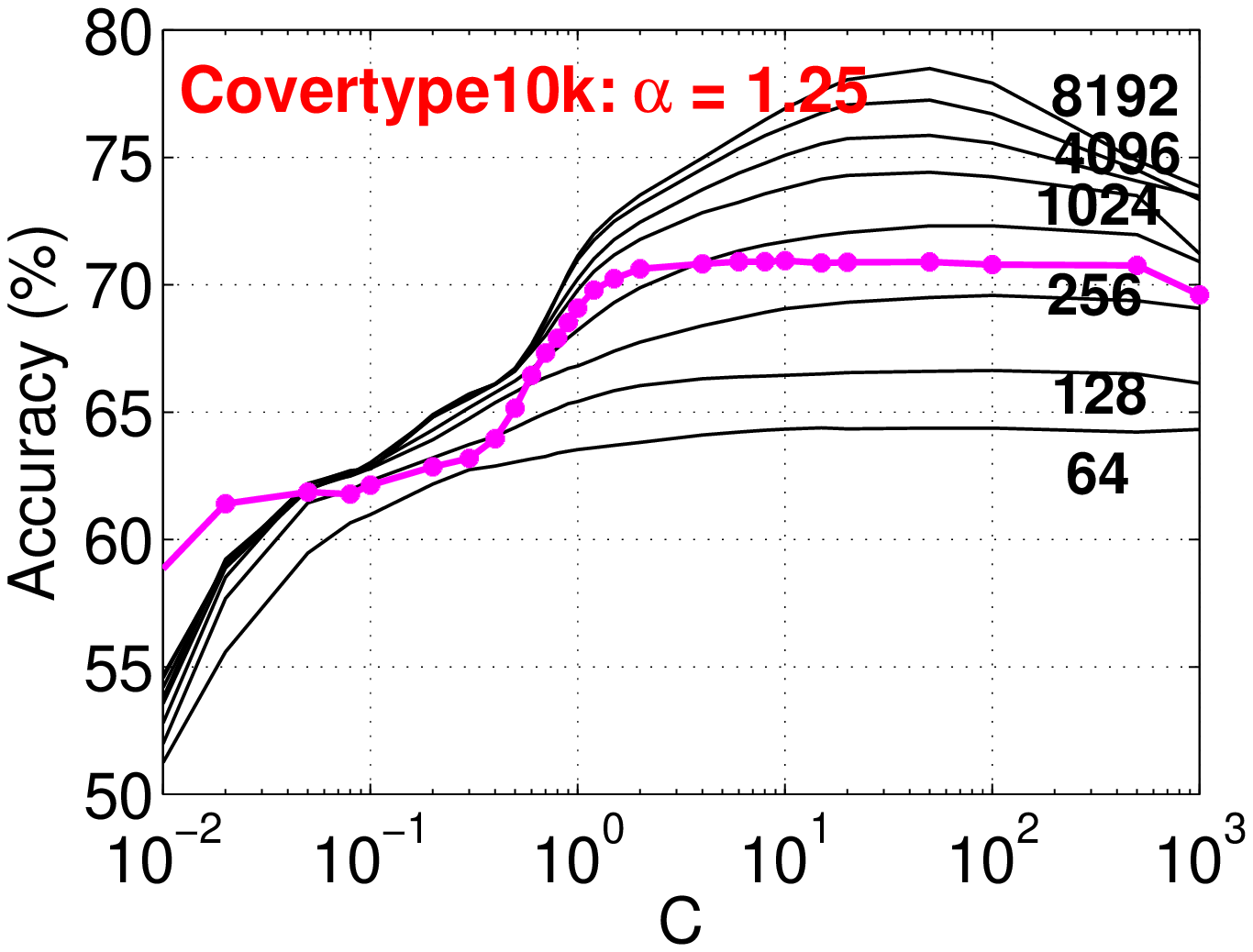}
}

\mbox{
\includegraphics[width=2.3in]{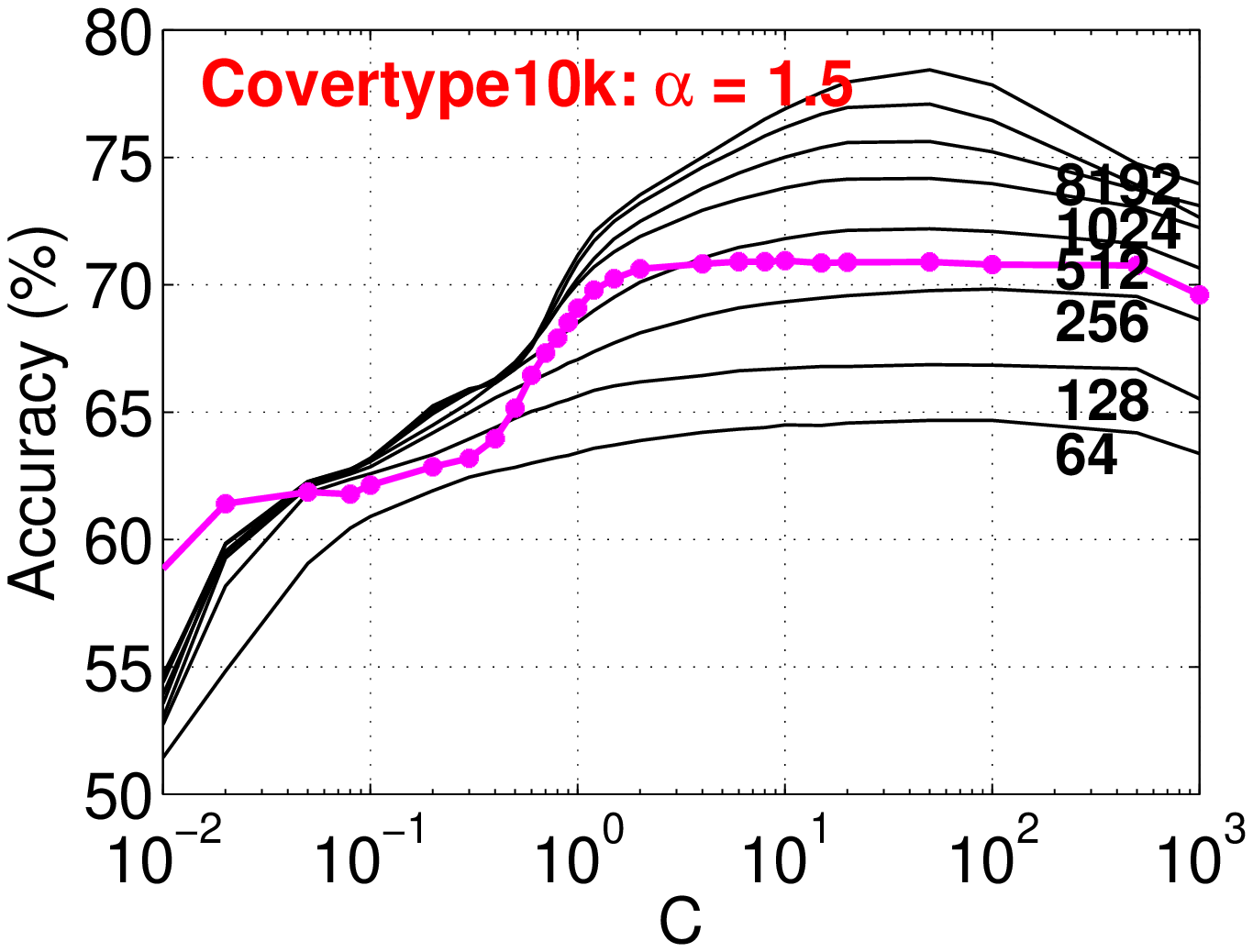}\hspace{-0.15in}
\includegraphics[width=2.3in]{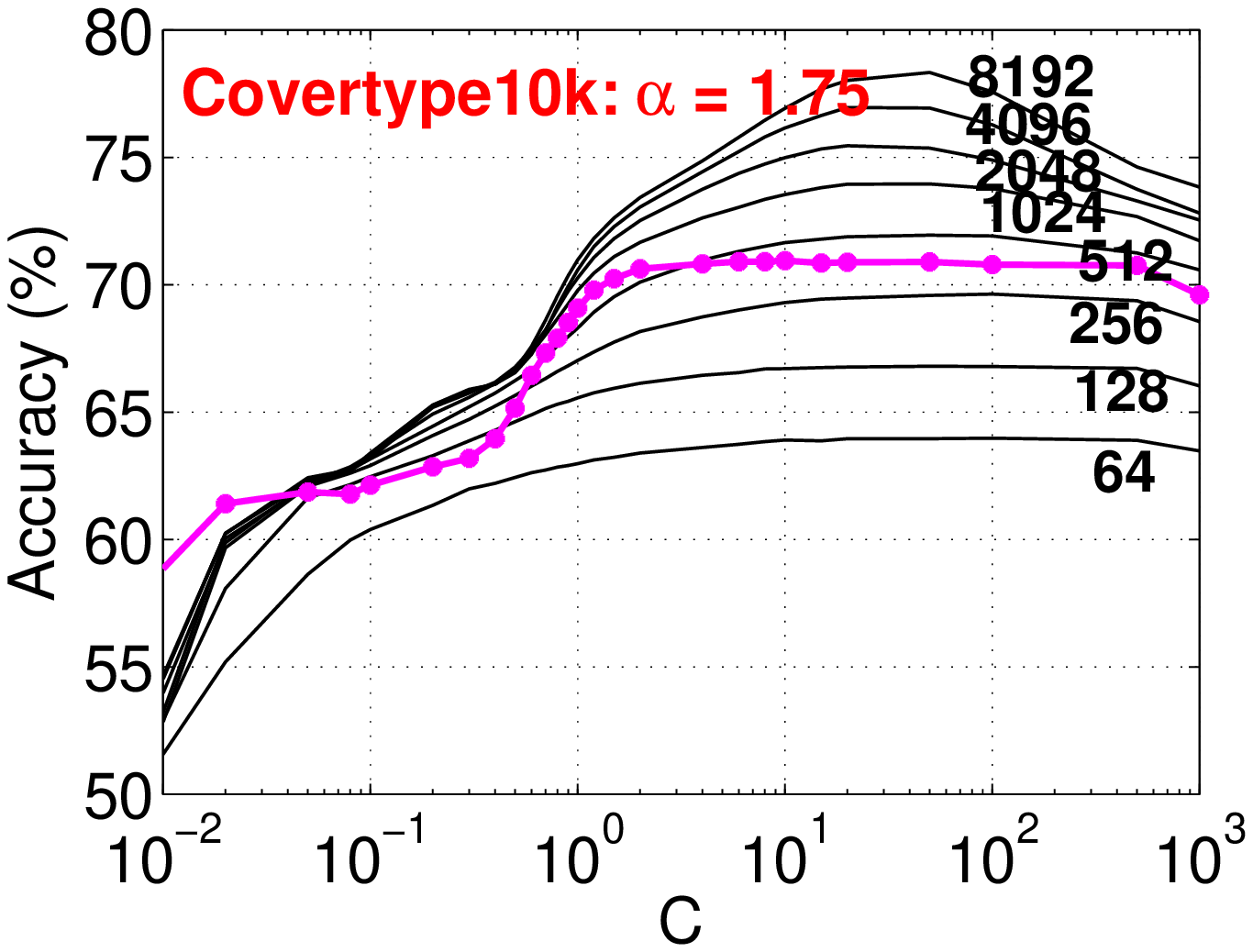}\hspace{-0.15in}
\includegraphics[width=2.3in]{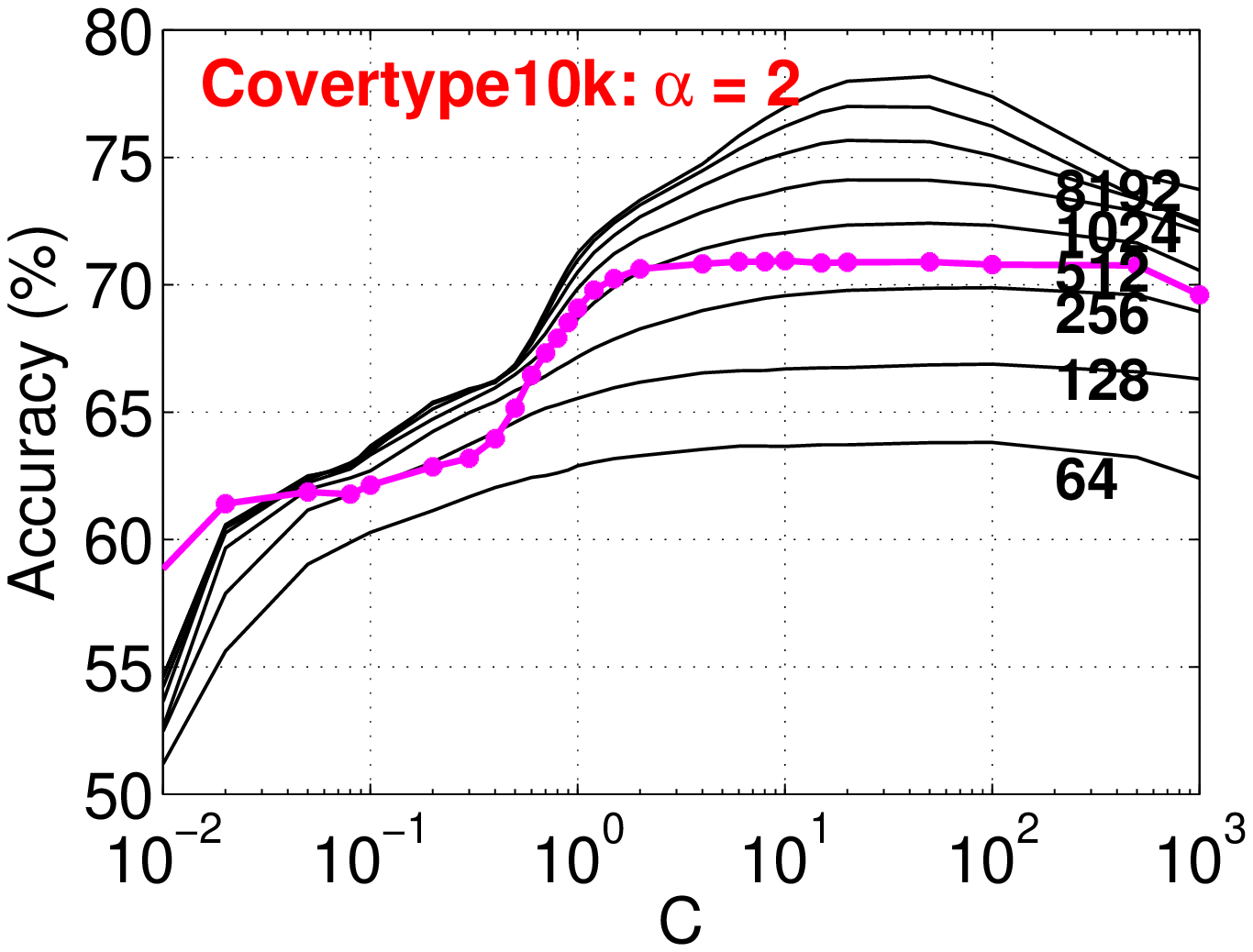}
}

\end{center}
\vspace{-0.2in}
\caption{\textbf{Covertype10k}. Classification accuracies of sign $\alpha$-stable random projections using $l_2$-regularized SVMs (with a tuning parameter $C\in[10^{-2},10^3]$) for $\alpha\in\{0.1,0.25,0.5,0.75,1,1.25,1.5,1.75,2\}$ and $k\in\{64,128,256,512,1024,2048,4096,8192\}$ projections. In each panel, the highest point (i.e., best accuracy) at $k=8192$ was reported in Table~\ref{tab_data}. In addition, each panel also presents the accuracies of linear SVM (the pink curve marked by *). All experiments were conducted by LIBLINEAR.}\label{fig_Covertype10kSRP}
\end{figure}

\begin{figure}[h!]
\begin{center}

\mbox{
\includegraphics[width=2.3in]{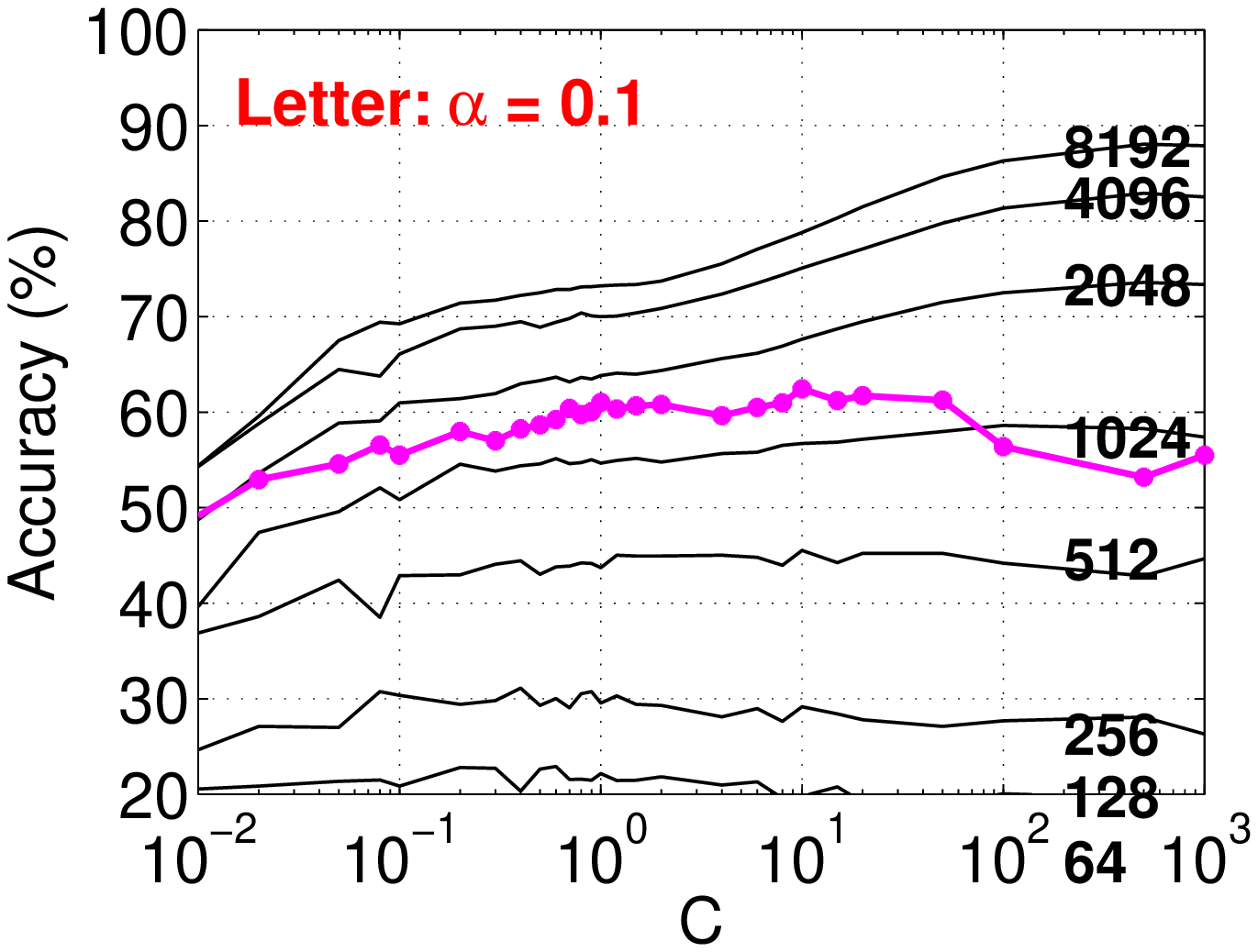}\hspace{-0.15in}
\includegraphics[width=2.3in]{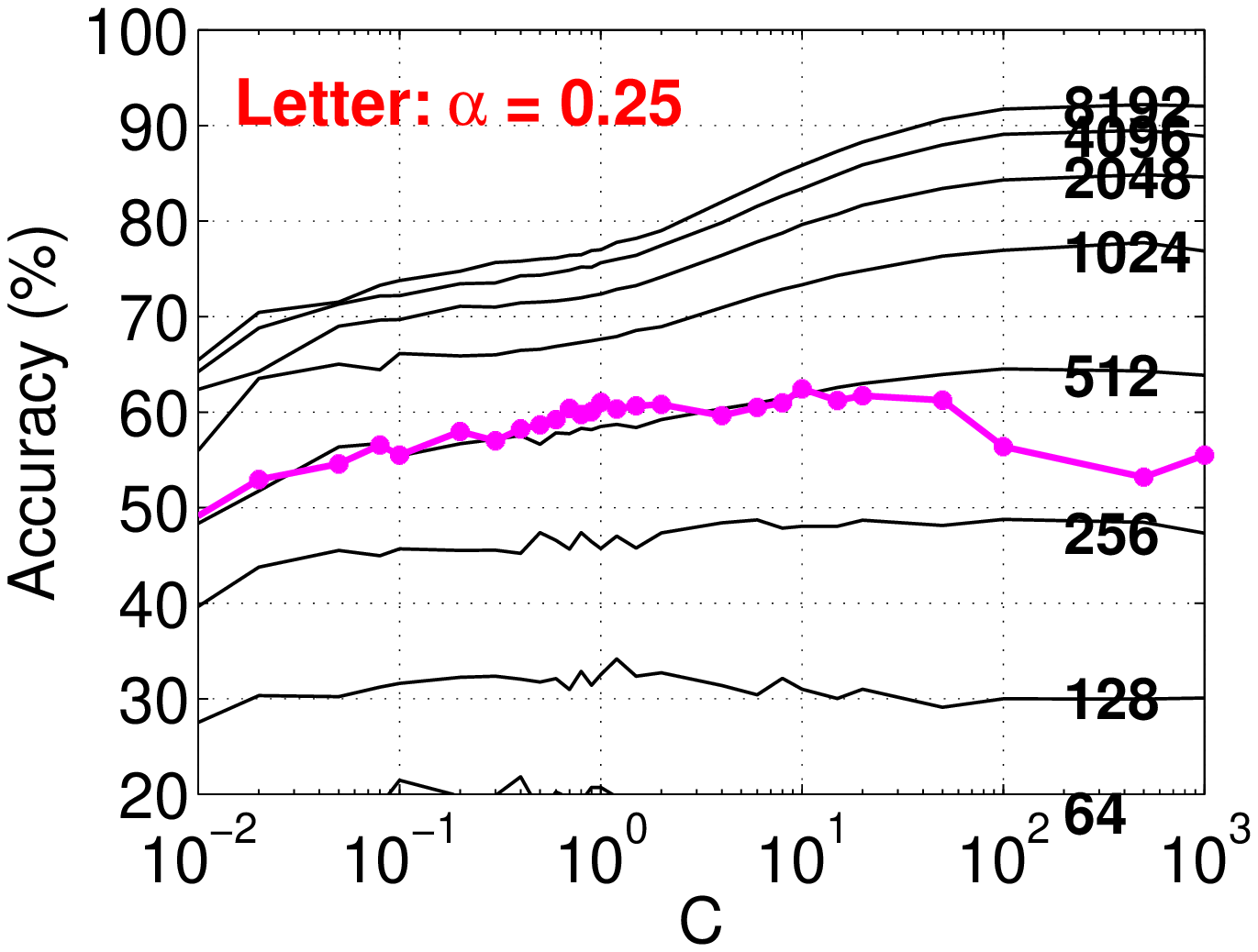}\hspace{-0.15in}
\includegraphics[width=2.3in]{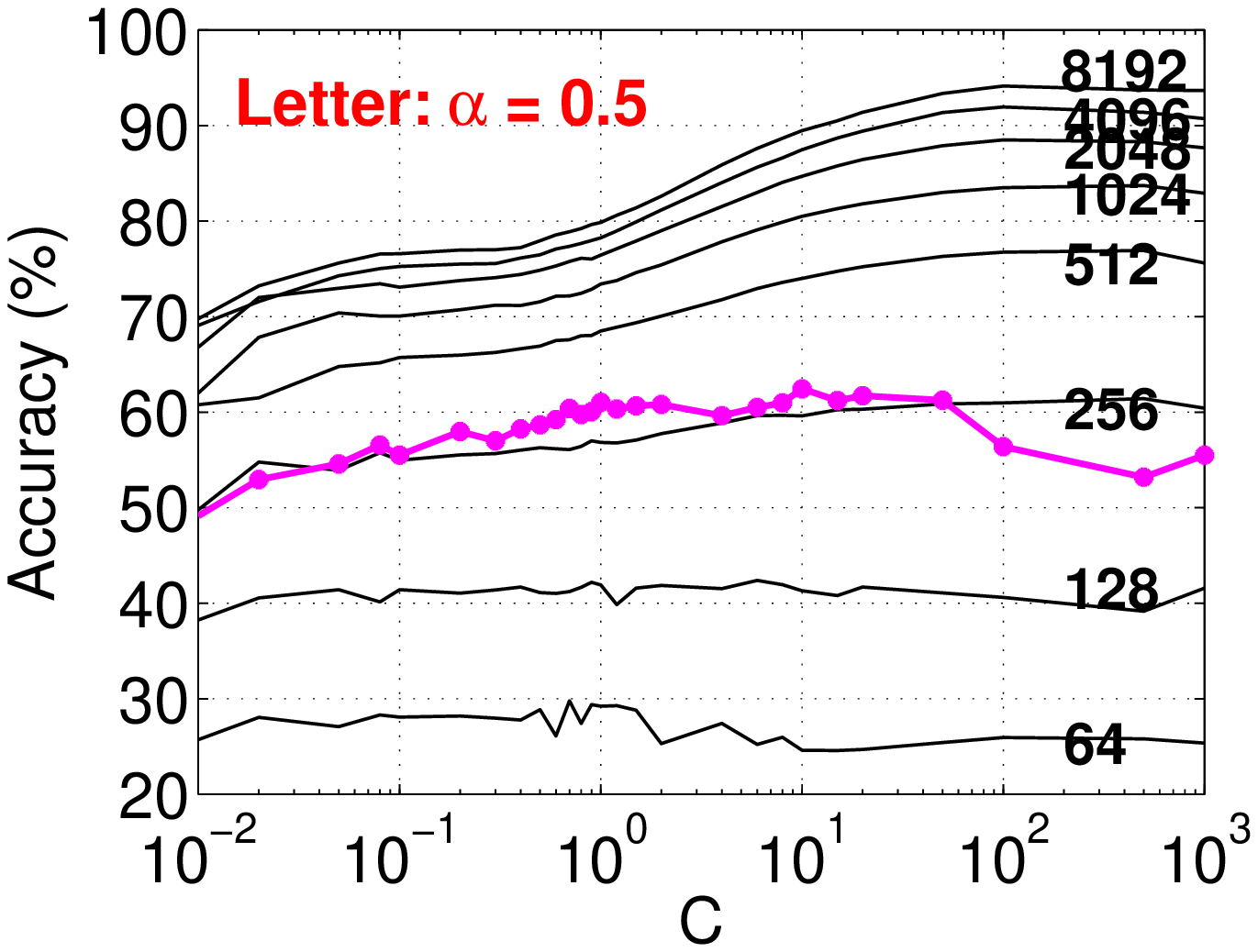}
}

\mbox{
\includegraphics[width=2.3in]{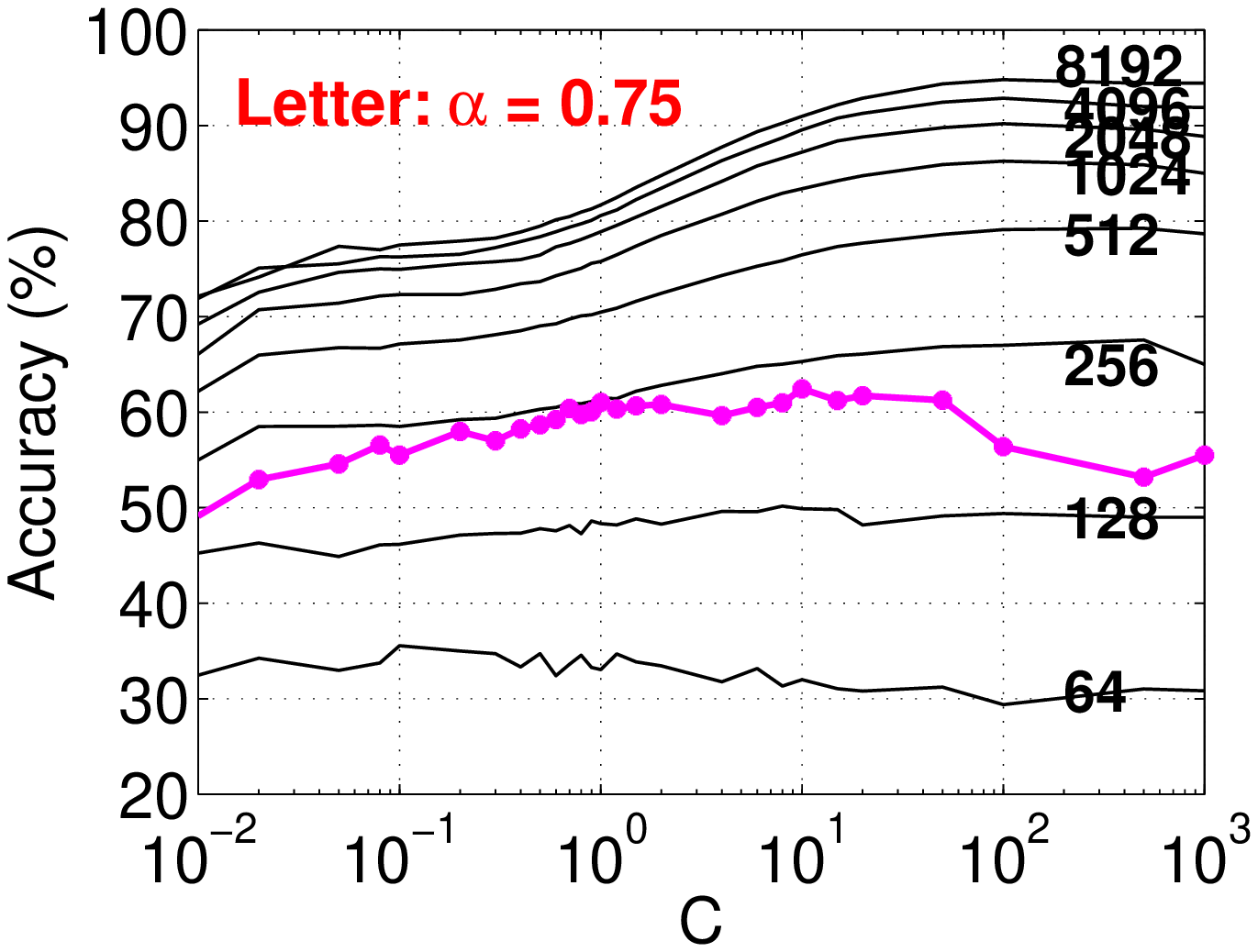}\hspace{-0.15in}
\includegraphics[width=2.3in]{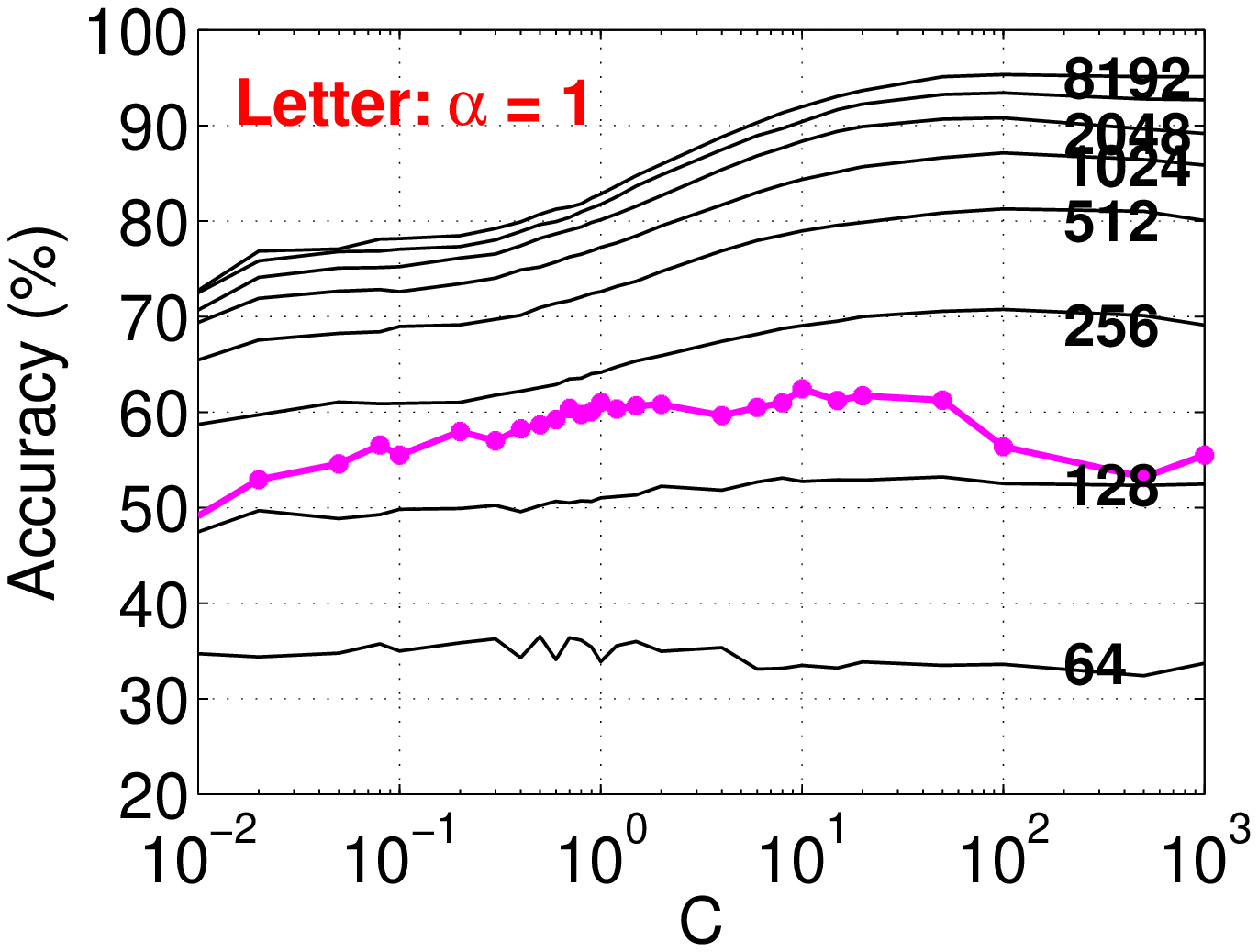}\hspace{-0.15in}
\includegraphics[width=2.3in]{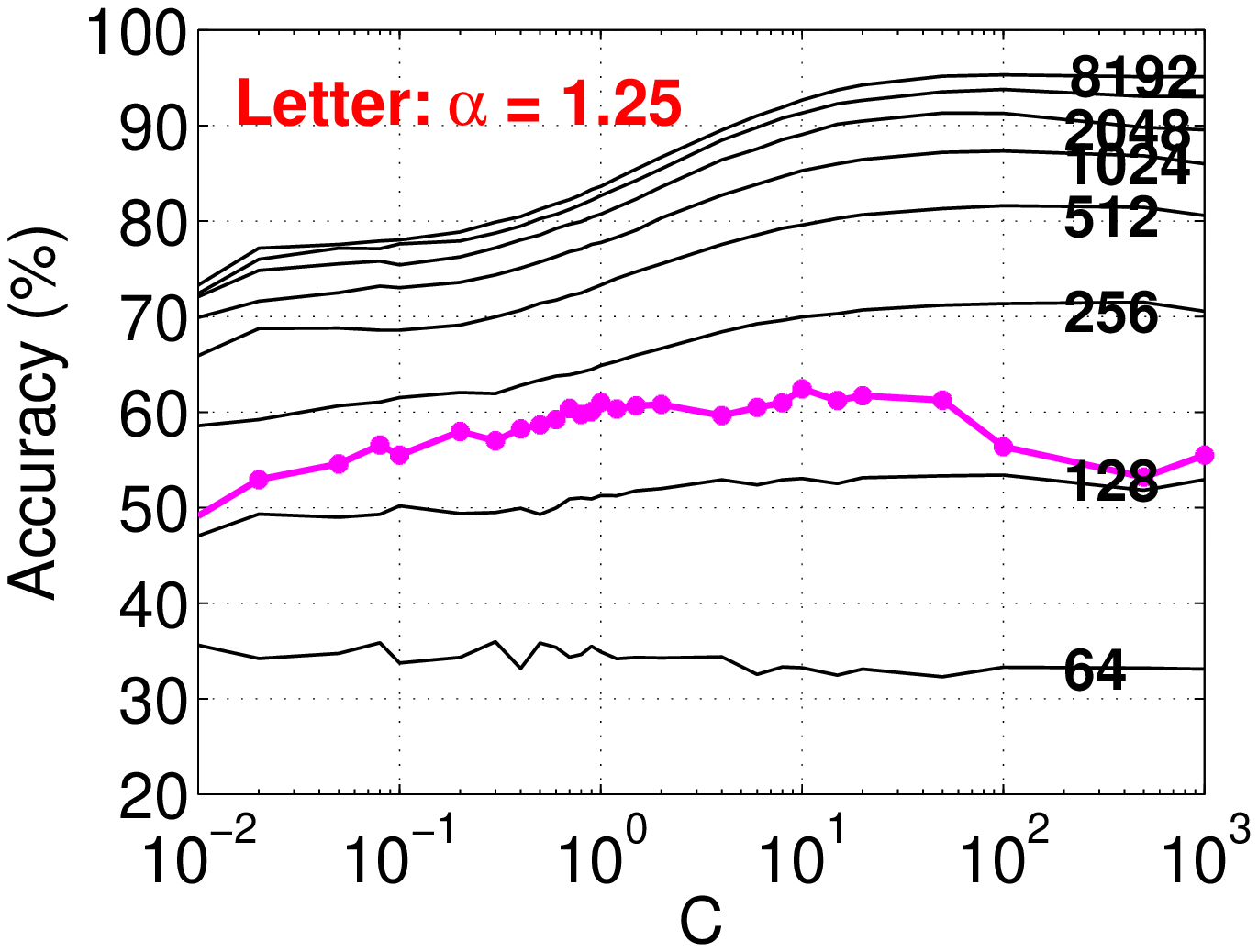}
}

\mbox{
\includegraphics[width=2.3in]{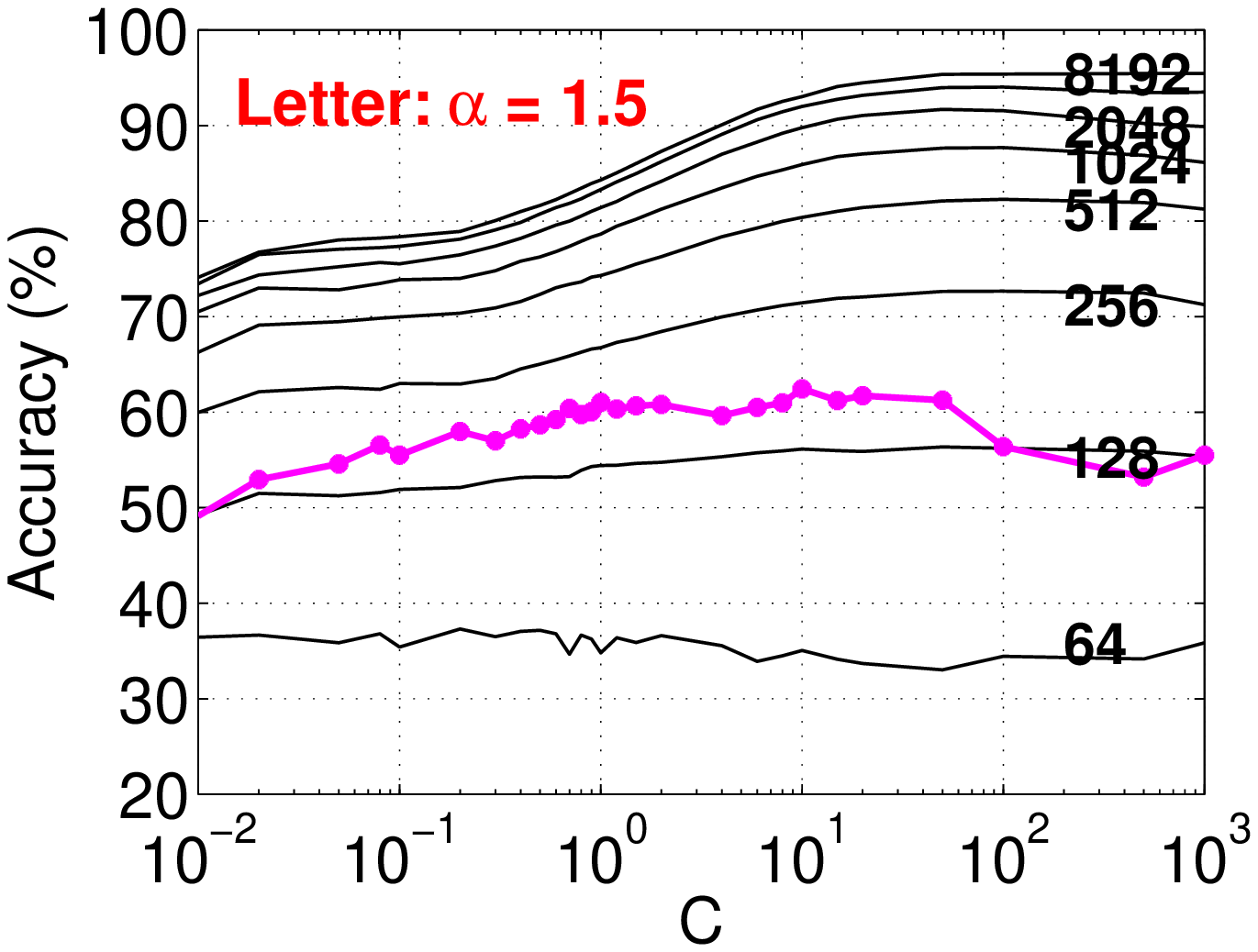}\hspace{-0.15in}
\includegraphics[width=2.3in]{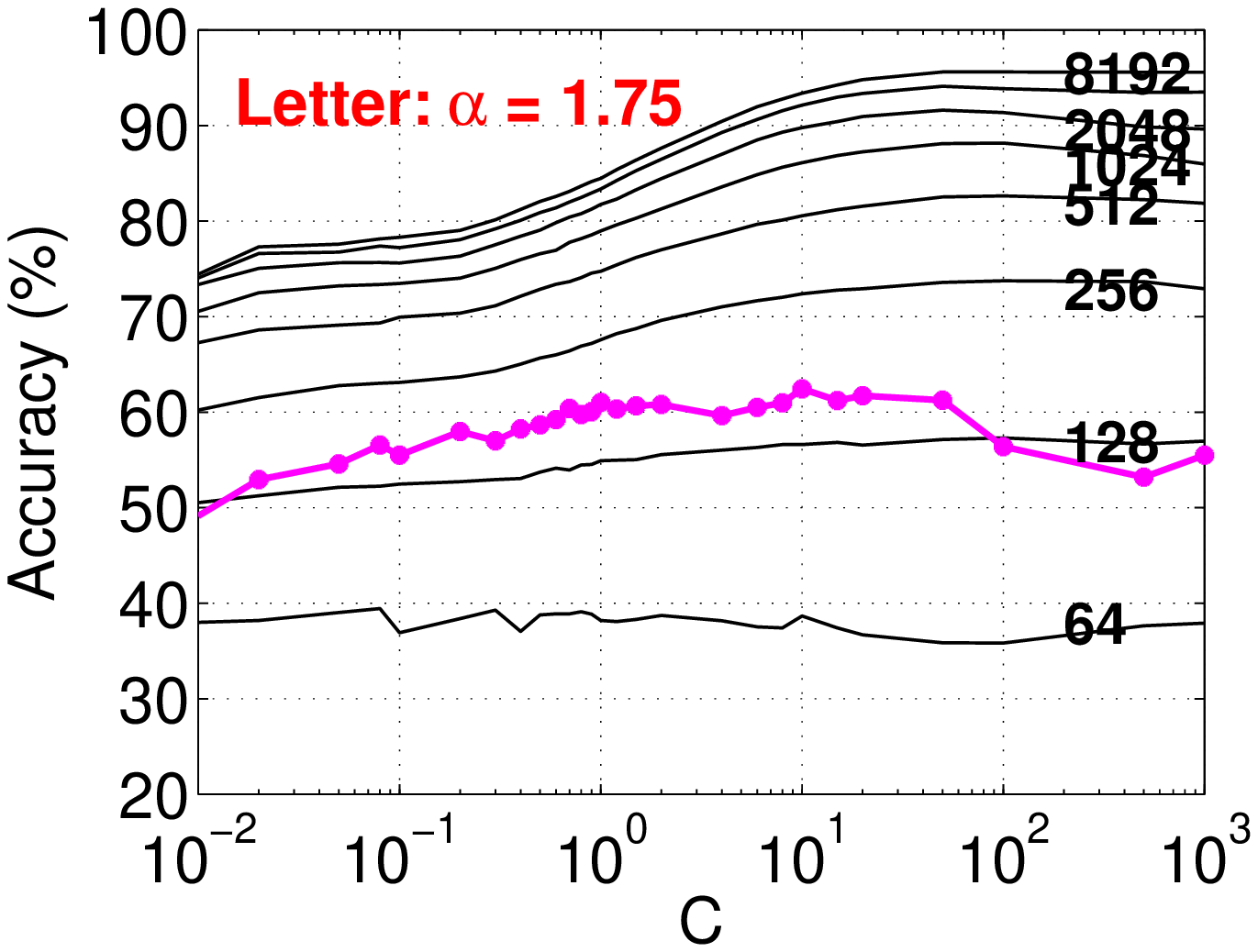}\hspace{-0.15in}
\includegraphics[width=2.3in]{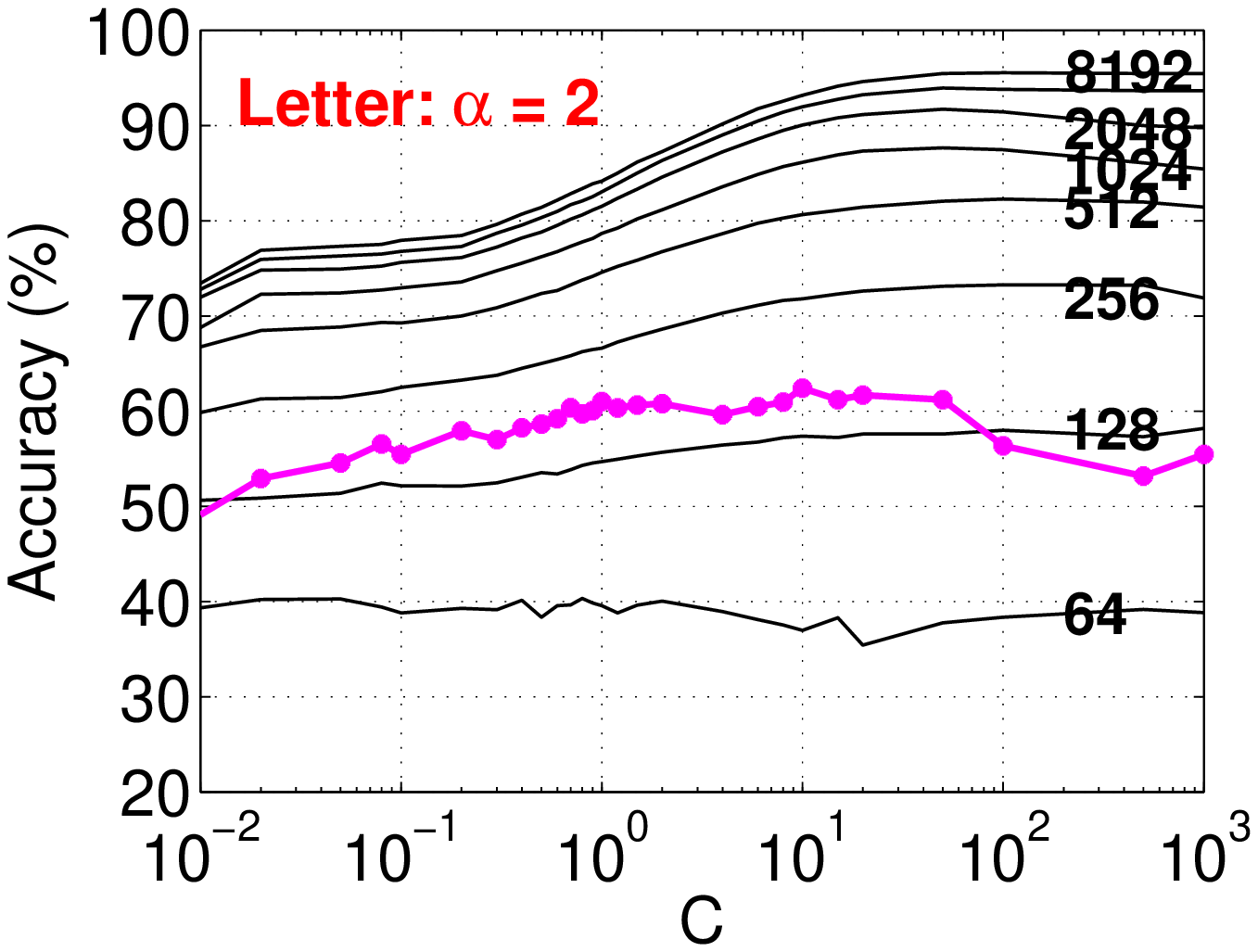}
}

\end{center}
\vspace{-0.2in}
\caption{\textbf{Letter}. Classification accuracies of sign $\alpha$-stable random projections using $l_2$-regularized SVMs (with a tuning parameter $C\in[10^{-2},10^3]$) for $\alpha\in\{0.1,0.25,0.5,0.75,1,1.25,1.5,1.75,2\}$ and $k\in\{64,128,256,512,1024,2048,4096,8192\}$ projections. In each panel, the highest point (i.e., best accuracy) at $k=8192$ was reported in Table~\ref{tab_data}. In addition, each panel also presents the accuracies of linear SVM (the pink curve marked by *). All experiments were conducted by LIBLINEAR.}\label{fig_LetterSRP}
\end{figure}

\begin{figure}[h!]
\begin{center}

\mbox{
\includegraphics[width=2.3in]{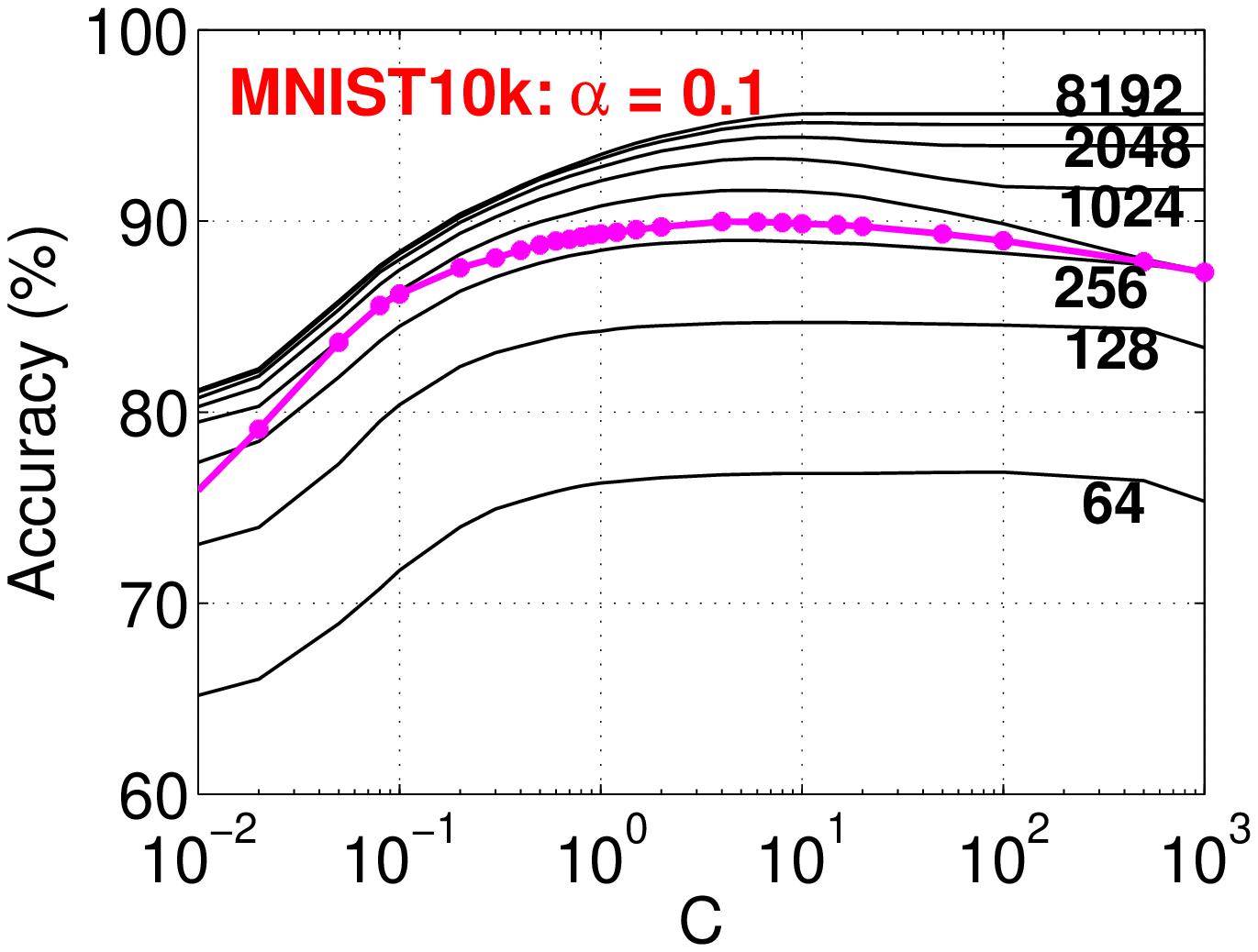}\hspace{-0.15in}
\includegraphics[width=2.3in]{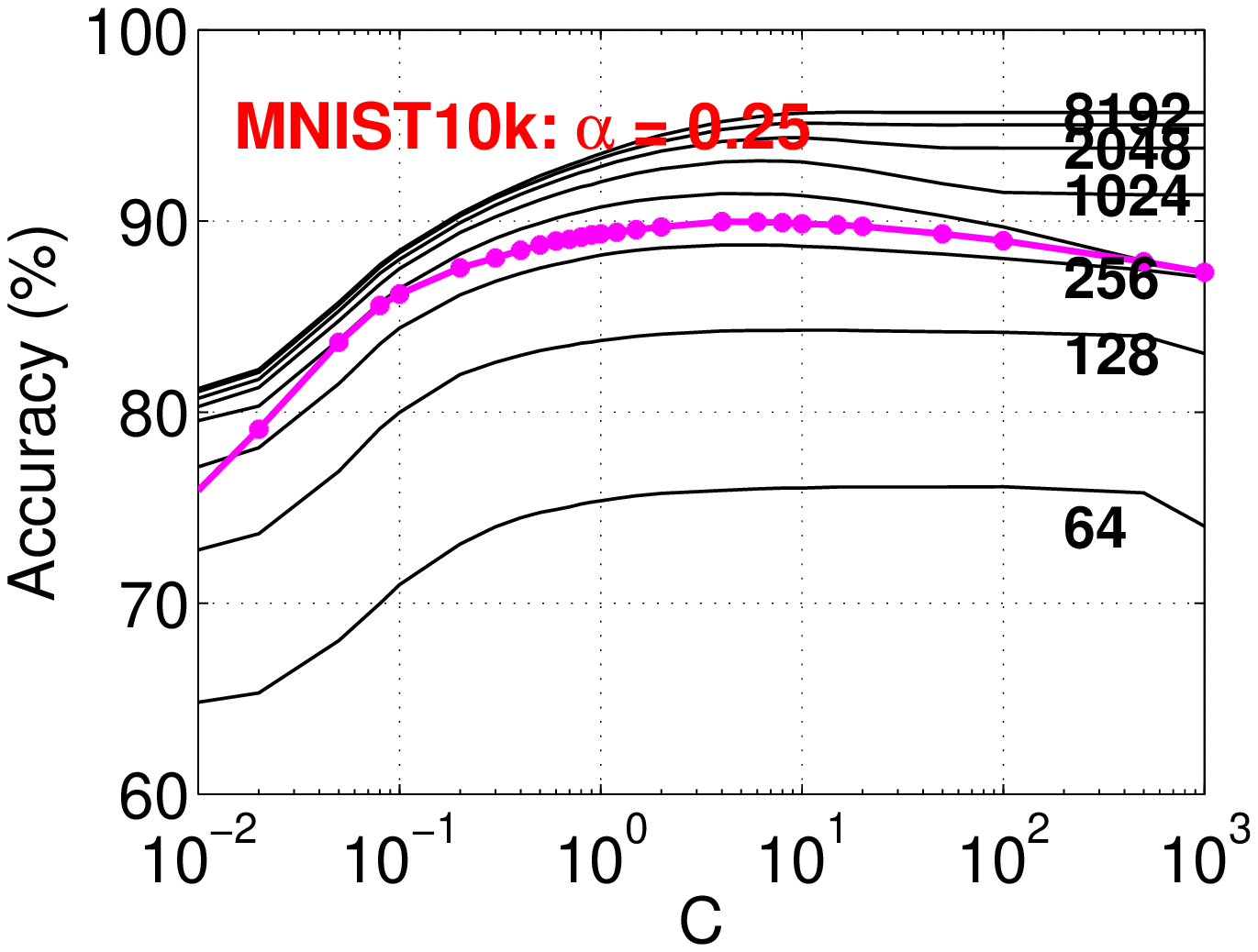}\hspace{-0.15in}
\includegraphics[width=2.3in]{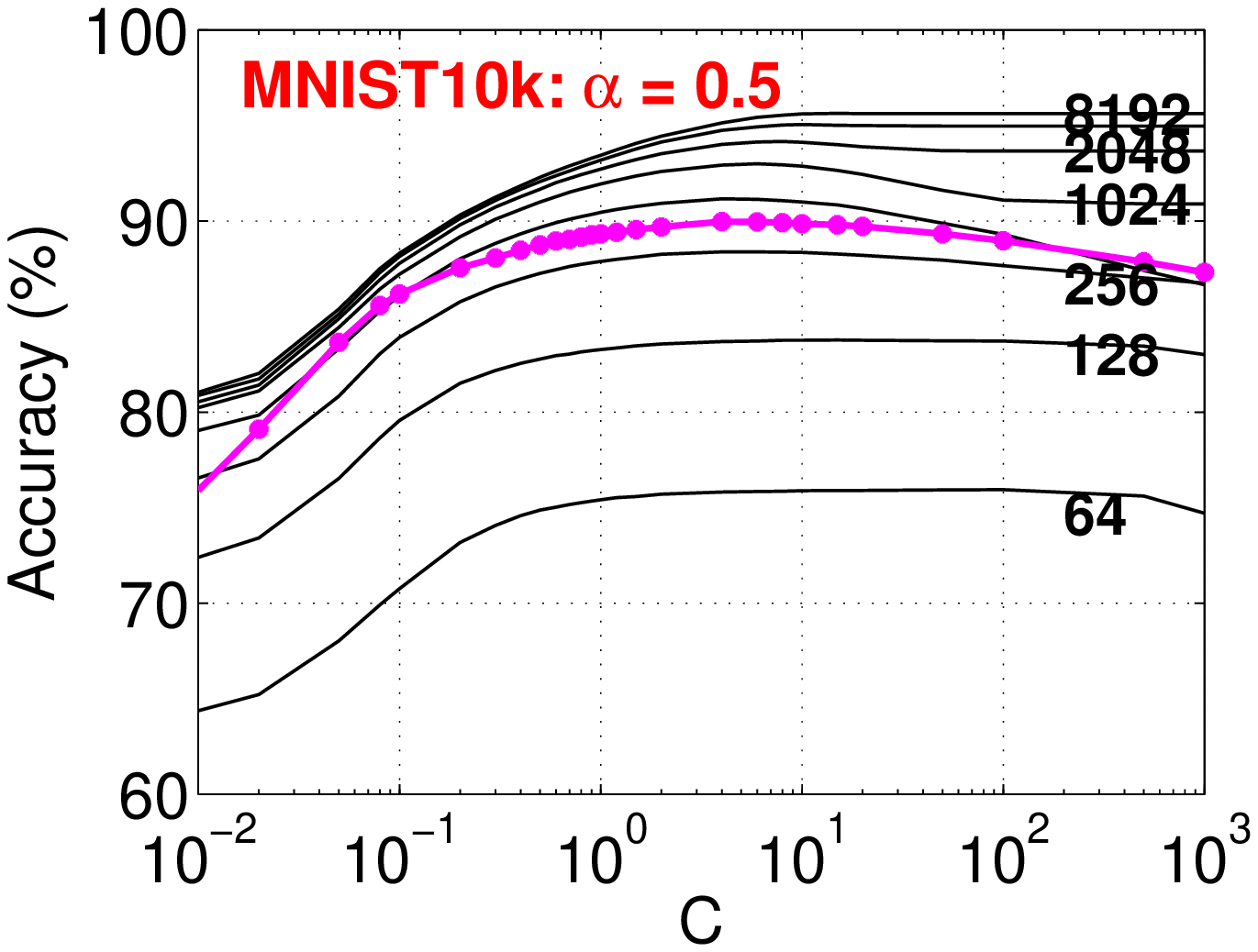}
}

\mbox{
\includegraphics[width=2.3in]{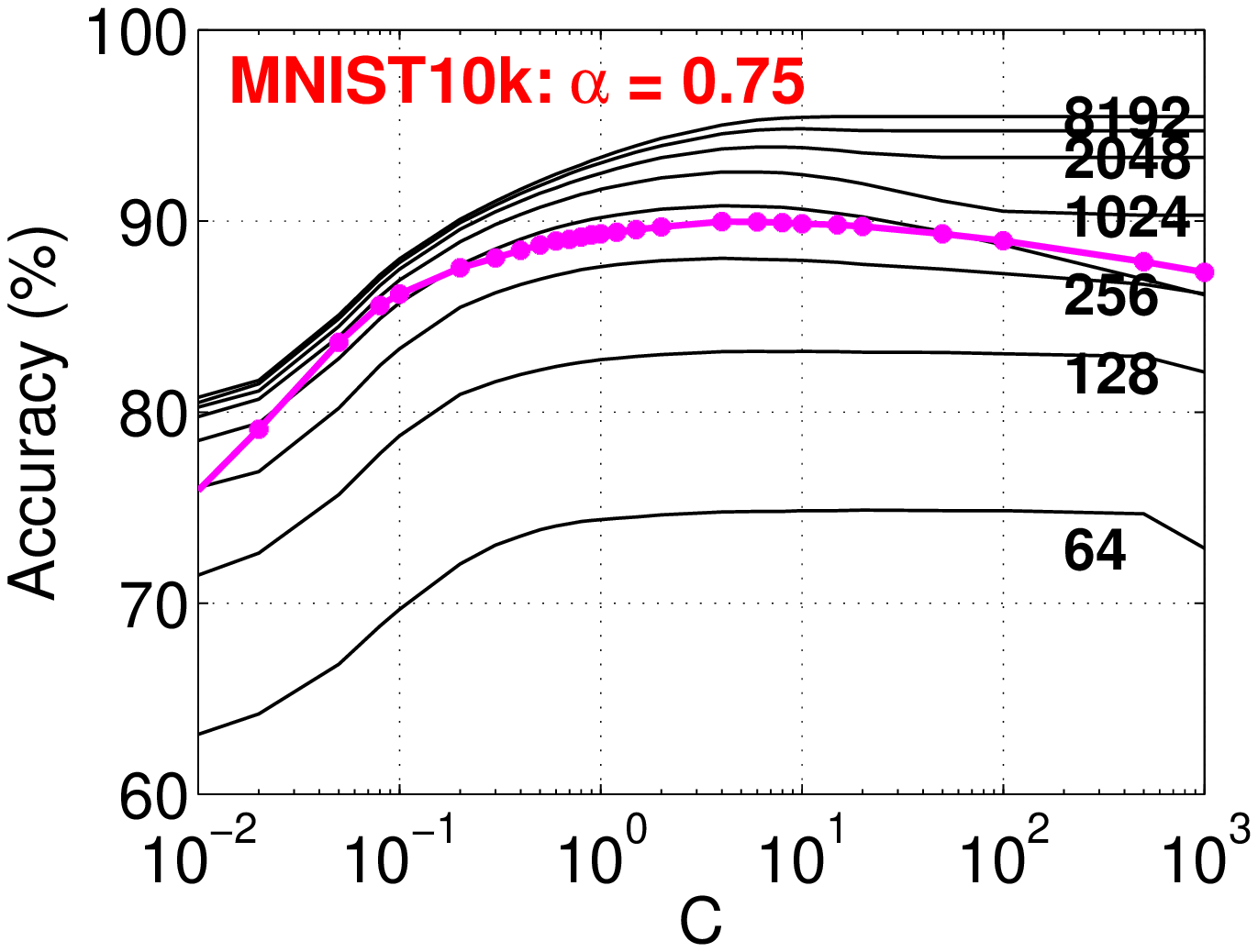}\hspace{-0.15in}
\includegraphics[width=2.3in]{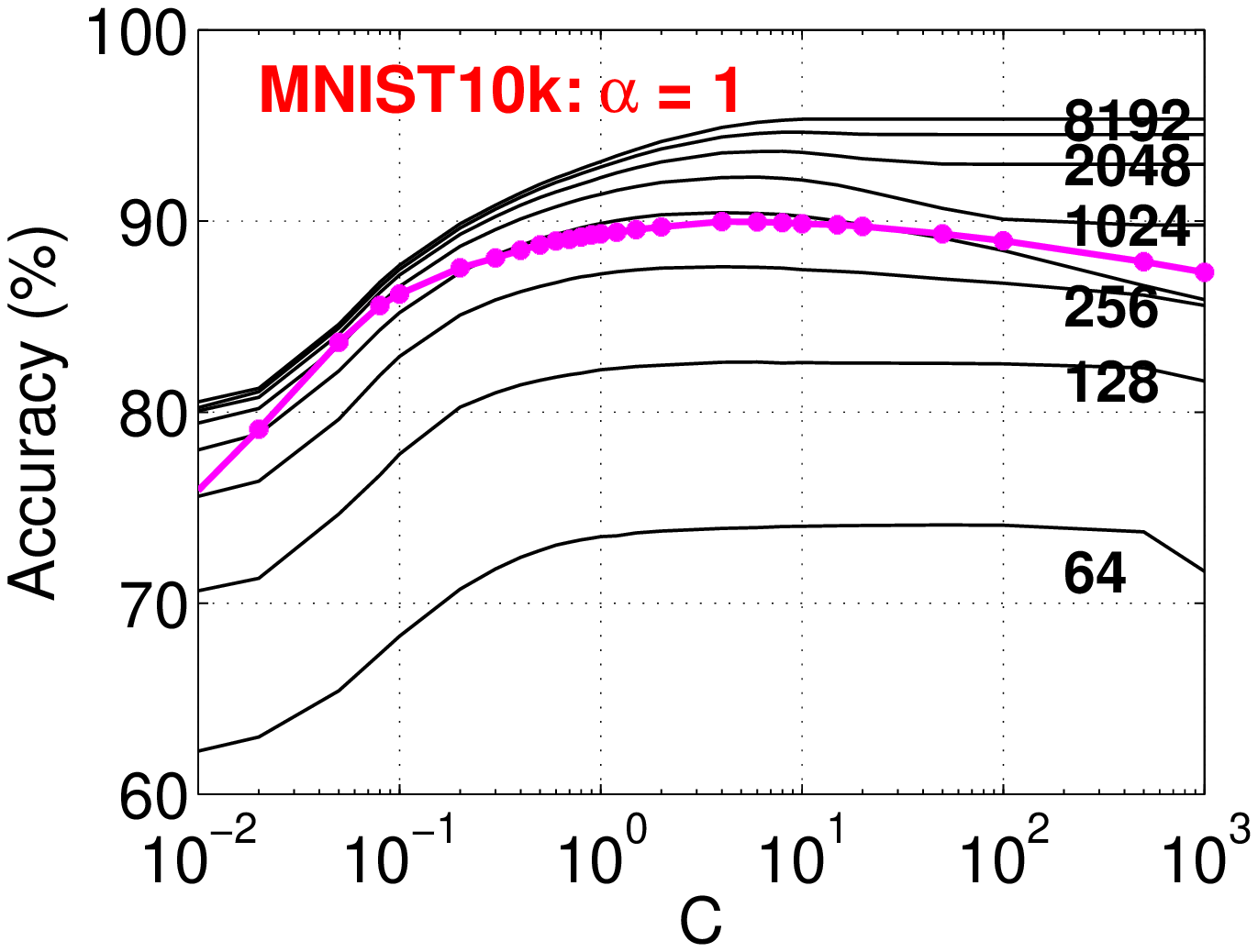}\hspace{-0.15in}
\includegraphics[width=2.3in]{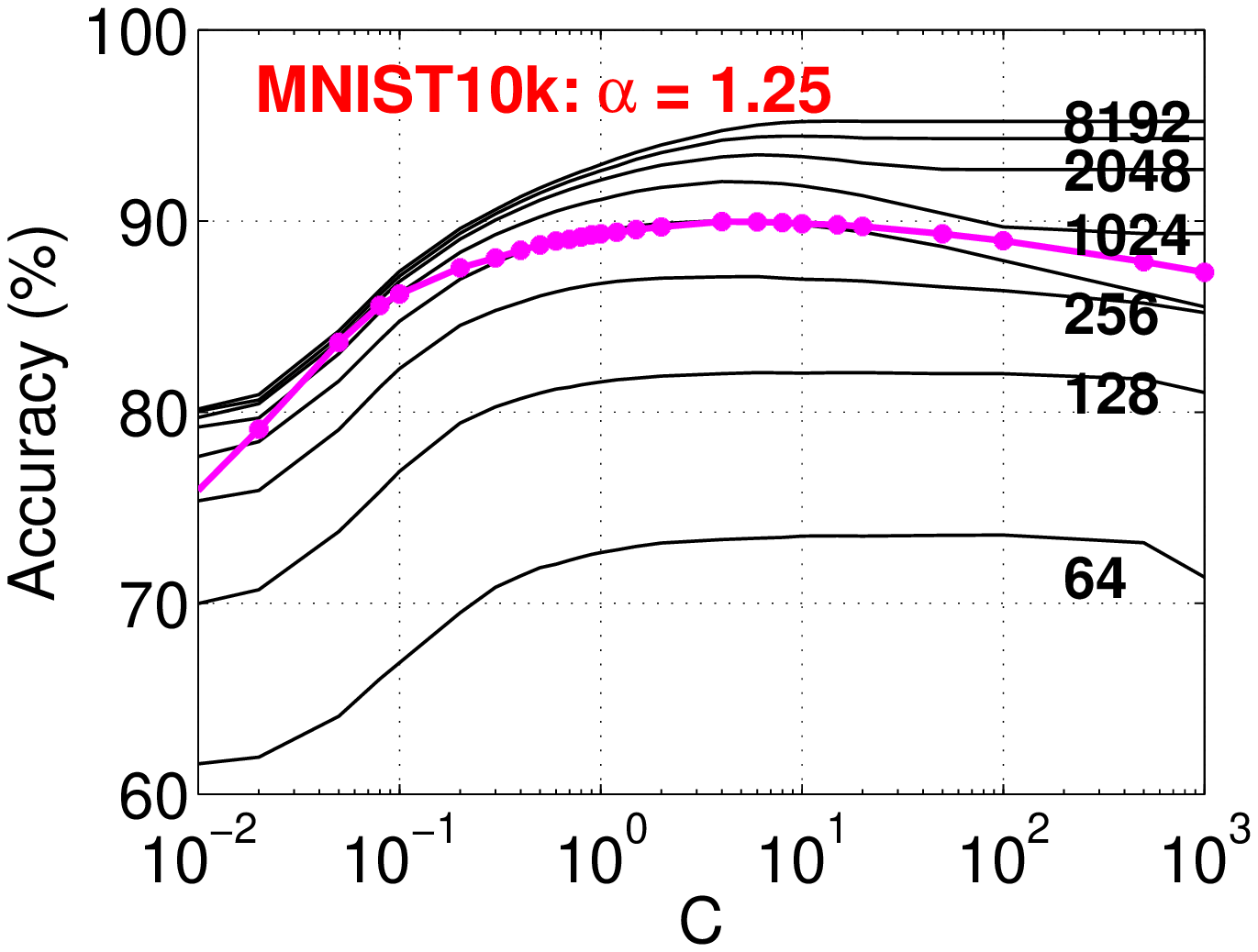}
}

\mbox{
\includegraphics[width=2.3in]{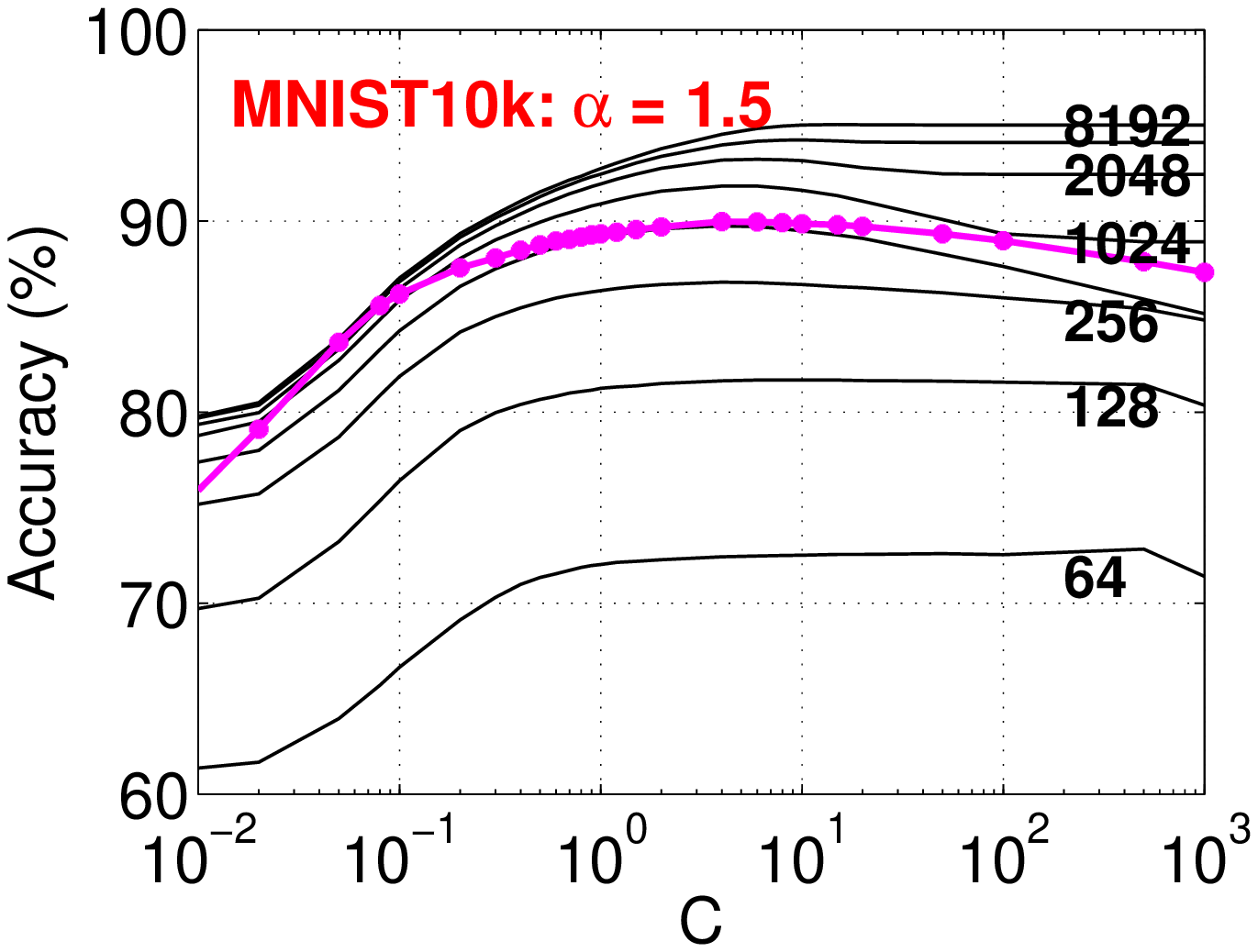}\hspace{-0.15in}
\includegraphics[width=2.3in]{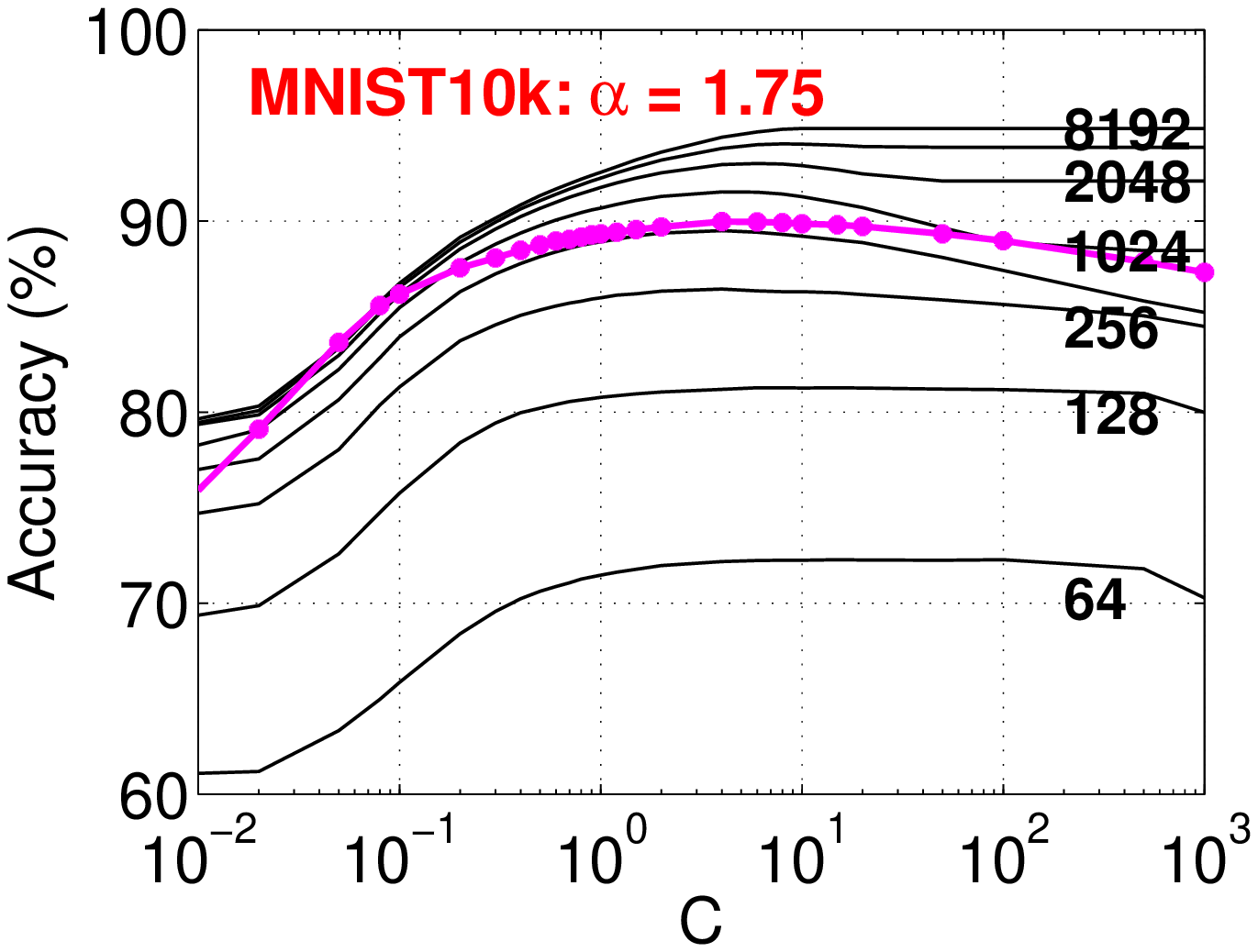}\hspace{-0.15in}
\includegraphics[width=2.3in]{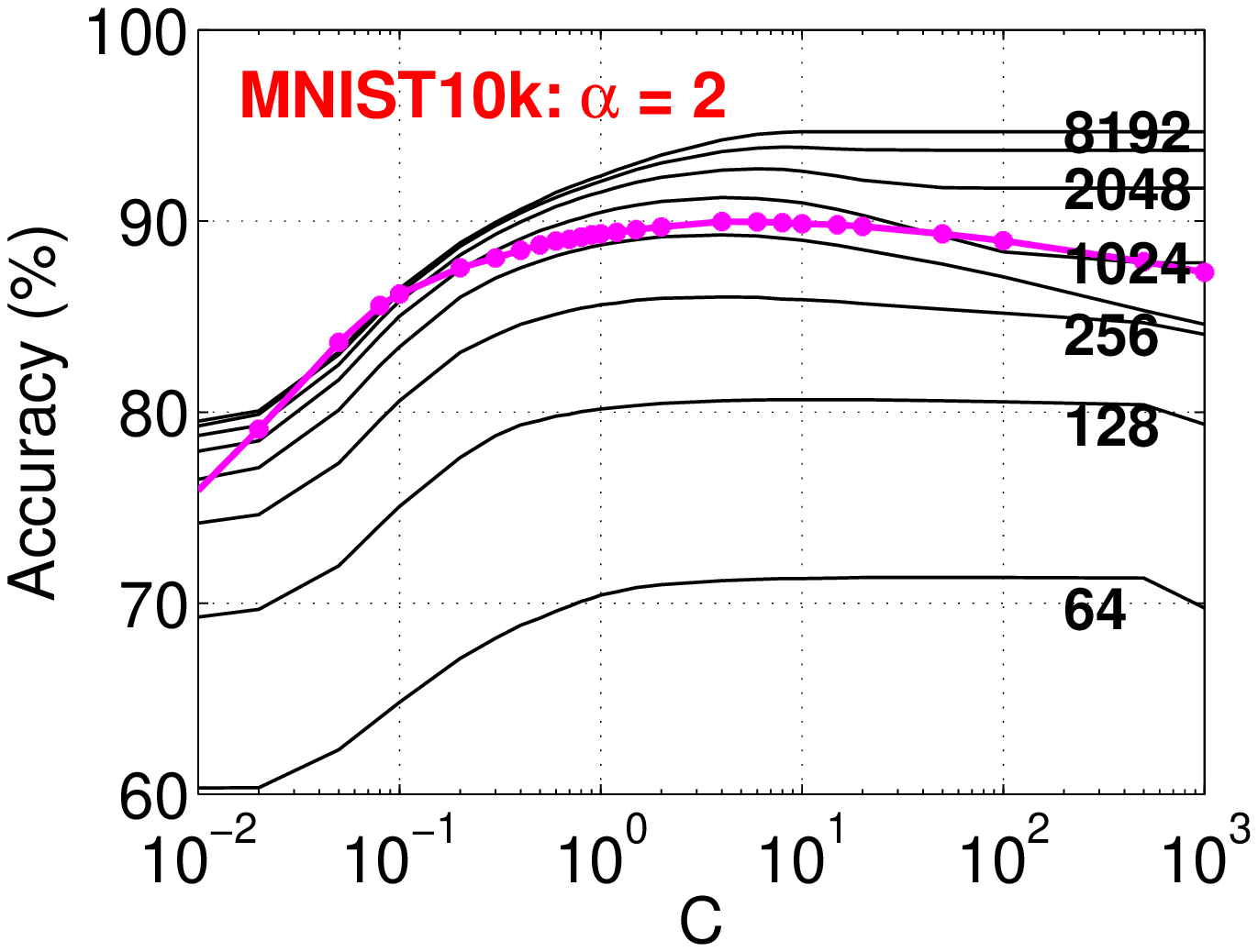}
}

\end{center}
\vspace{-0.2in}
\caption{\textbf{MNIST10k}. Classification accuracies of sign $\alpha$-stable random projections using $l_2$-regularized SVMs (with a tuning parameter $C\in[10^{-2},10^3]$) for $\alpha\in\{0.1,0.25,0.5,0.75,1,1.25,1.5,1.75,2\}$ and $k\in\{64,128,256,512,1024,2048,4096,8192\}$ projections. In each panel, the highest point (i.e., best accuracy) at $k=8192$ was reported in Table~\ref{tab_data}. In addition, each panel also presents the accuracies of linear SVM (the pink curve marked by *). All experiments were conducted by LIBLINEAR. }\label{fig_MNIST10kSRP}
\end{figure}

\begin{figure}[h!]
\begin{center}

\mbox{
\includegraphics[width=2.3in]{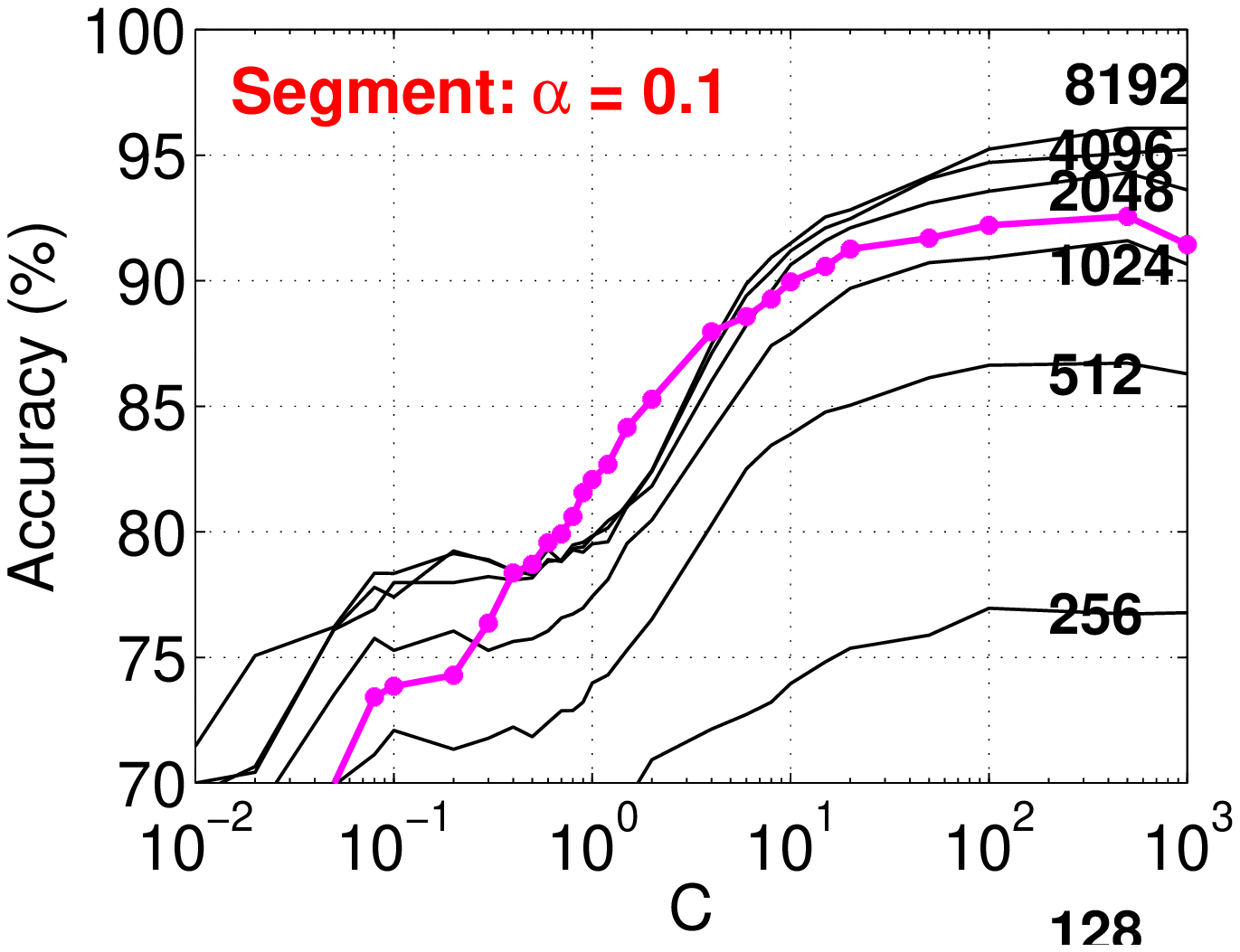}\hspace{-0.15in}
\includegraphics[width=2.3in]{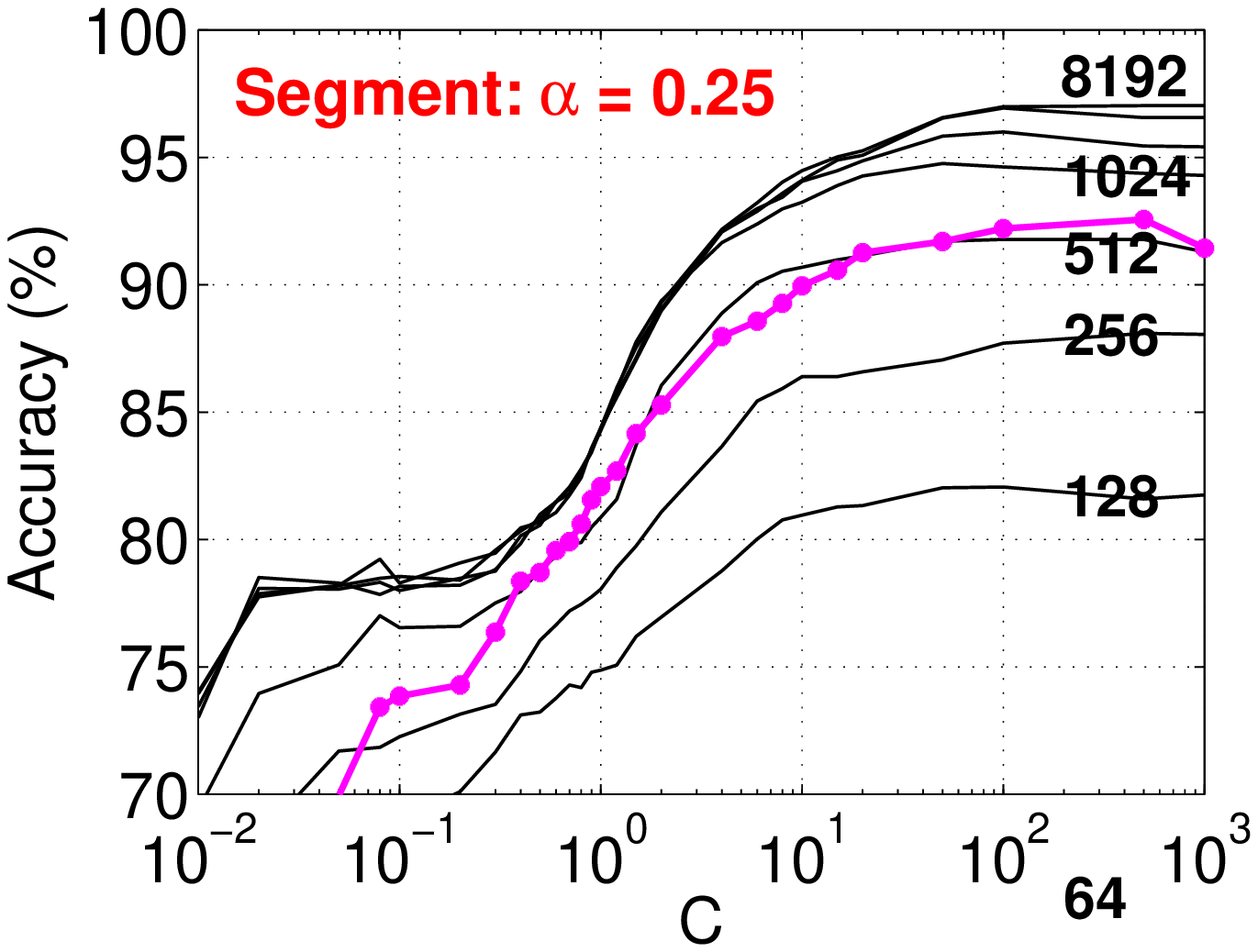}\hspace{-0.15in}
\includegraphics[width=2.3in]{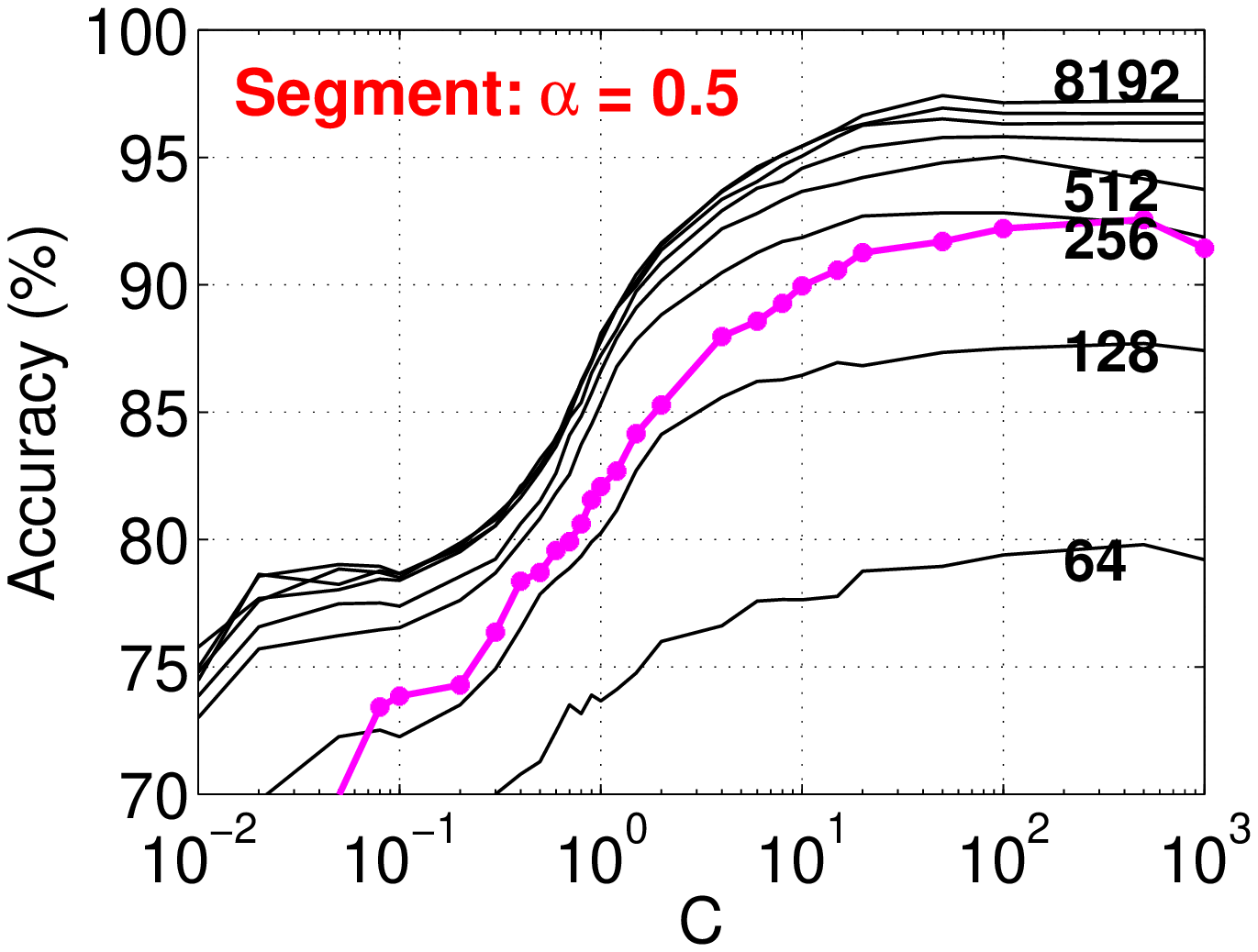}
}

\mbox{
\includegraphics[width=2.3in]{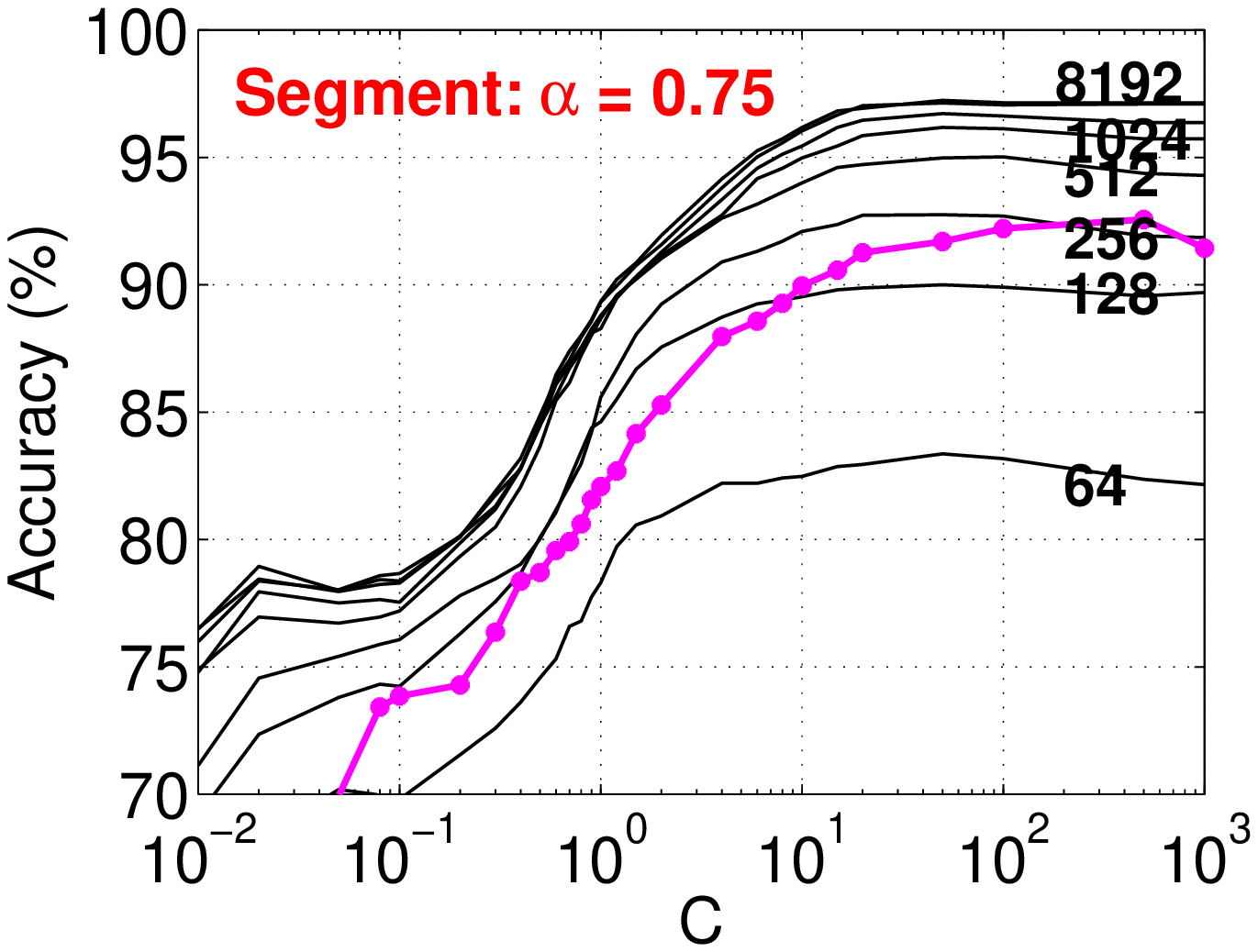}\hspace{-0.15in}
\includegraphics[width=2.3in]{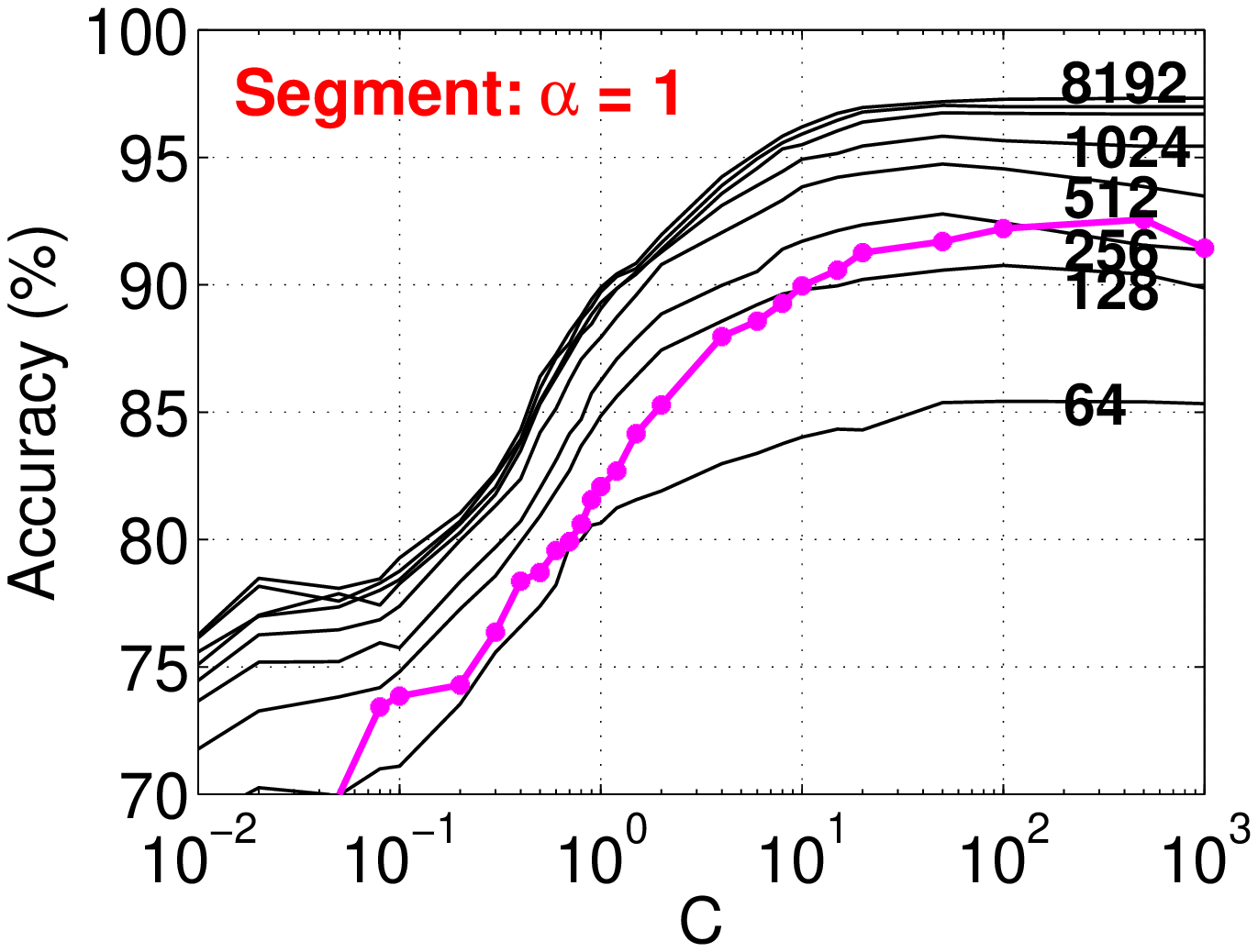}\hspace{-0.15in}
\includegraphics[width=2.3in]{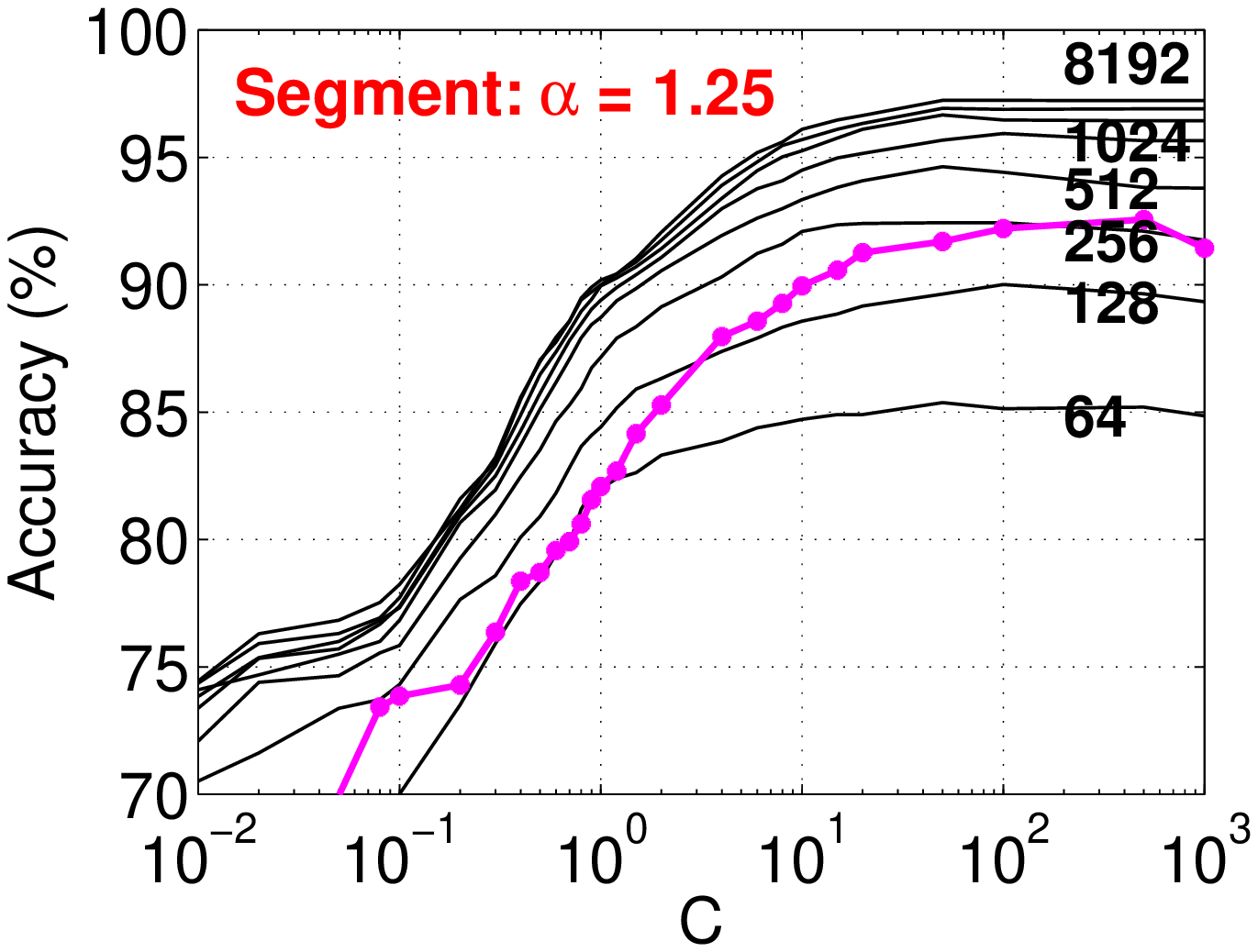}
}

\mbox{
\includegraphics[width=2.3in]{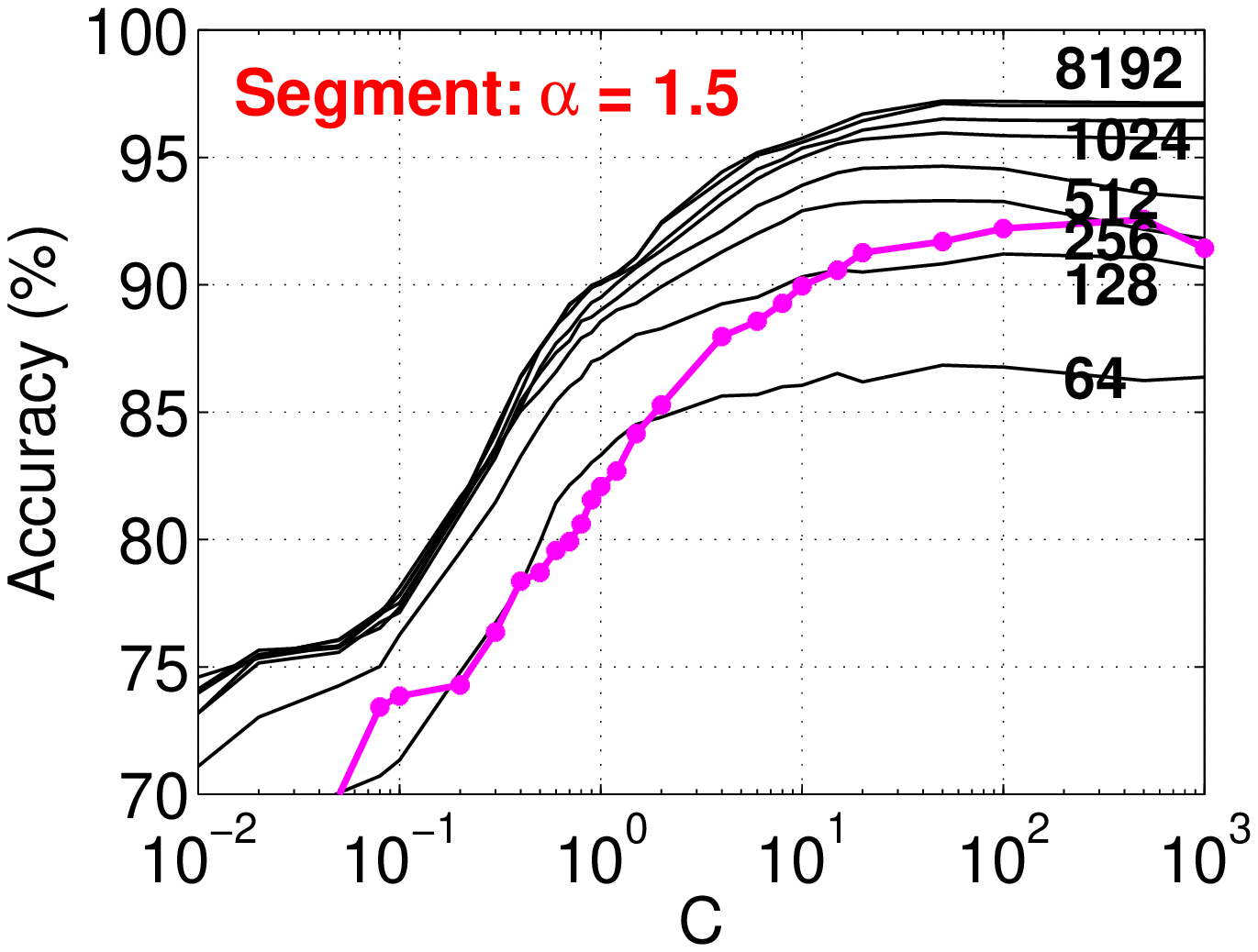}\hspace{-0.15in}
\includegraphics[width=2.3in]{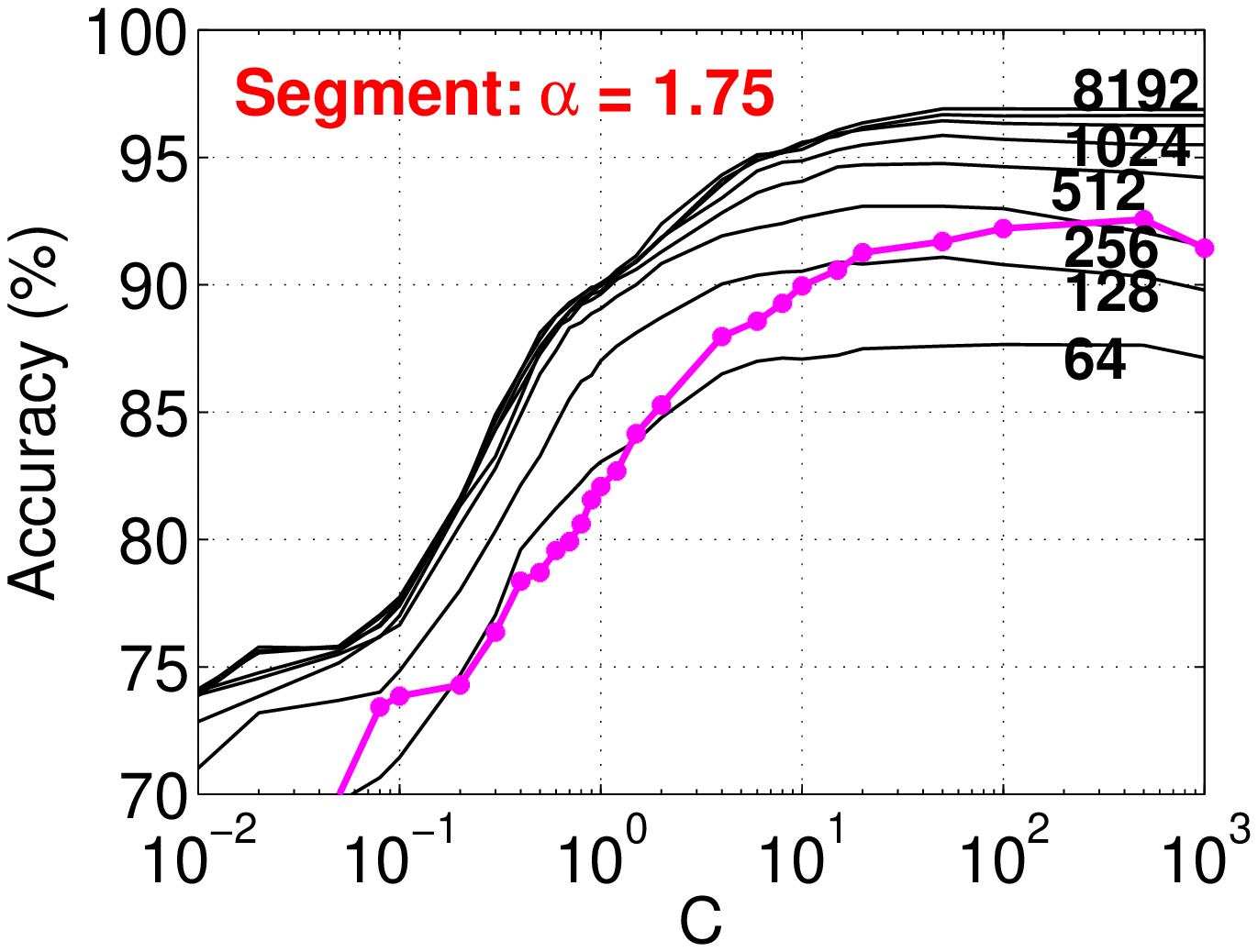}\hspace{-0.15in}
\includegraphics[width=2.3in]{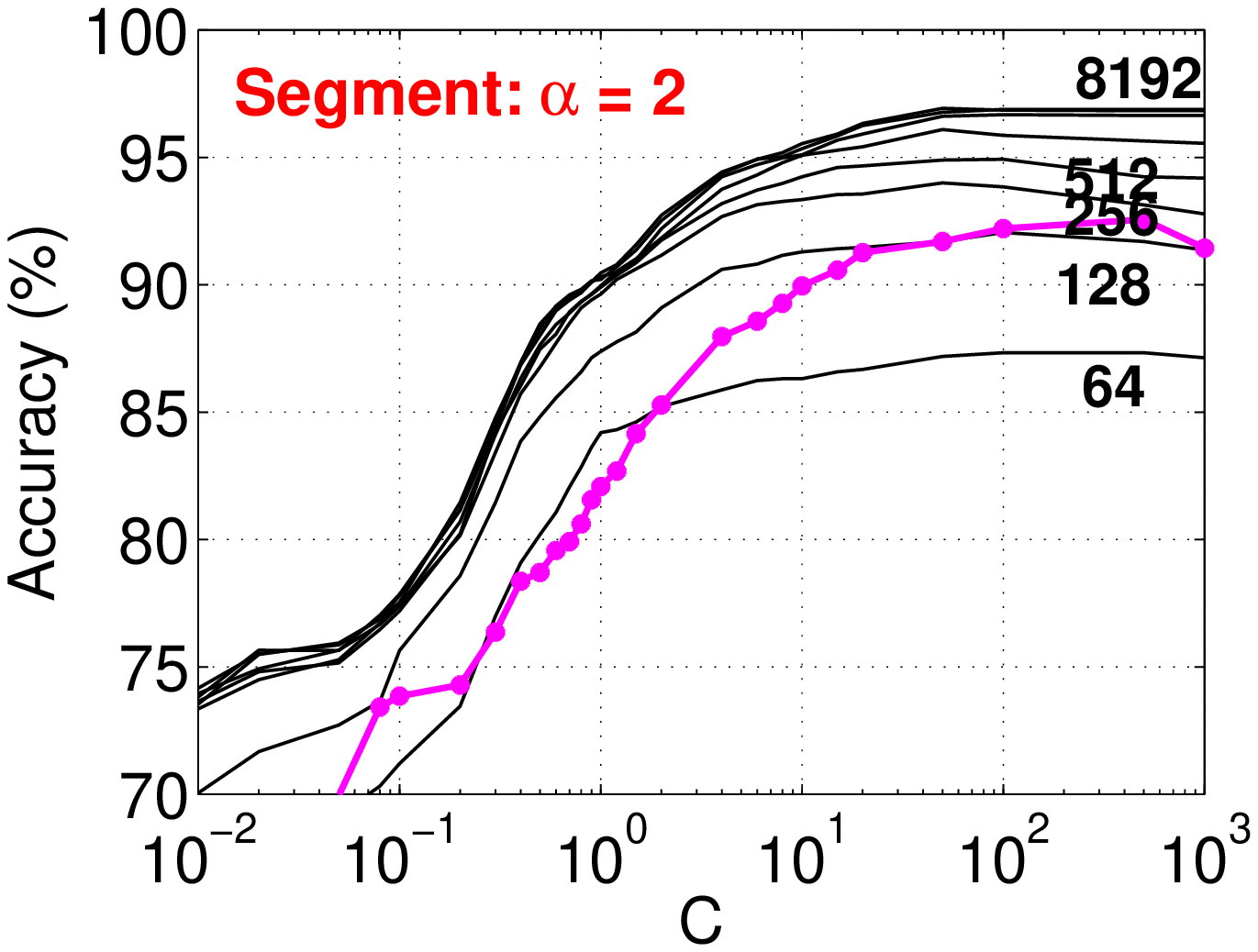}
}

\end{center}
\vspace{-0.2in}
\caption{\textbf{Segment}. Classification accuracies of sign $\alpha$-stable random projections using $l_2$-regularized SVMs (with a tuning parameter $C\in[10^{-2},10^3]$) for $\alpha\in\{0.1,0.25,0.5,0.75,1,1.25,1.5,1.75,2\}$ and $k\in\{64,128,256,512,1024,2048,4096,8192\}$ projections. In each panel, the highest point (i.e., best accuracy) at $k=8192$ was reported in Table~\ref{tab_data}. In addition, each panel also presents the accuracies of linear SVM (the pink curve marked by *). All experiments were conducted by LIBLINEAR. }\label{fig_SegmentSRP}
\end{figure}

Figures~\ref{fig_Covertype10kSRP} to ~\ref{fig_SegmentSRP} presents the detailed classification results of sign $\alpha$-stable random projections for selected 4 datasets, using $l_2$-regularized linear SVM (with a regularization parameter $C\in[10^{-2},10^3]$). In each figure, we present the results for $k\in\{64,128,256,512,1024,2048,4096,8192\}$ projections and $\alpha\in\{0.1,0.25,0.5,0.75,1,1.25,1.5,1.75,2\}$. All experiments were conducted using LIBLINEAR~\cite{Article:Fan_JMLR08} and we repeated each randomized experiment 5 times and reported the average results. The classification results are very stable (i.e., very small variance) unless $k$ is too small.

The results (together with  Table~\ref{tab_data} and other figures later in the paper) show that, given enough projections (e.g., $8192$), the method of sign $\alpha$-stable random projections can typically achieve good accuracies.


\newpage\clearpage

\subsection{Detailed Comparisons with 0-Bit Consistent Weighted Sampling (CWS)}

\vspace{-0.05in}

\begin{figure}[h!]
\begin{center}

\mbox{
\includegraphics[width=2.3in]{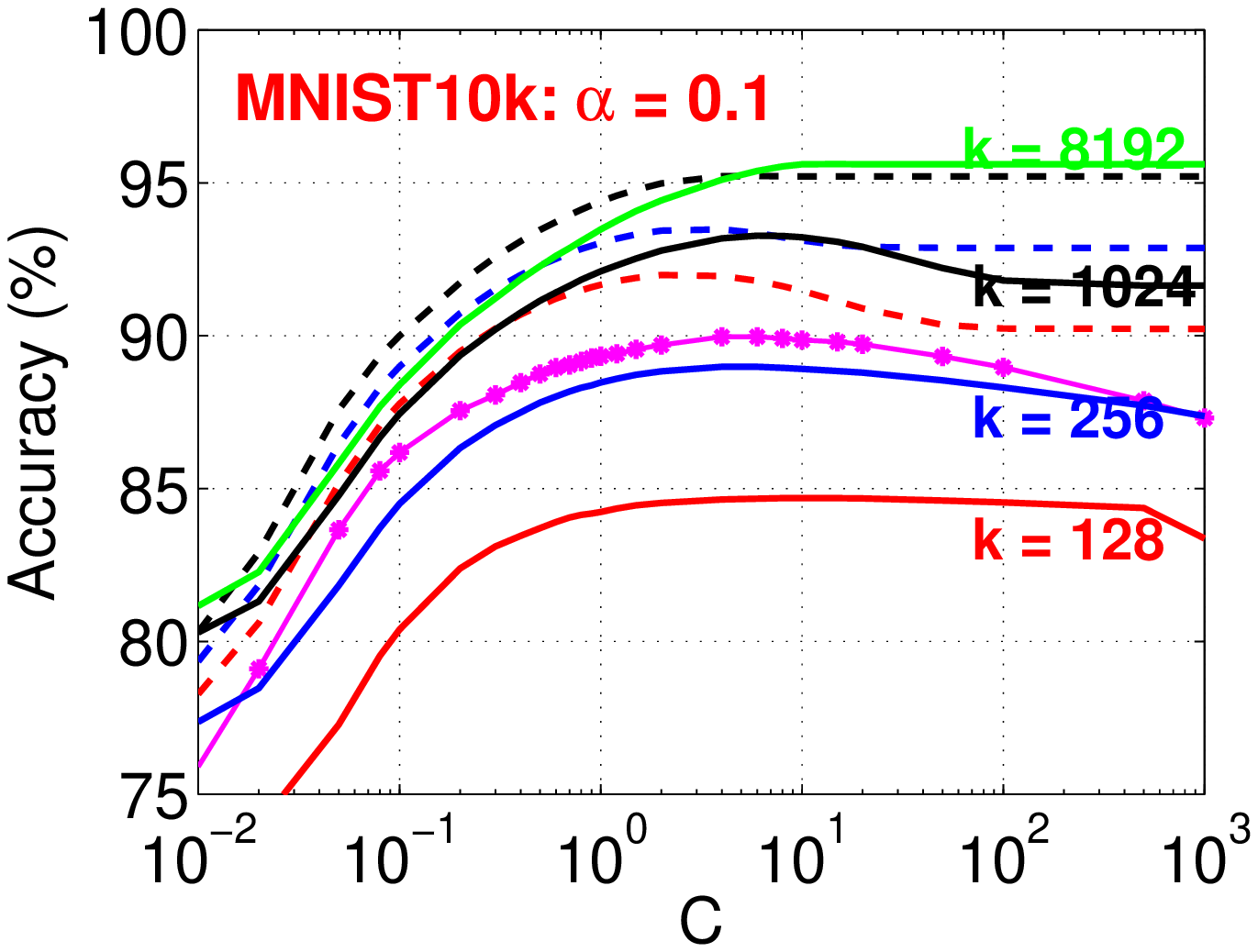}\hspace{-0.15in}
\includegraphics[width=2.3in]{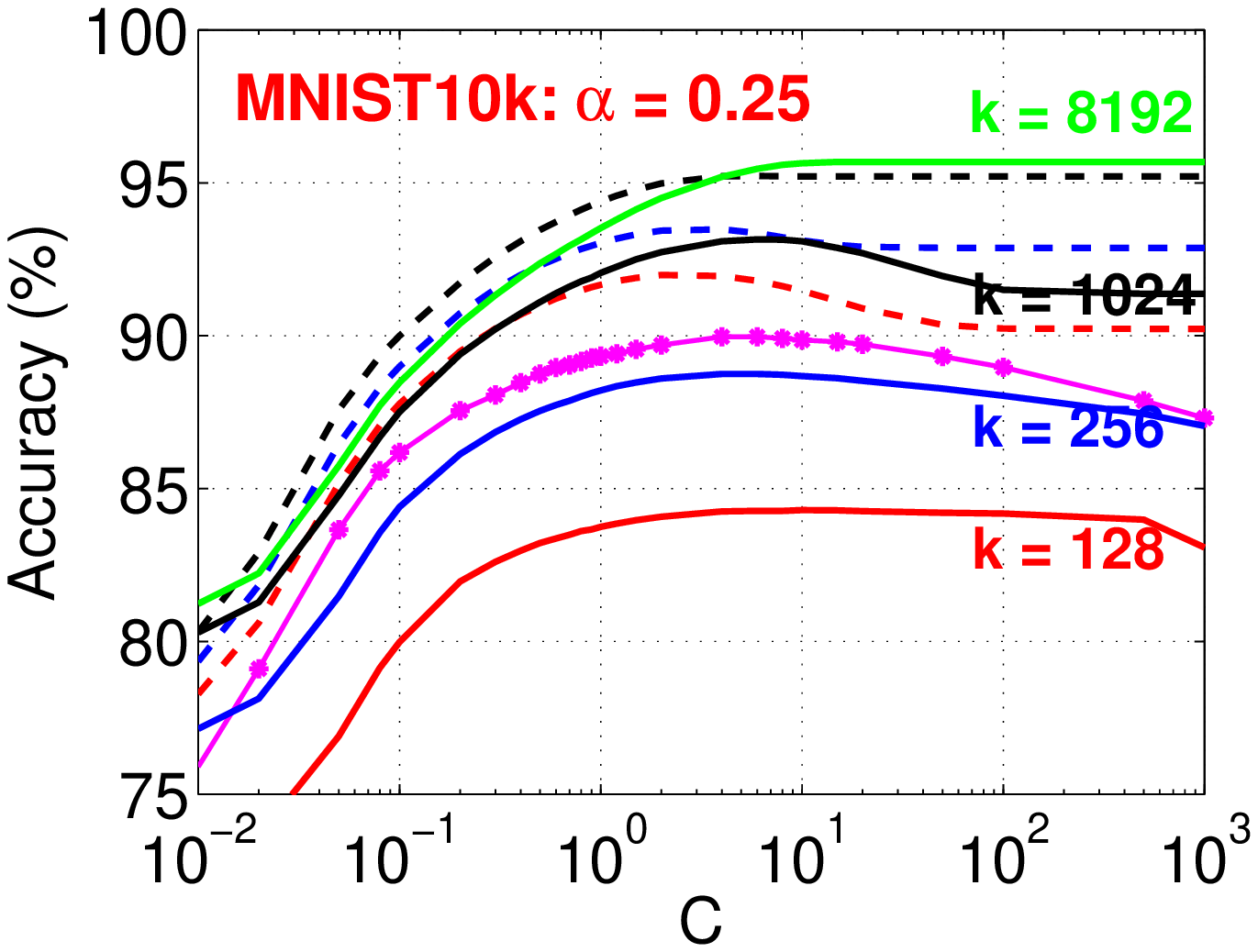}\hspace{-0.15in}
\includegraphics[width=2.3in]{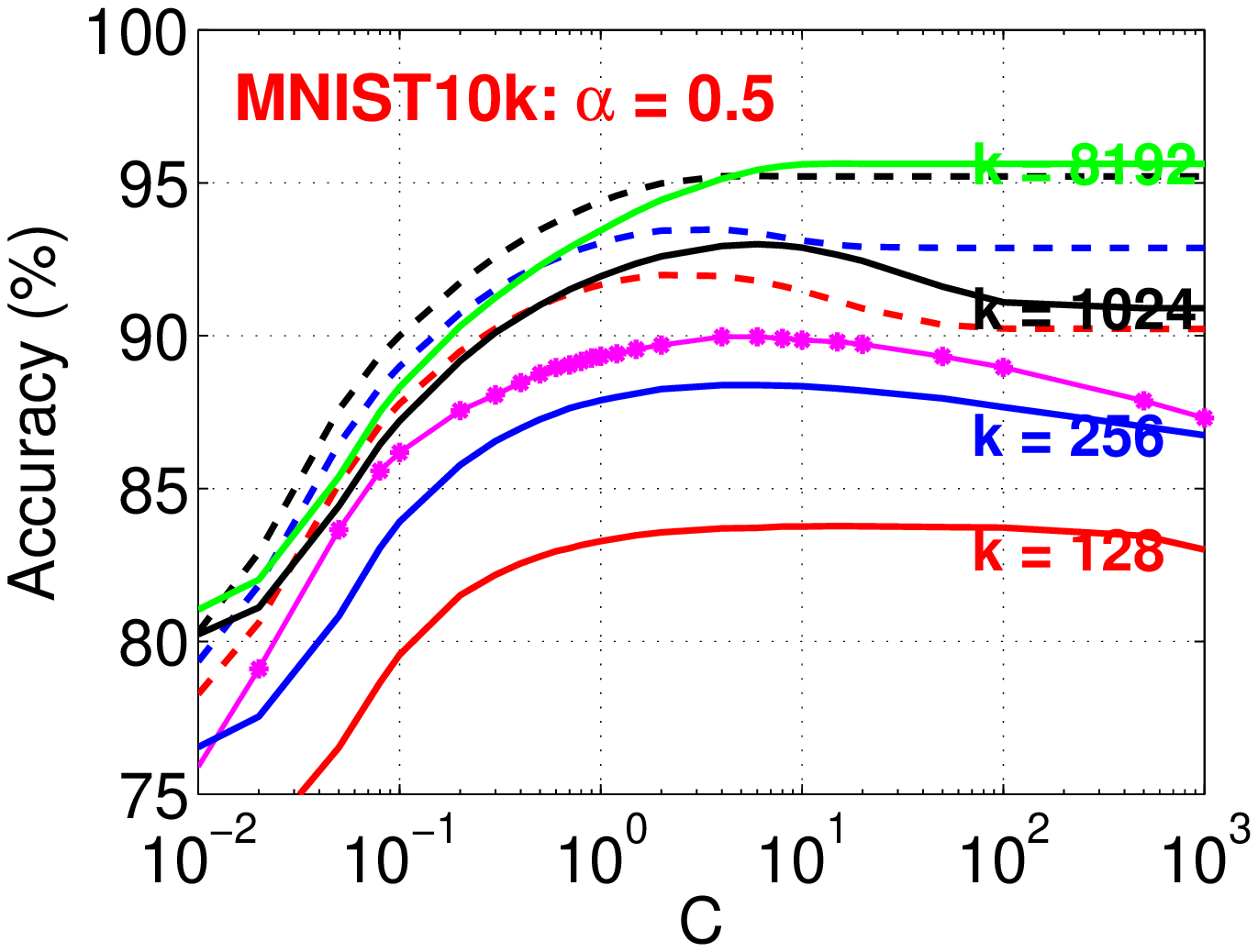}
}

\vspace{-0.15in}

\mbox{
\includegraphics[width=2.3in]{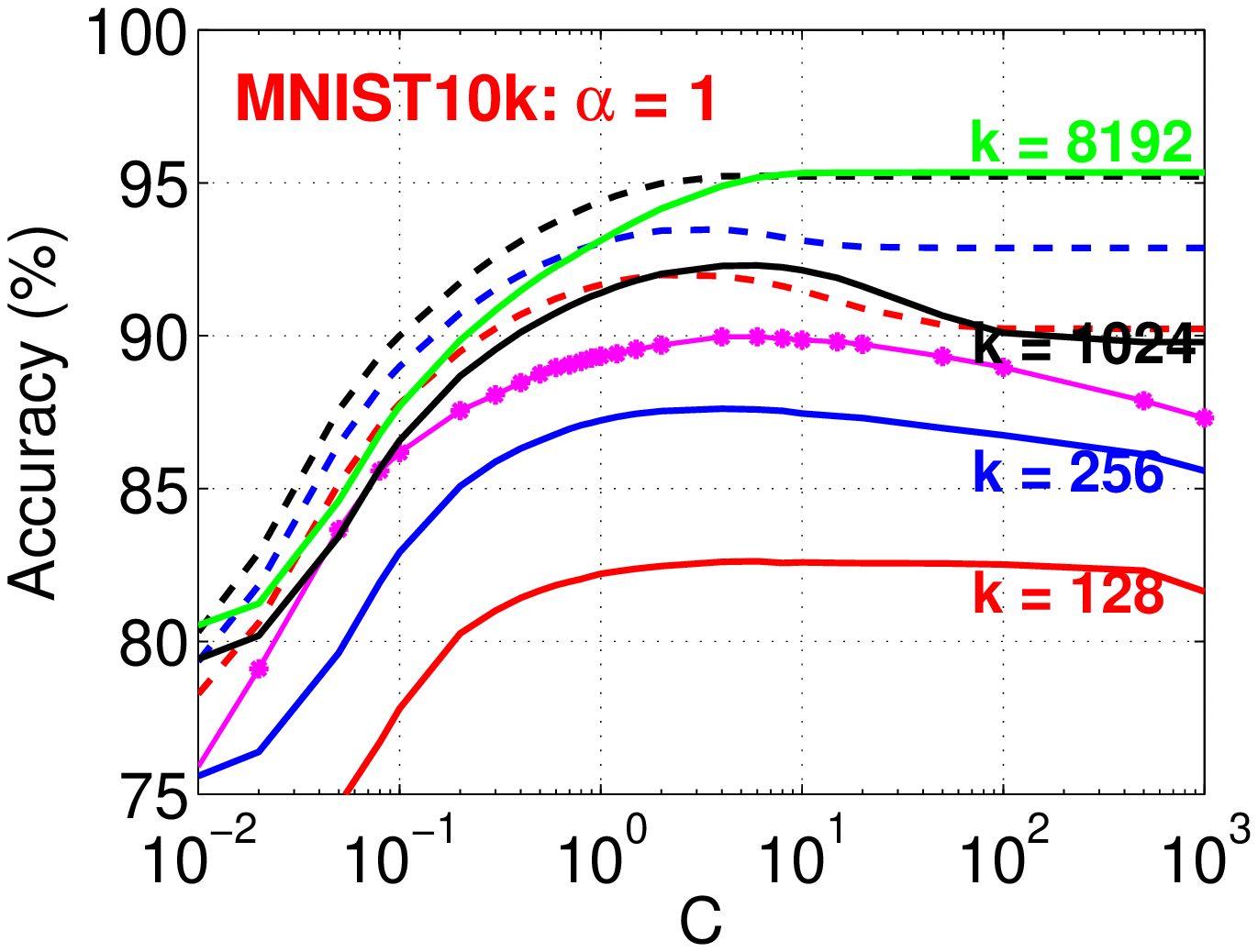}\hspace{-0.15in}
\includegraphics[width=2.3in]{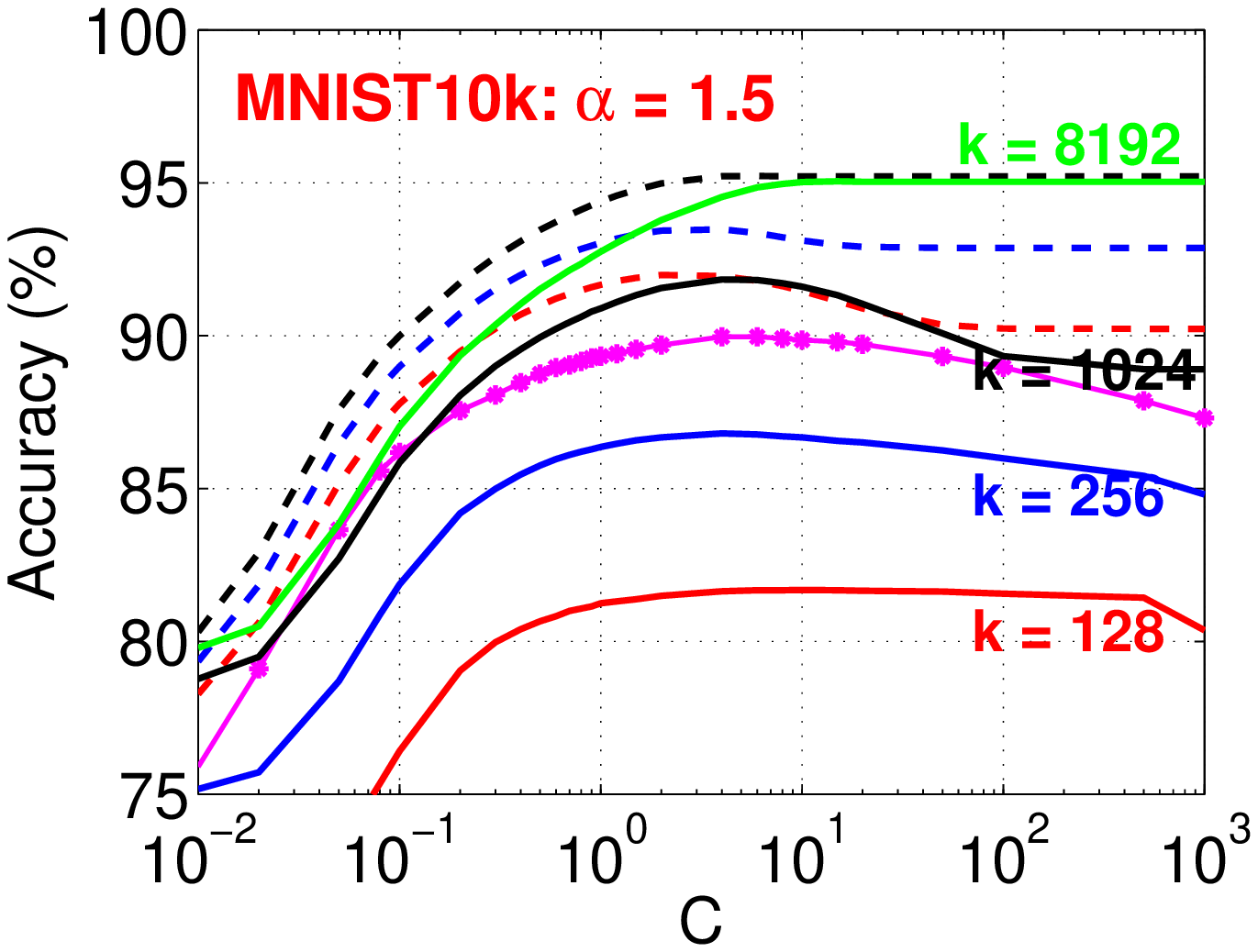}\hspace{-0.15in}
\includegraphics[width=2.3in]{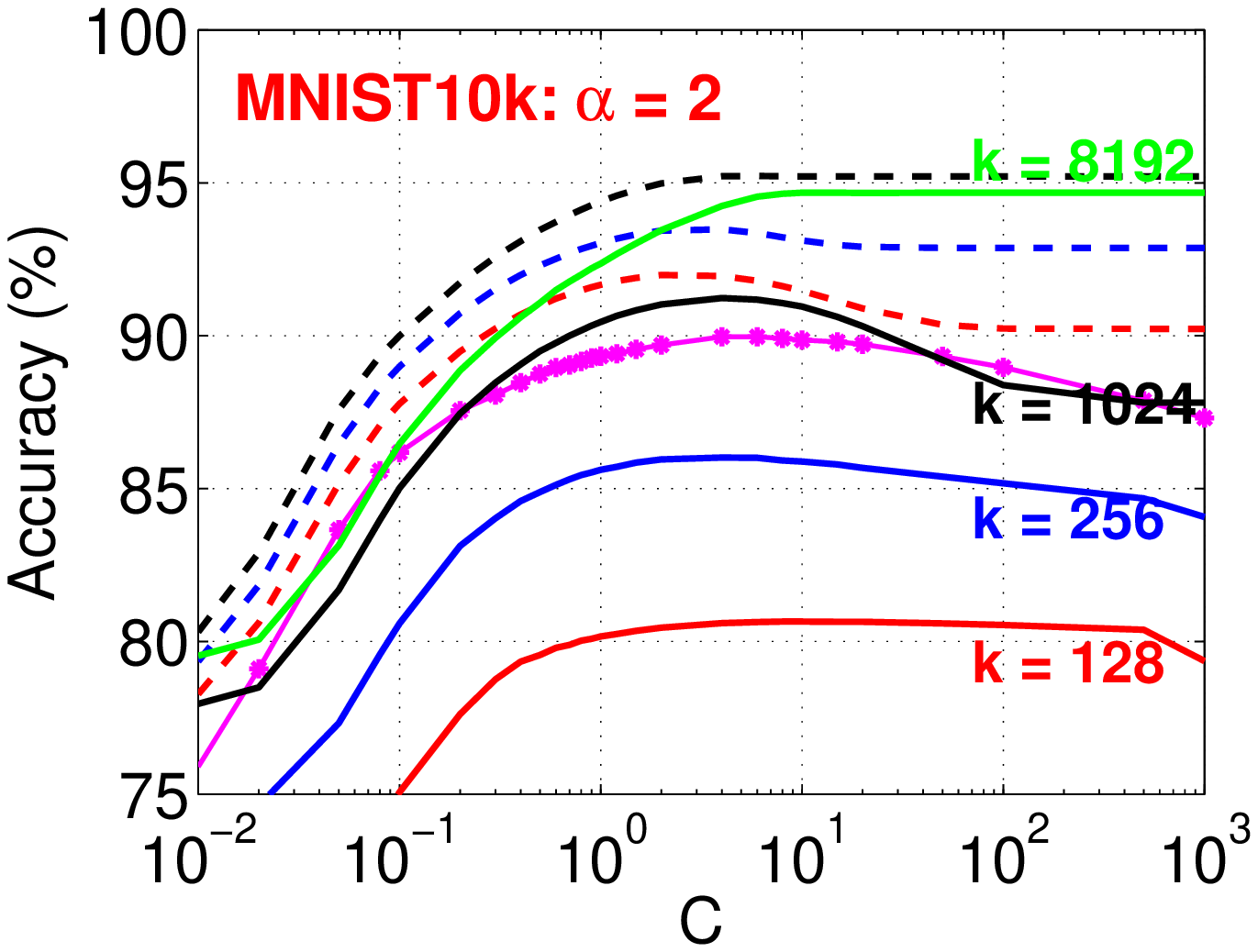}
}

\vspace{0.0in}

\mbox{
\includegraphics[width=2.3in]{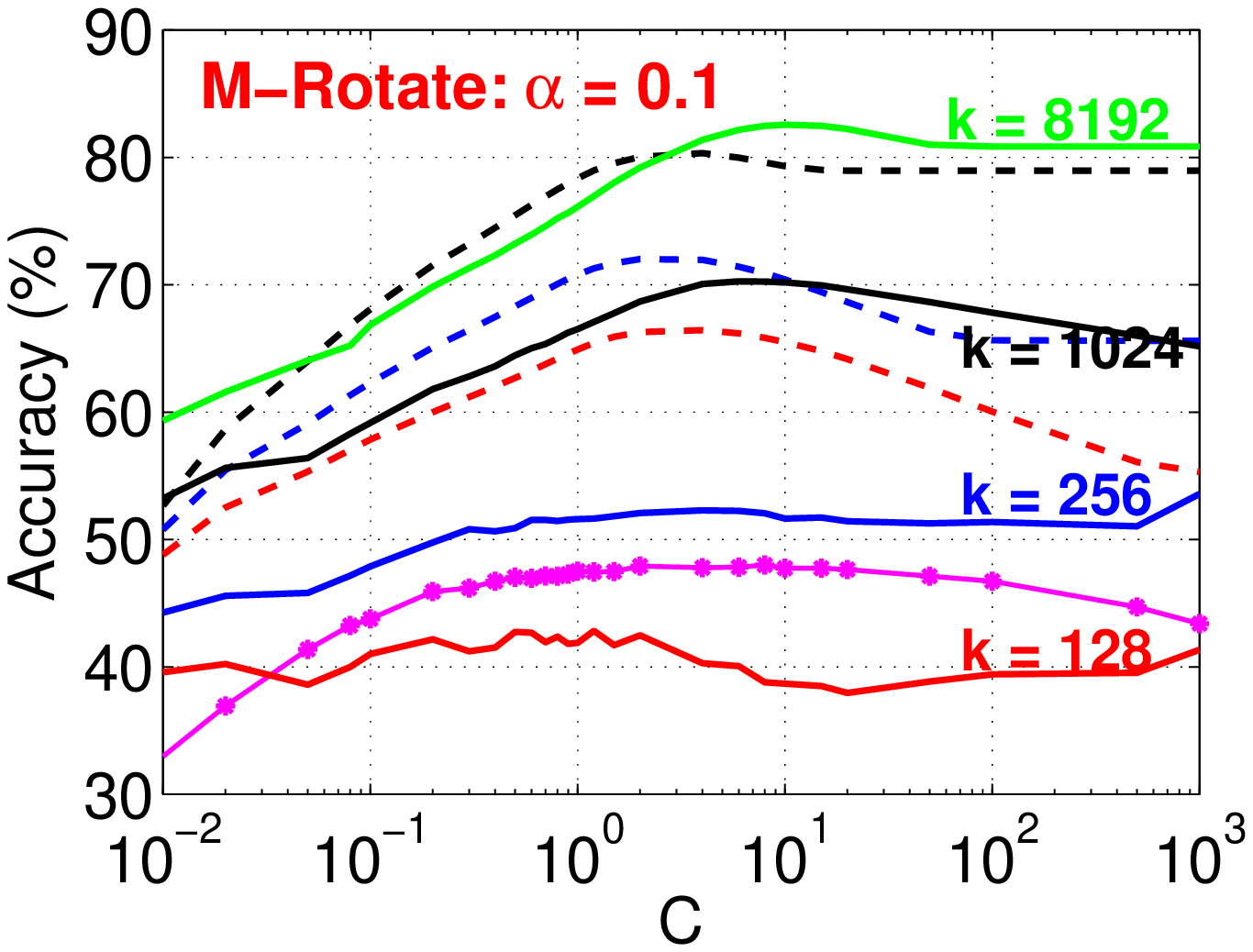}\hspace{-0.15in}
\includegraphics[width=2.3in]{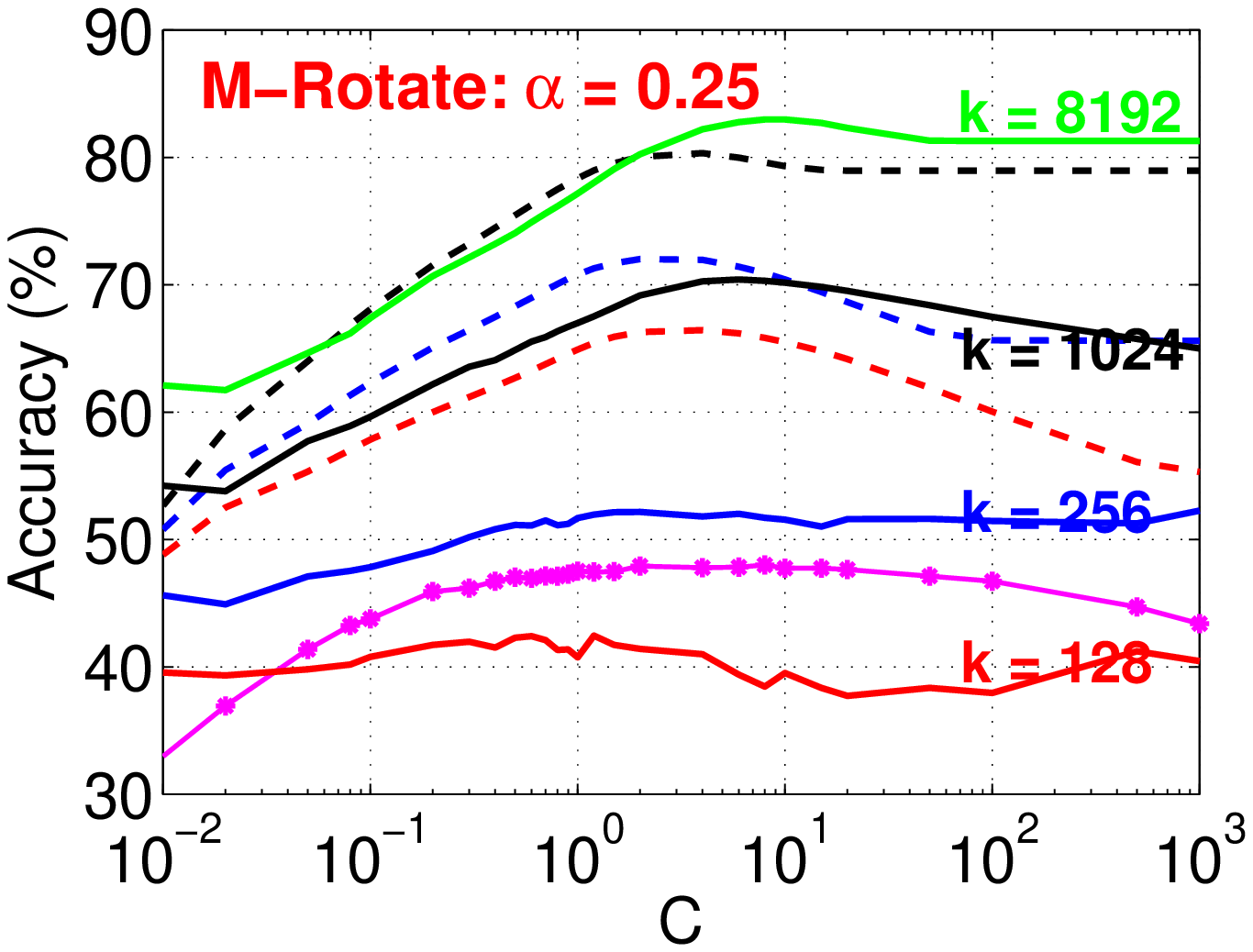}\hspace{-0.15in}
\includegraphics[width=2.3in]{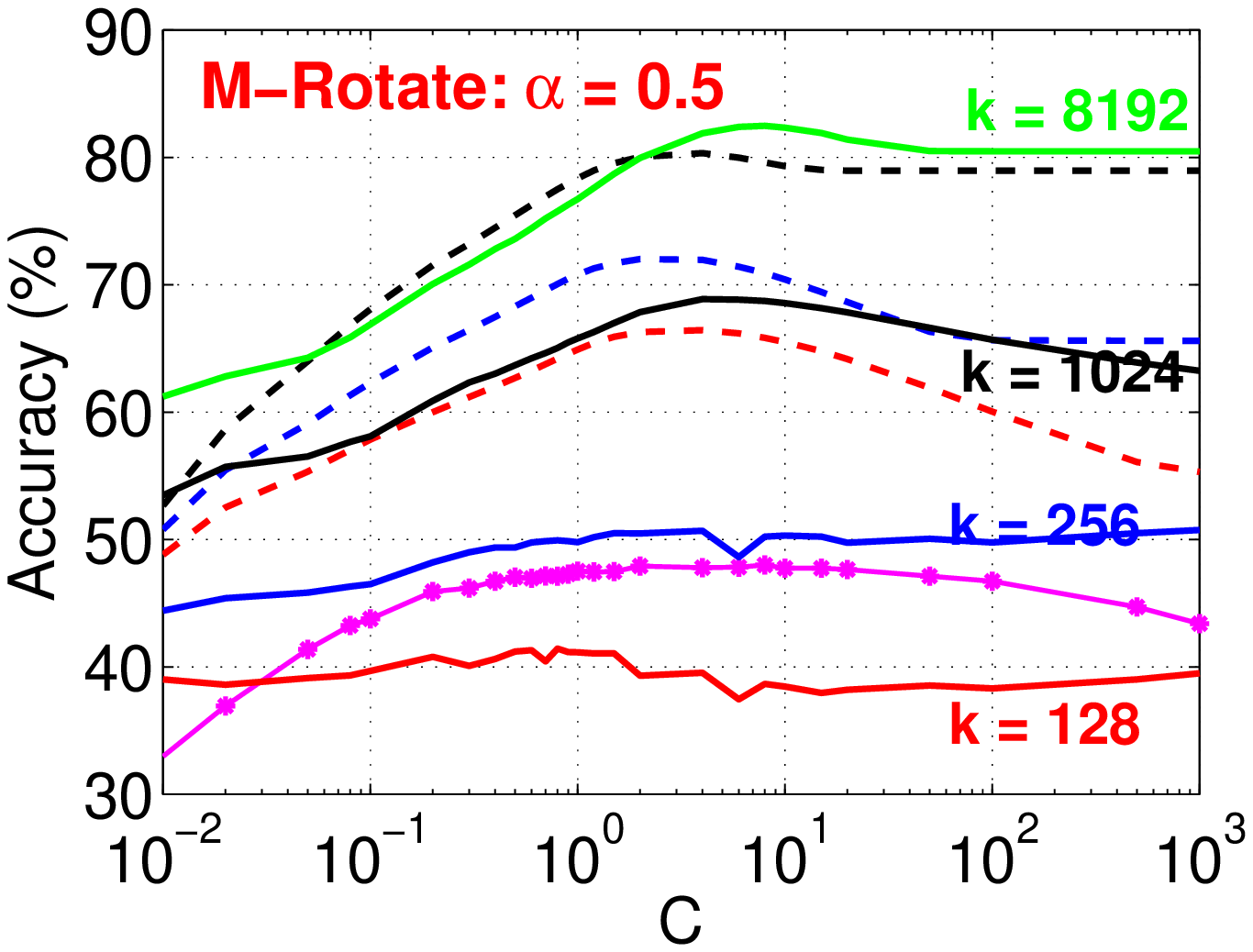}
}

\vspace{-0.15in}

\mbox{
\includegraphics[width=2.3in]{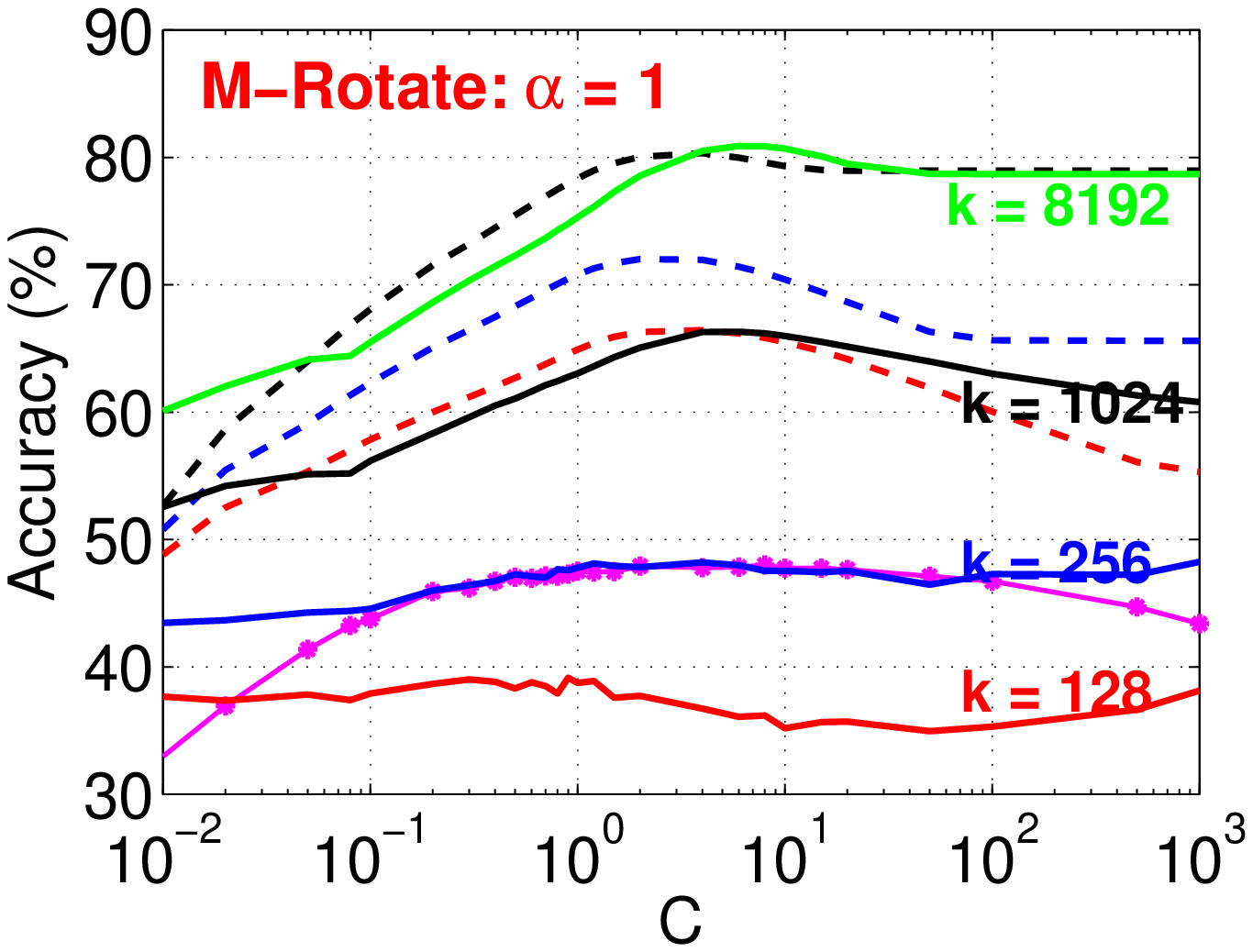}\hspace{-0.15in}
\includegraphics[width=2.3in]{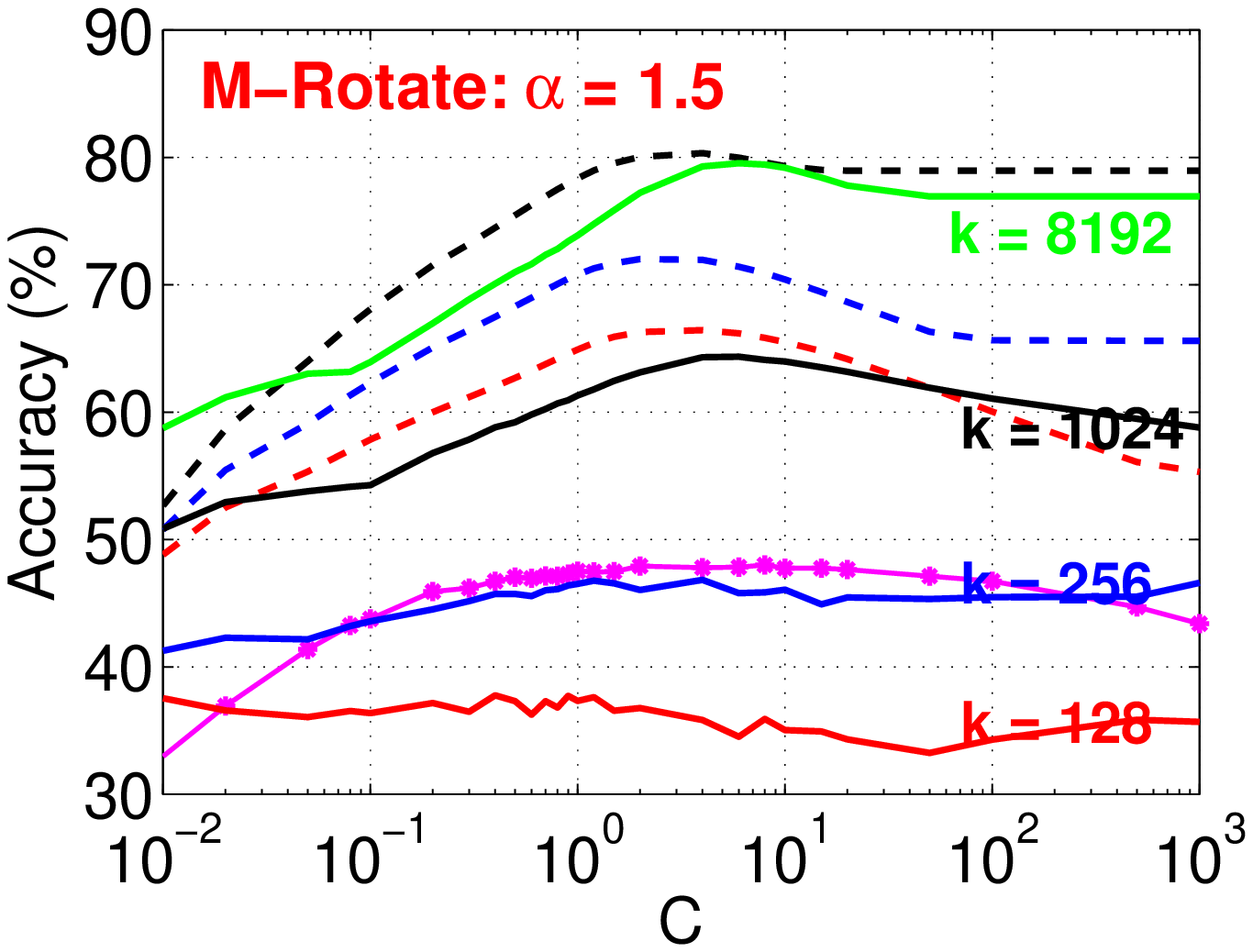}\hspace{-0.15in}
\includegraphics[width=2.3in]{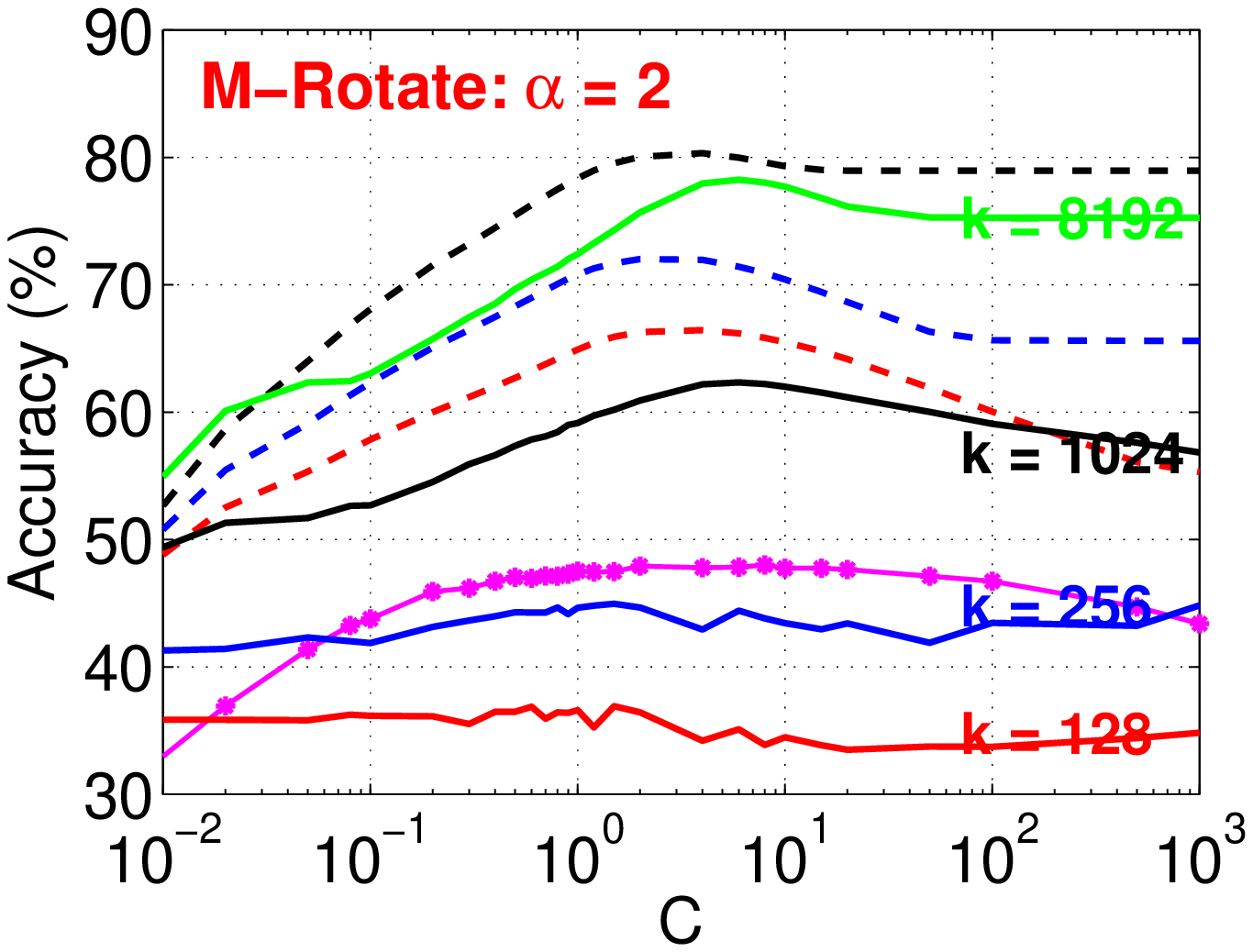}
}

\end{center}
\vspace{-0.35in}
\caption{\textbf{MNIST10k} (top 2 rows) and {\bf M-Rotate} (bottom 2 rows). We compare sign $\alpha$-stable random projections with 0-bit consistent weighted sampling (CWS). Each panel (for each $\alpha$) consists of 8 curves. The solid (pink) curve marked by * represents the results of linear SVM. Four solid curves (labelled by $k=128$, $k=256$, $k=1024$, and $k=8192$, respectively) represent the results of sign $\alpha$-stable random projections for 4 different $k$ values.  The 3 dashed curves  correspond to the results of 0-bit CWS for $k=128, 256, 1024$ (a higher curve for a higher $k$ value). These experimental results, all conducted using LIBLINEAR, show that 0-bit CWS requires much fewer samples to achieve the sample accuracies.   }\label{fig_CWS1}
\end{figure}

\begin{figure}[h!]
\begin{center}

\mbox{
\includegraphics[width=2.3in]{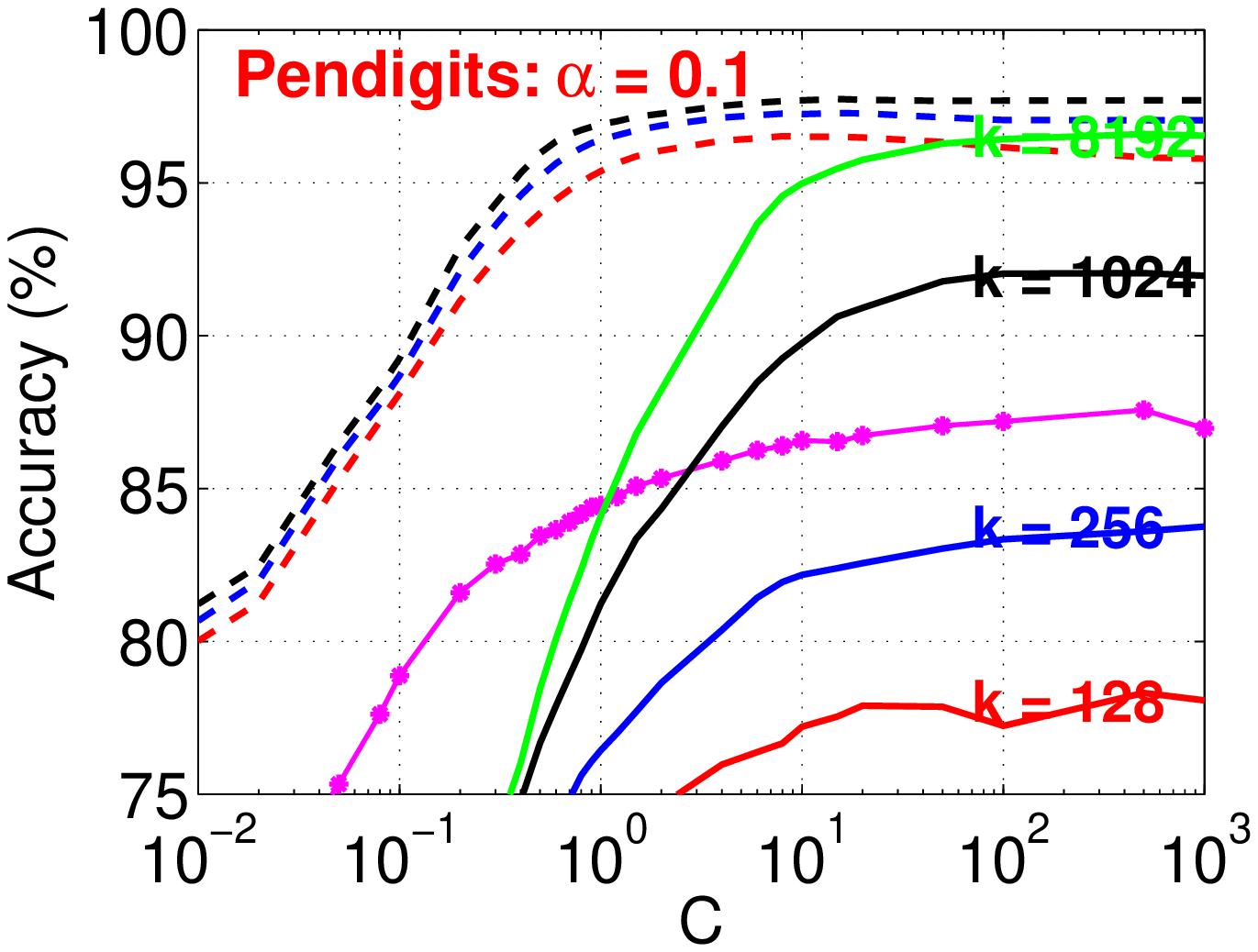}\hspace{-0.15in}
\includegraphics[width=2.3in]{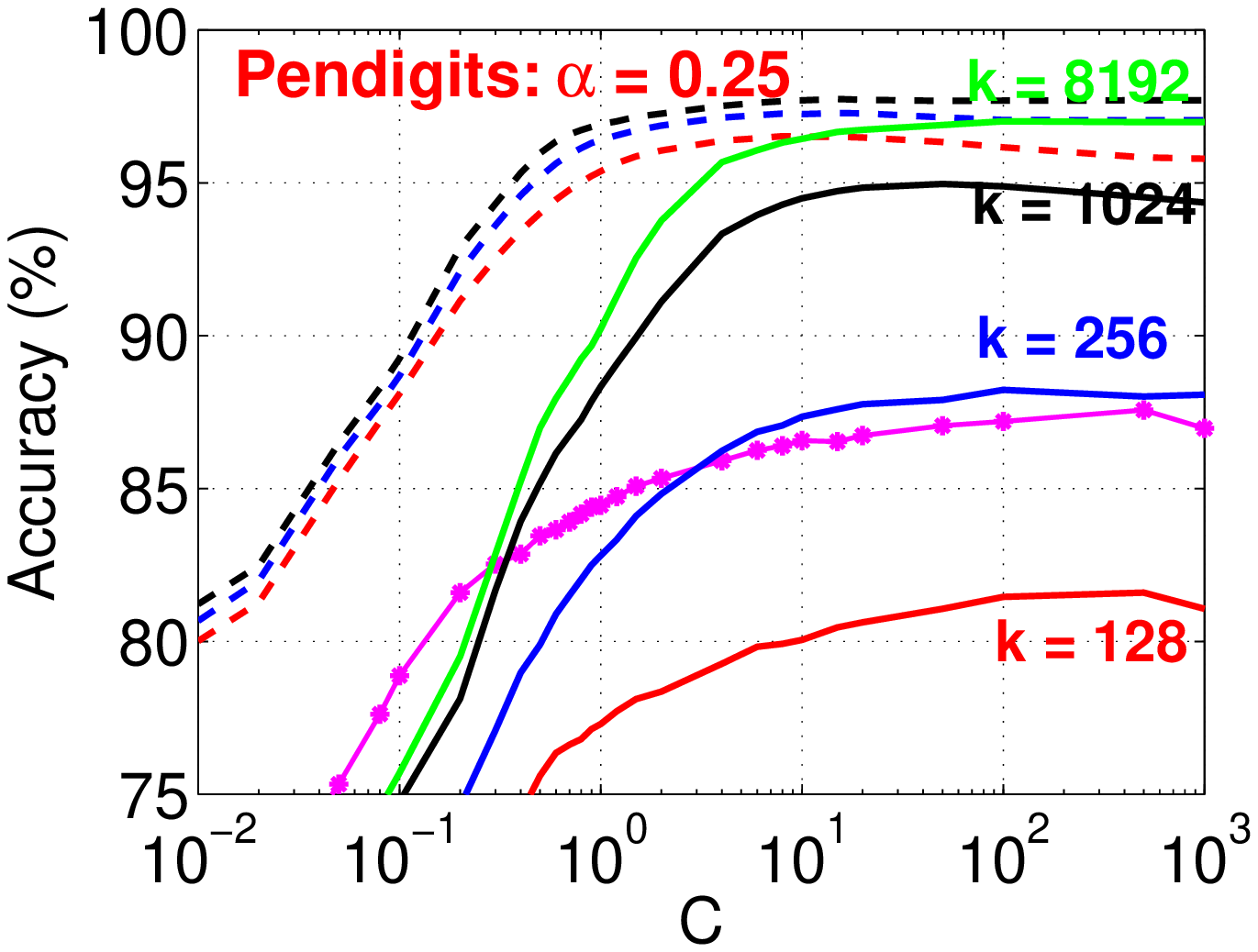}\hspace{-0.15in}
\includegraphics[width=2.3in]{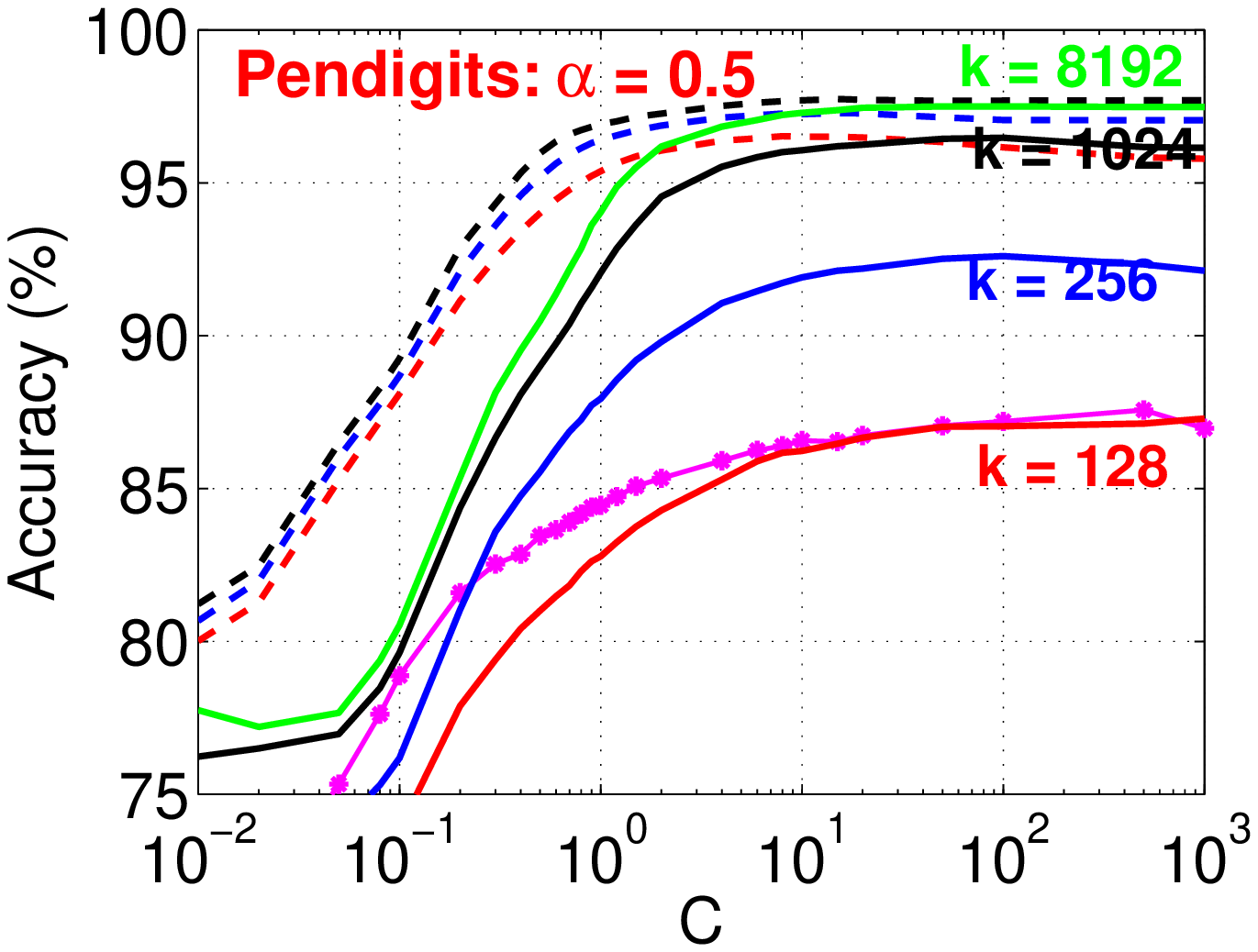}
}
\mbox{
\includegraphics[width=2.3in]{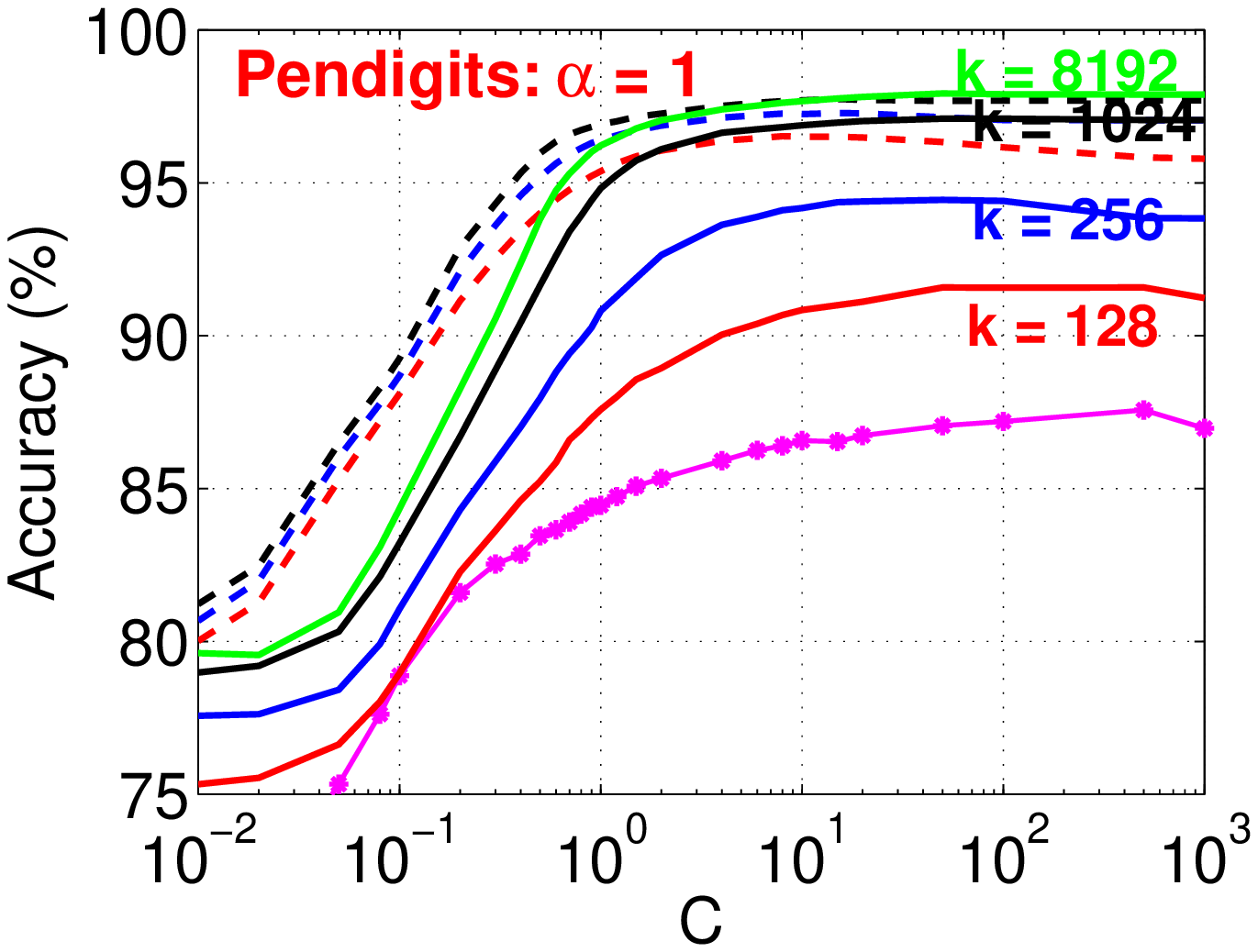}\hspace{-0.15in}
\includegraphics[width=2.3in]{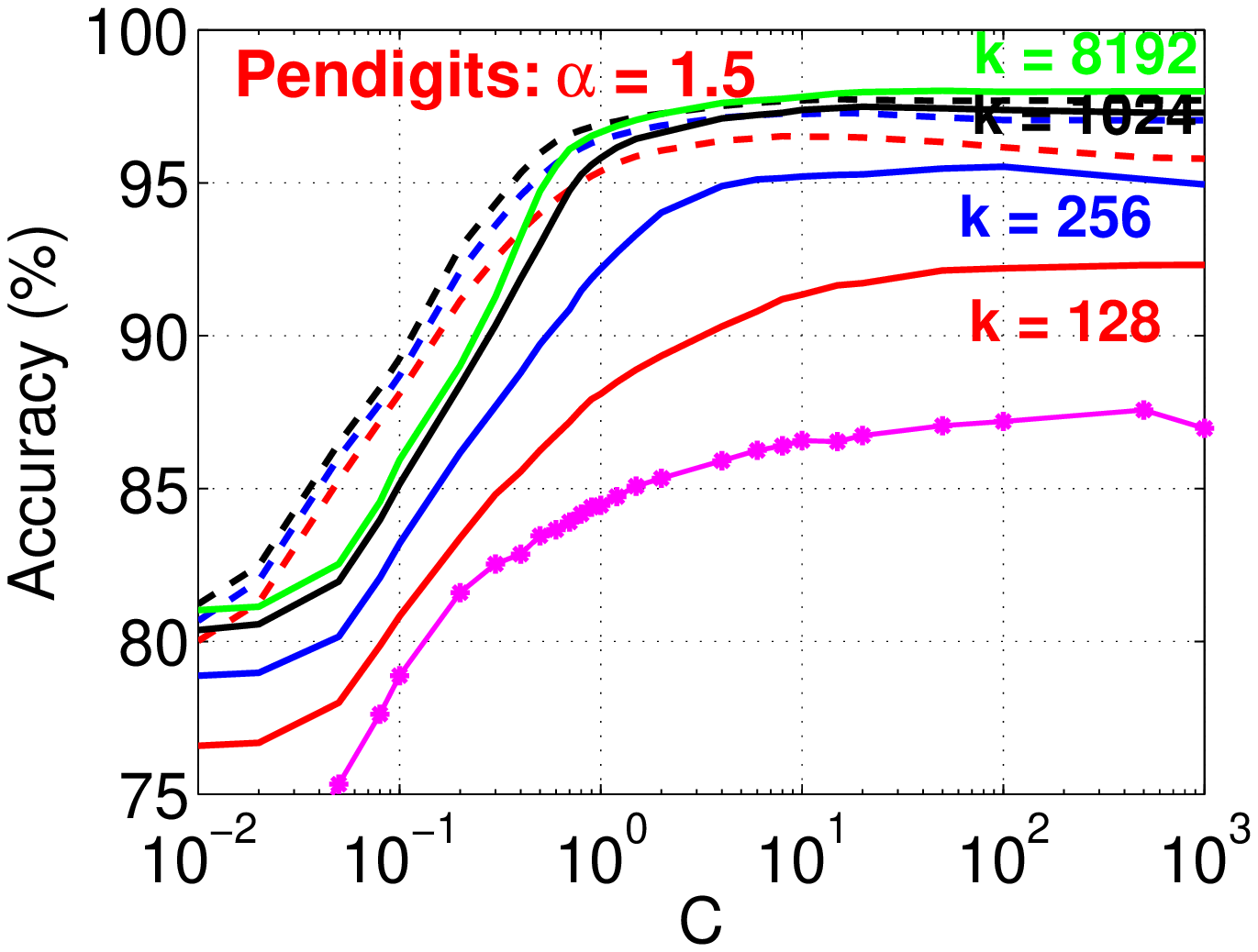}\hspace{-0.15in}
\includegraphics[width=2.3in]{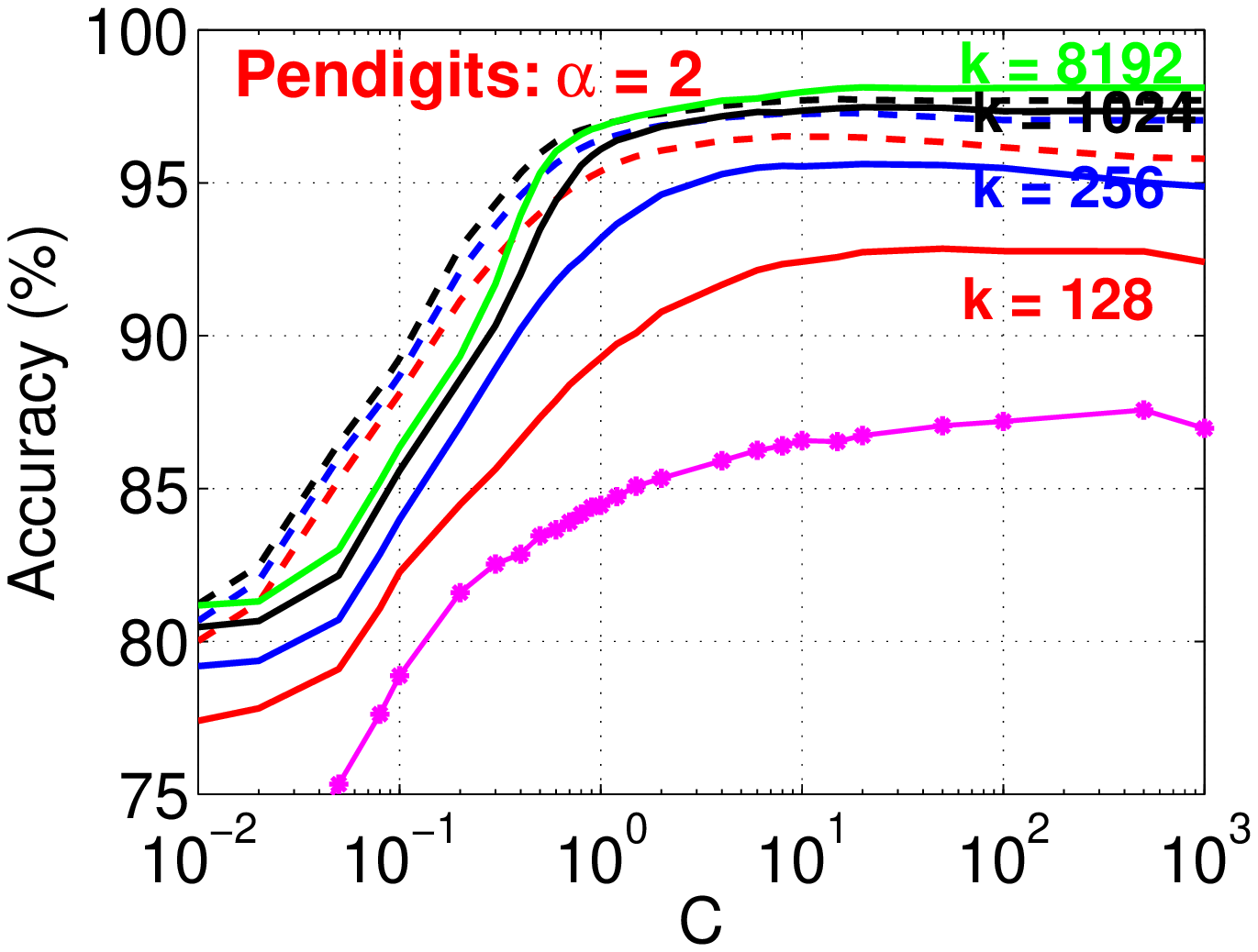}
}

\vspace{0.3in}

\mbox{
\includegraphics[width=2.3in]{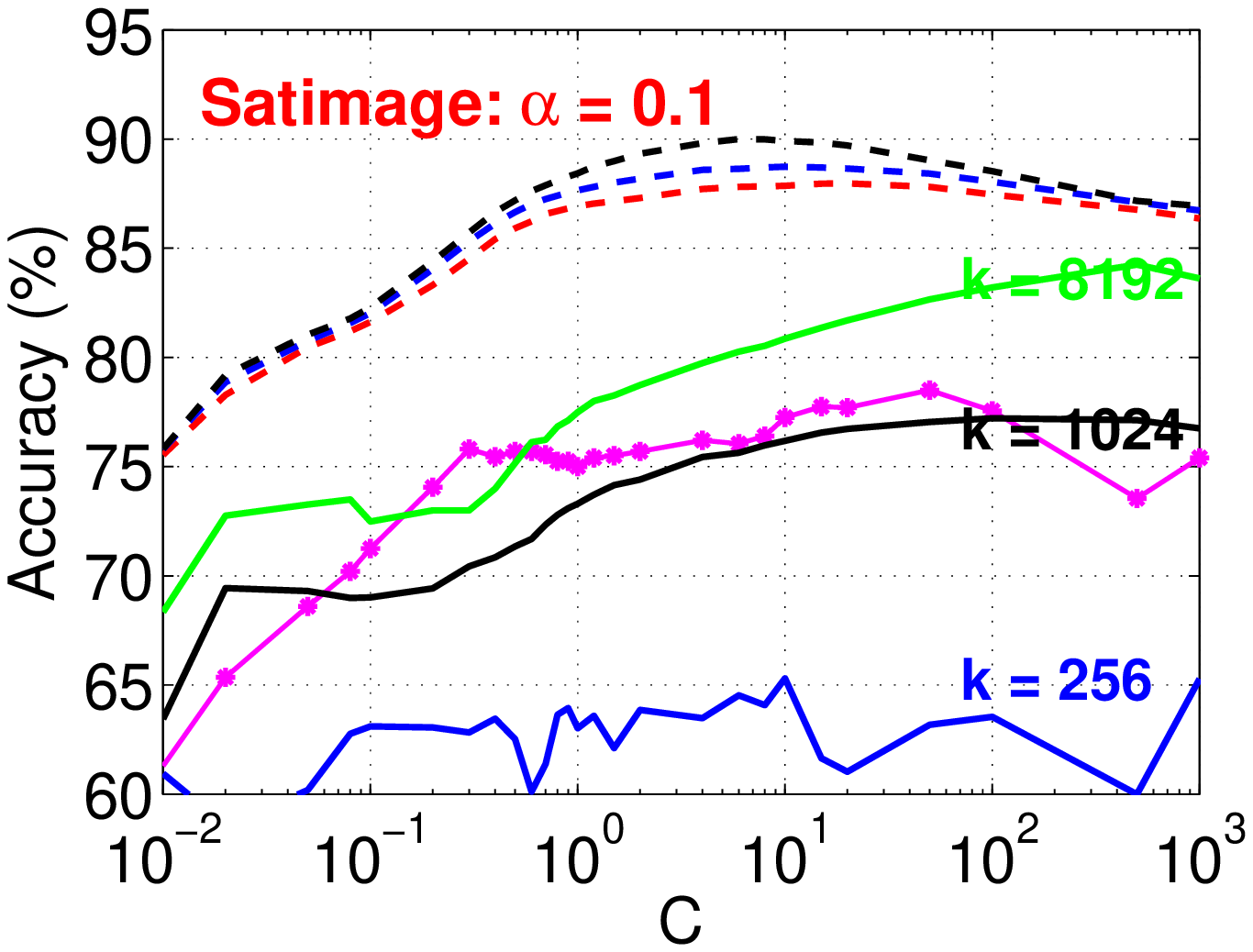}\hspace{-0.15in}
\includegraphics[width=2.3in]{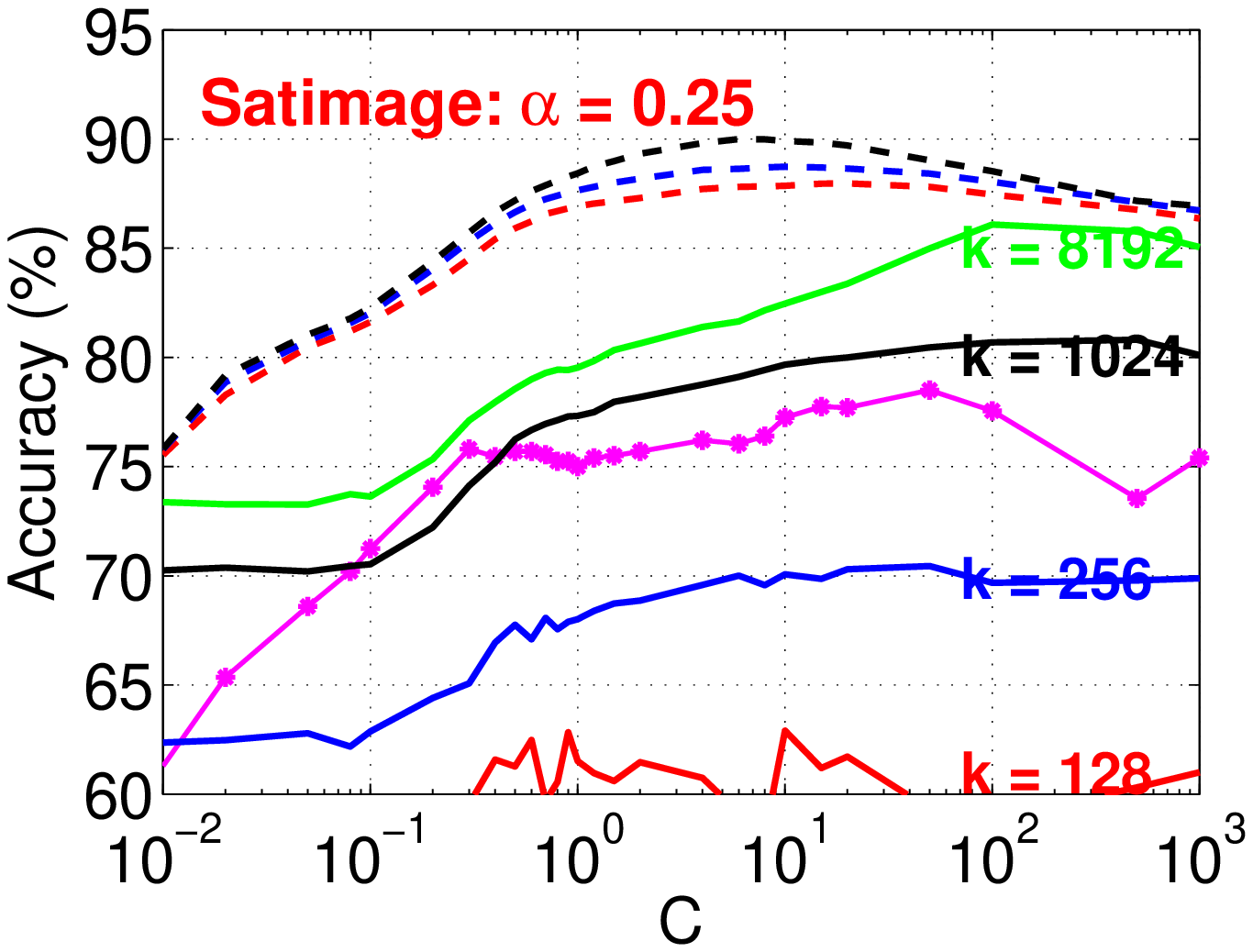}\hspace{-0.15in}
\includegraphics[width=2.3in]{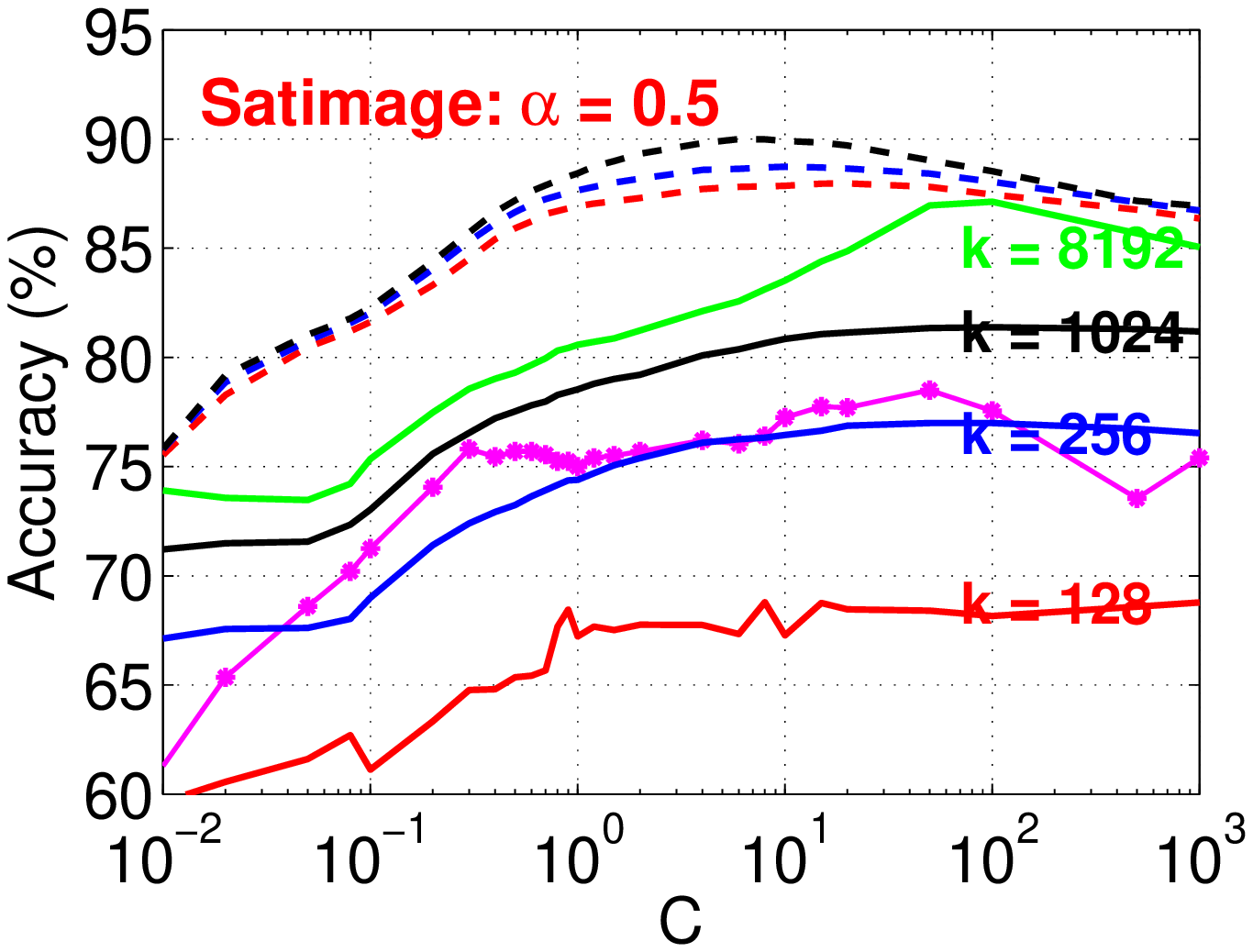}
}
\mbox{
\includegraphics[width=2.3in]{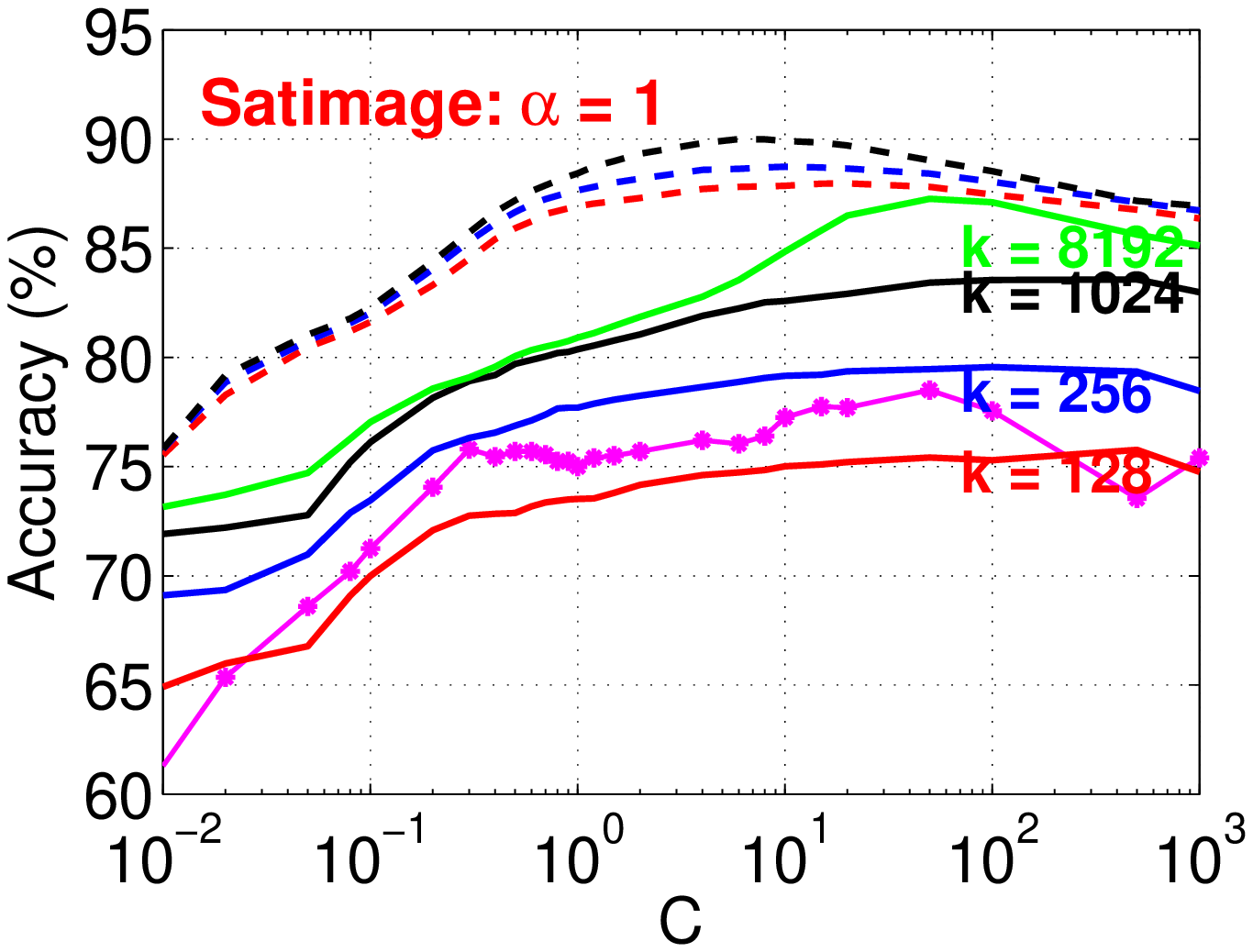}\hspace{-0.15in}
\includegraphics[width=2.3in]{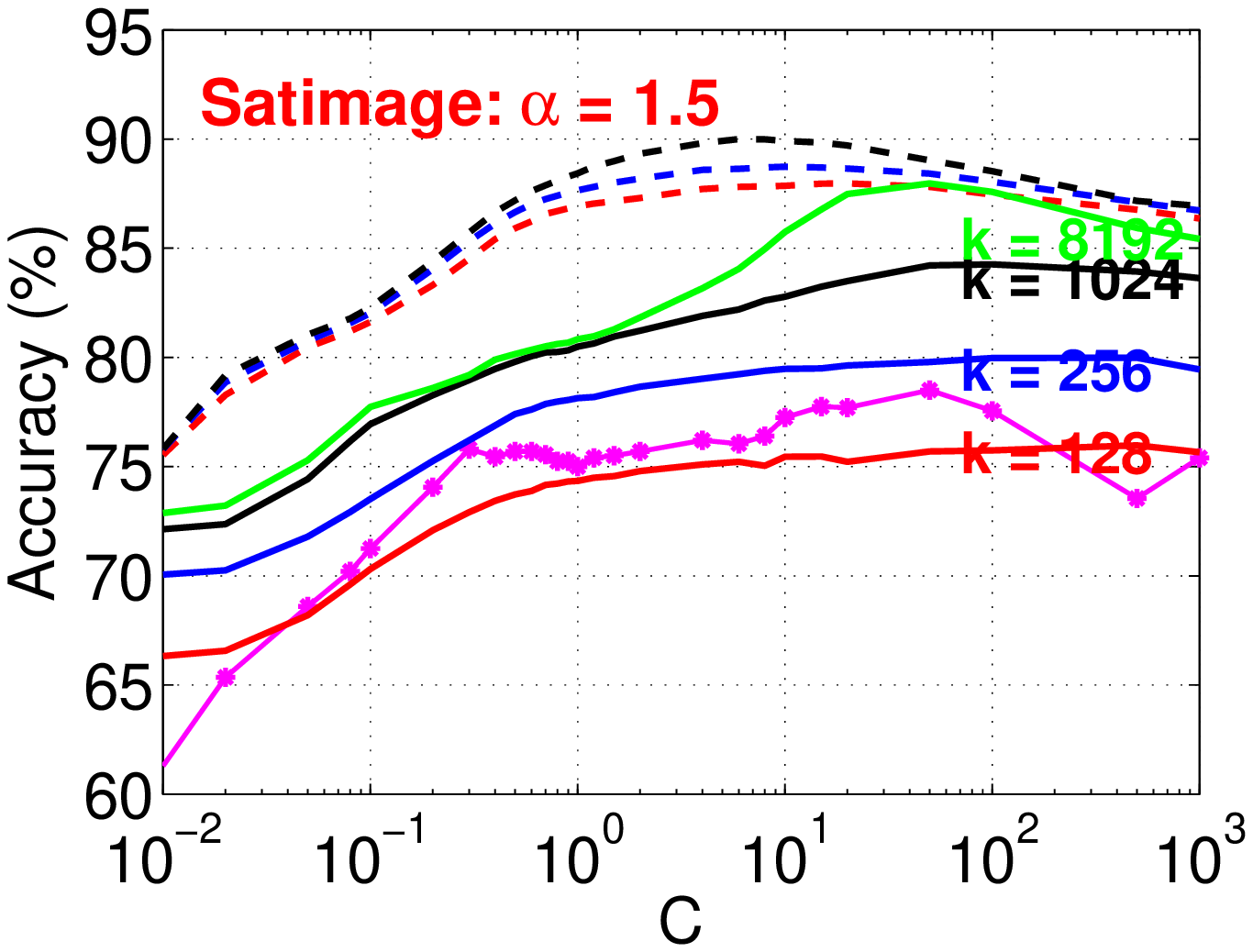}\hspace{-0.15in}
\includegraphics[width=2.3in]{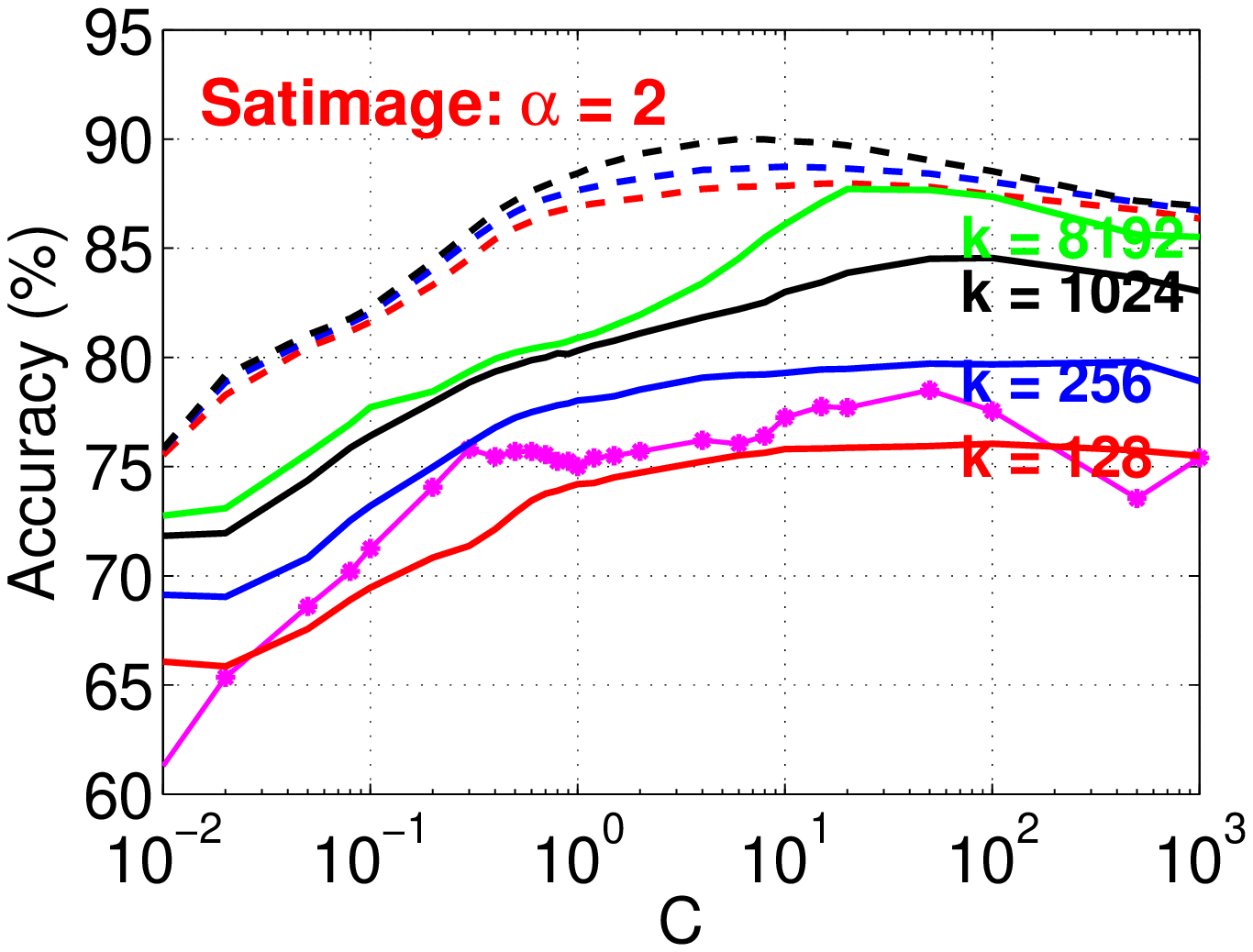}
}

\end{center}
\vspace{-0.3in}
\caption{\textbf{Pendigits} and {\bf Satimage}. We compare sign $\alpha$-stable random projections with 0-bit consistent weighted sampling (CWS). Each panel (for each $\alpha$) consists of 8 curves. The solid (pink) curve marked by * represents the results of linear SVM. Four solid curves (labelled by $k=128$, $k=256$, $k=1024$, and $k=8192$, respectively) represent the results of sign $\alpha$-stable random projections for 4 different $k$ values.  The 3 dashed curves  correspond to the results of 0-bit CWS for $k=128, 256, 1024$ (a higher curve for a higher $k$ value). These experimental results, all conducted using LIBLINEAR, show that 0-bit CWS requires much fewer samples to achieve the sample accuracies.    }\label{fig_CWS2}
\end{figure}

\begin{figure}[h!]
\begin{center}

\mbox{
\includegraphics[width=2.3in]{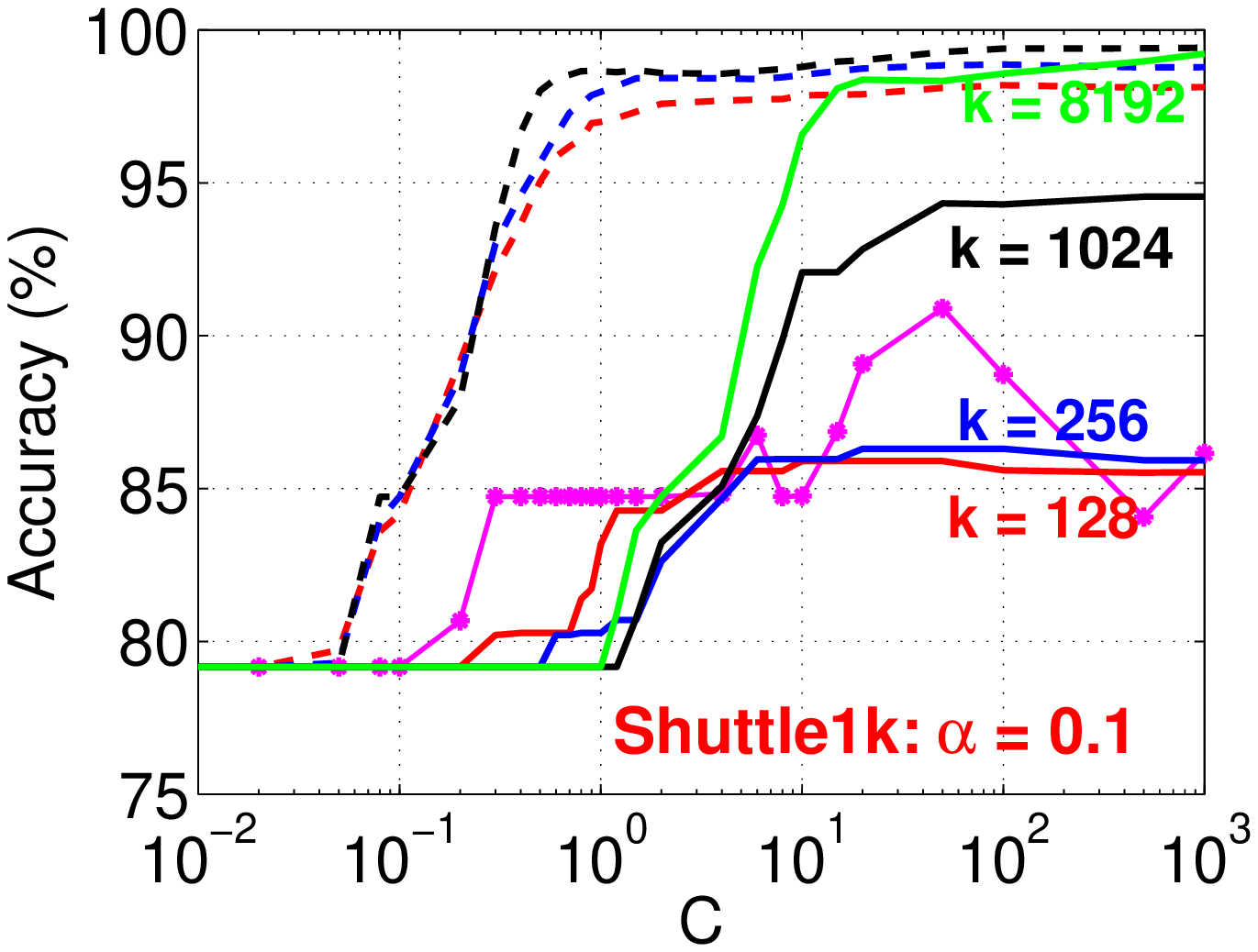}\hspace{-0.15in}
\includegraphics[width=2.3in]{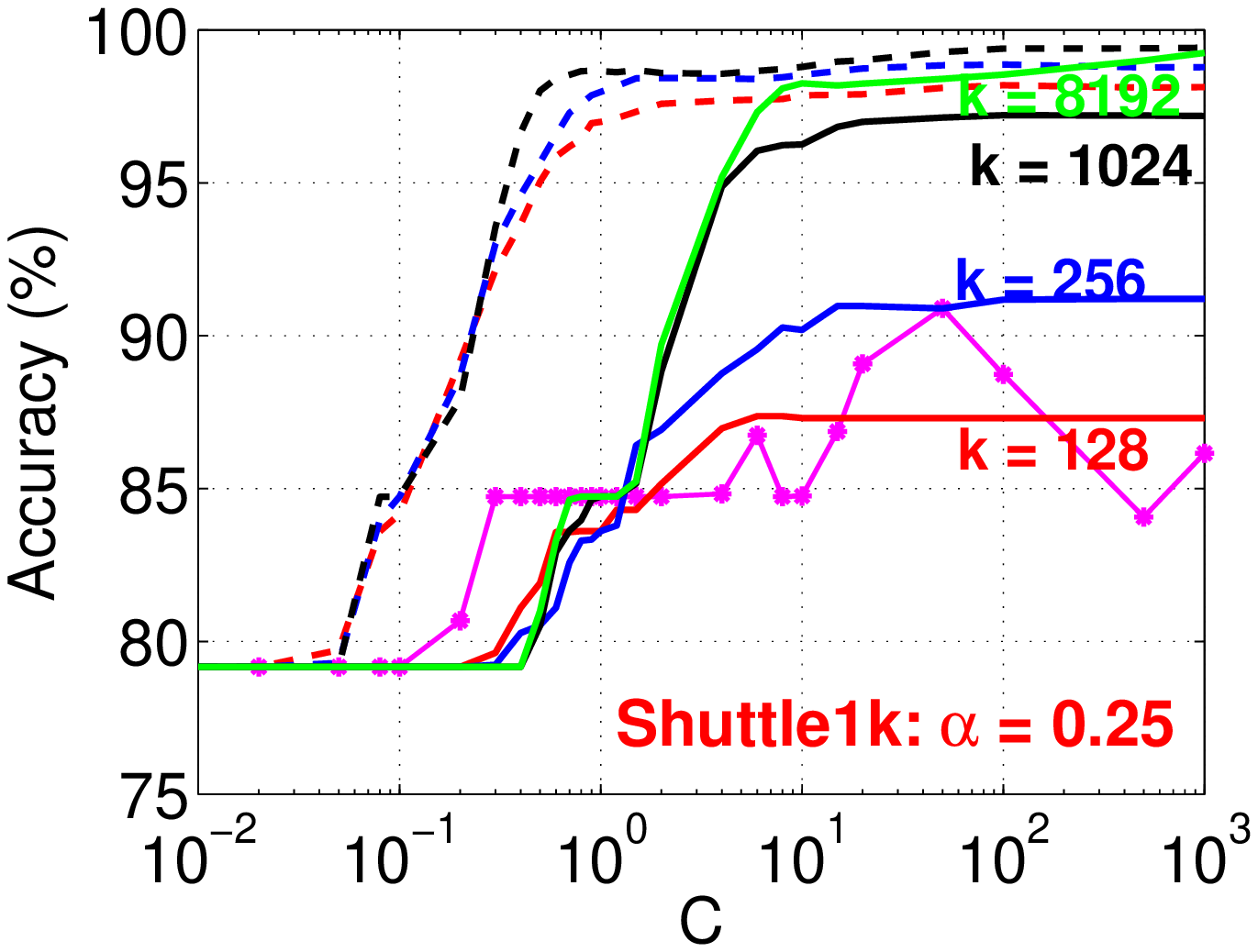}\hspace{-0.15in}
\includegraphics[width=2.3in]{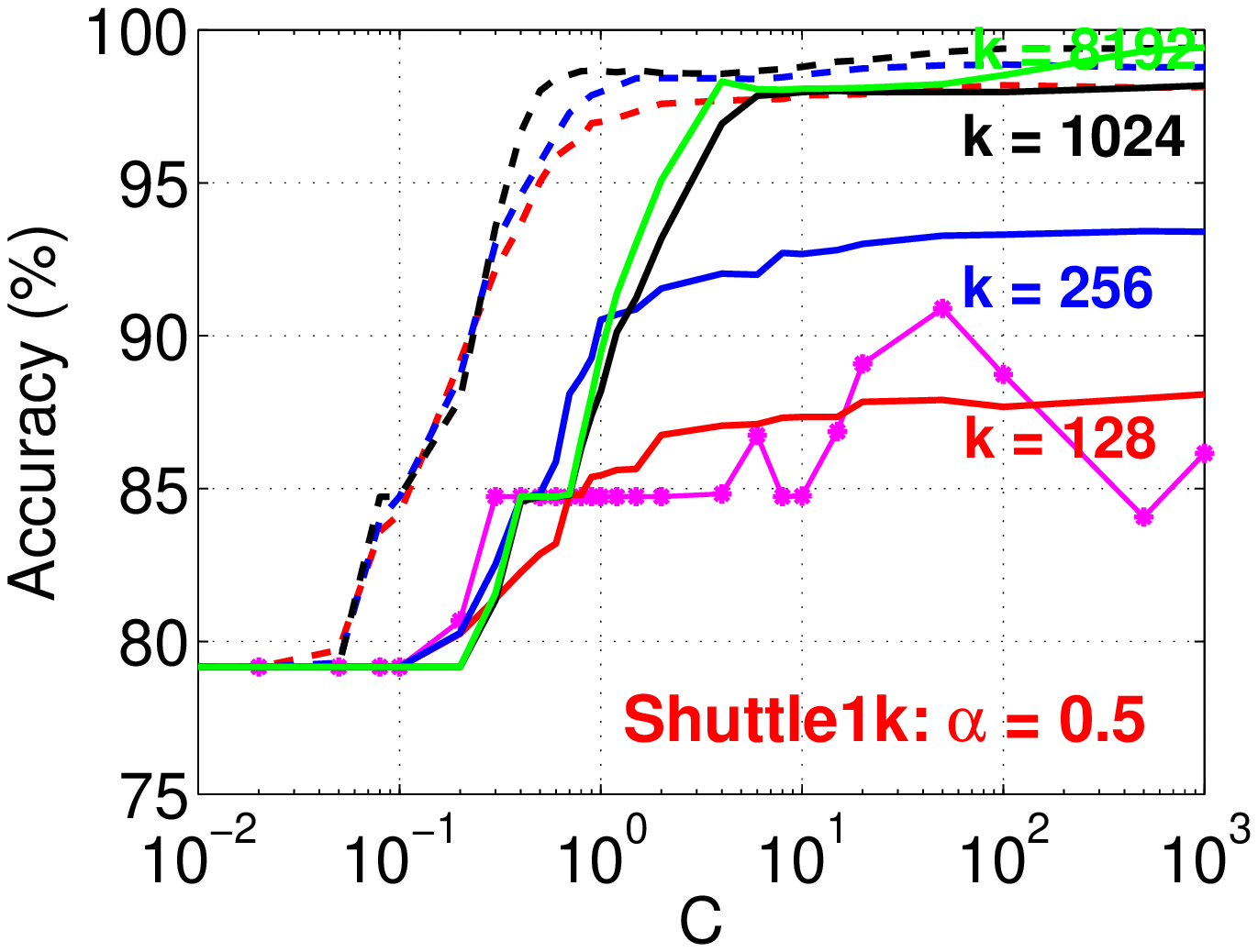}
}
\mbox{
\includegraphics[width=2.3in]{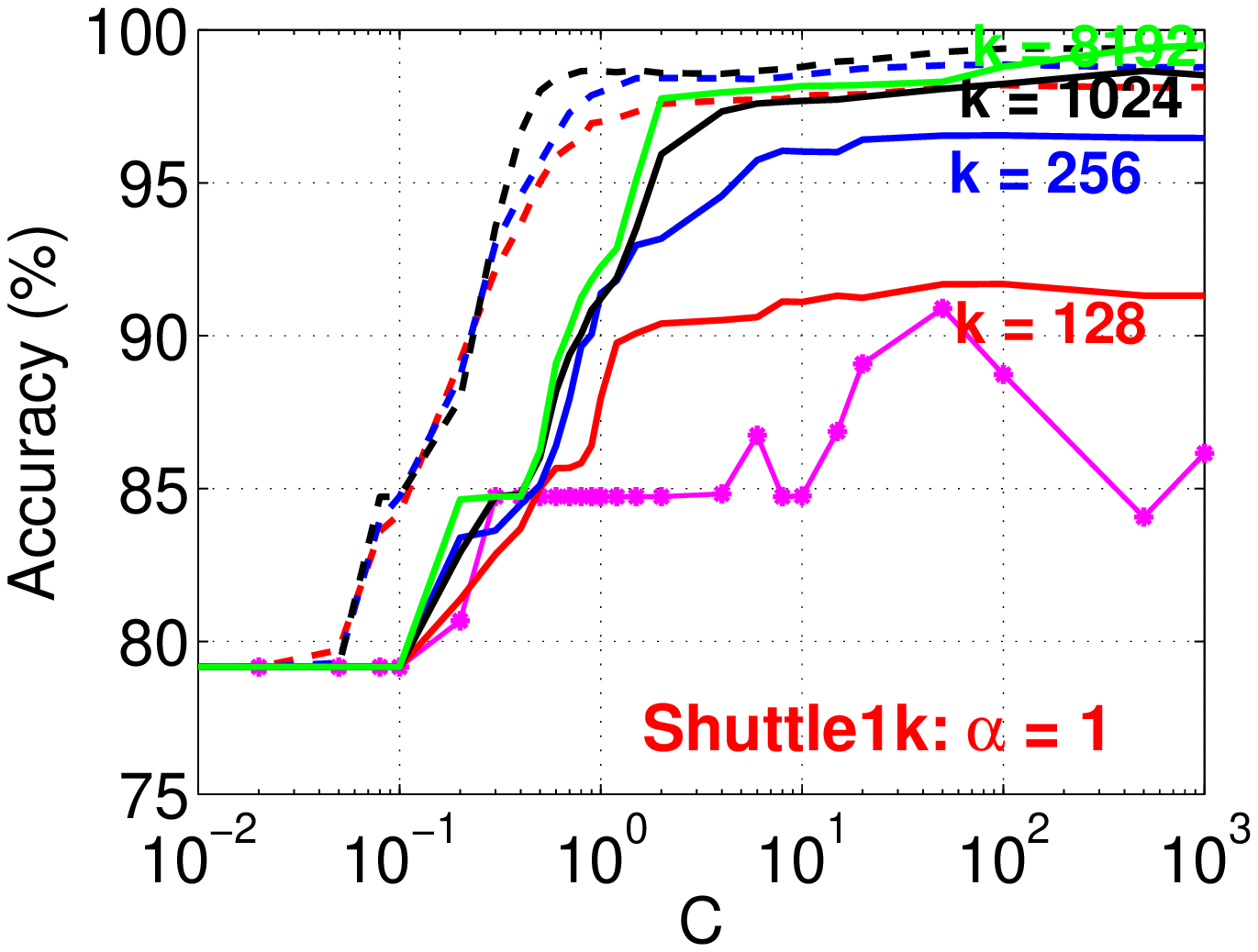}\hspace{-0.15in}
\includegraphics[width=2.3in]{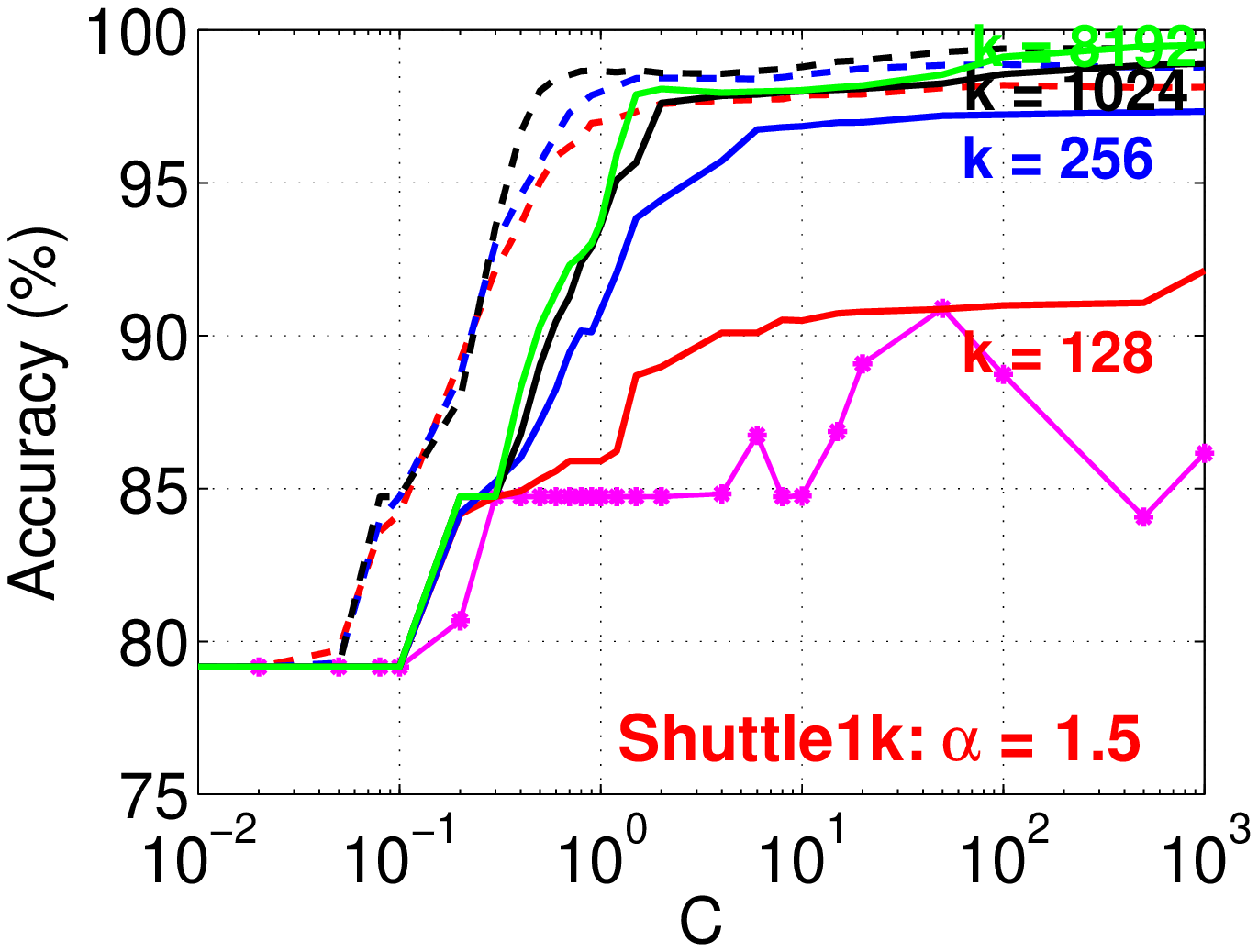}\hspace{-0.15in}
\includegraphics[width=2.3in]{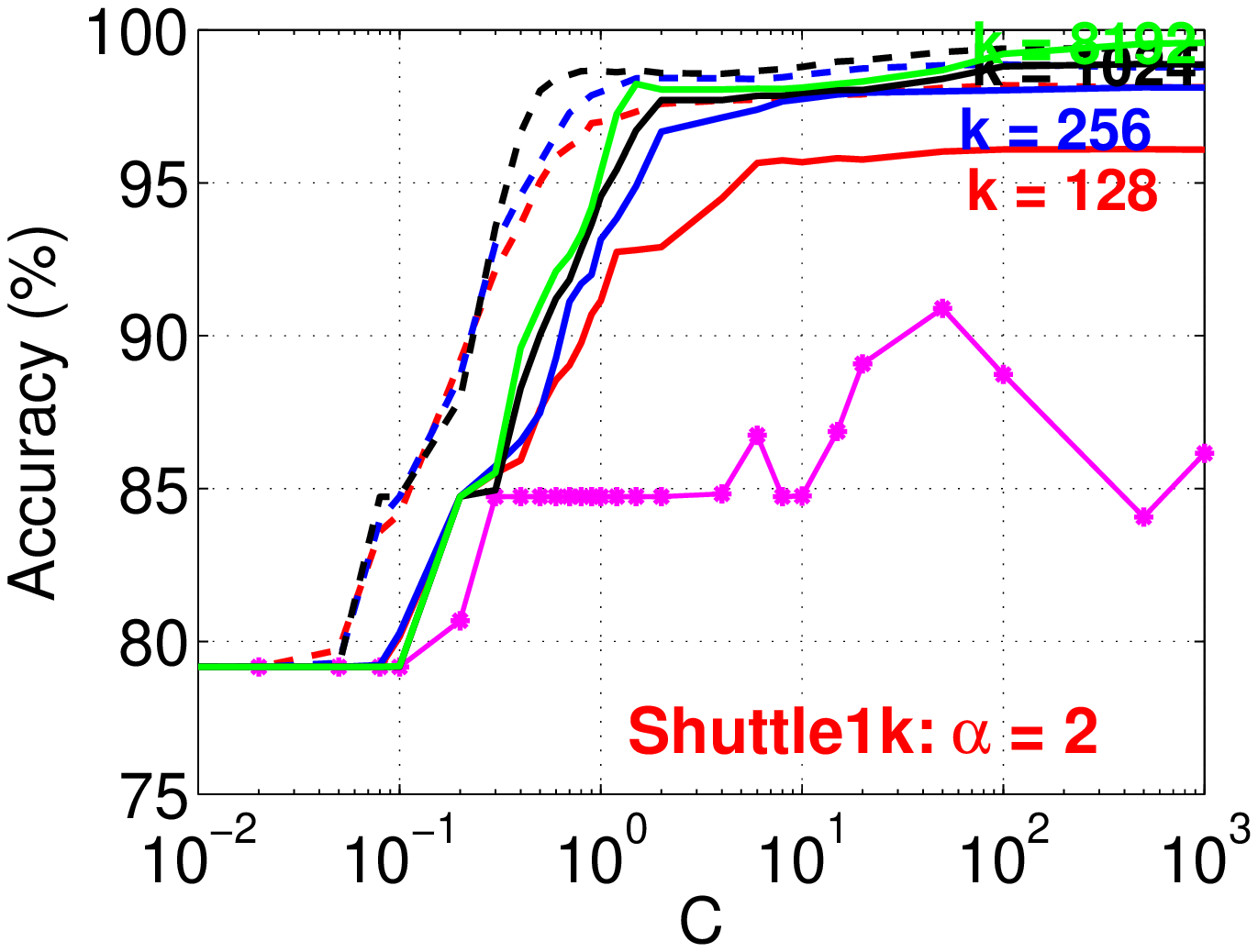}
}

\vspace{0.3in}

\mbox{
\includegraphics[width=2.3in]{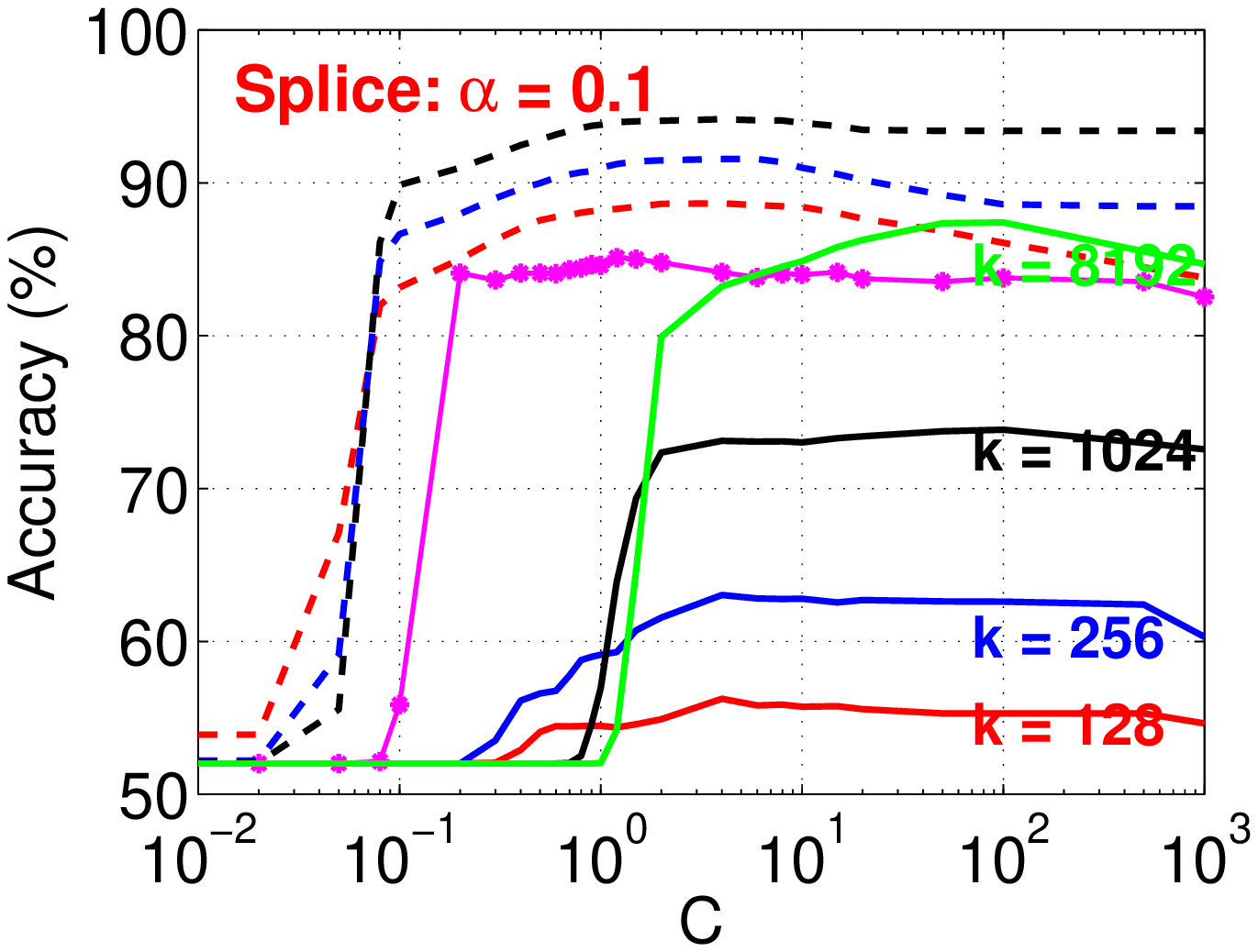}\hspace{-0.15in}
\includegraphics[width=2.3in]{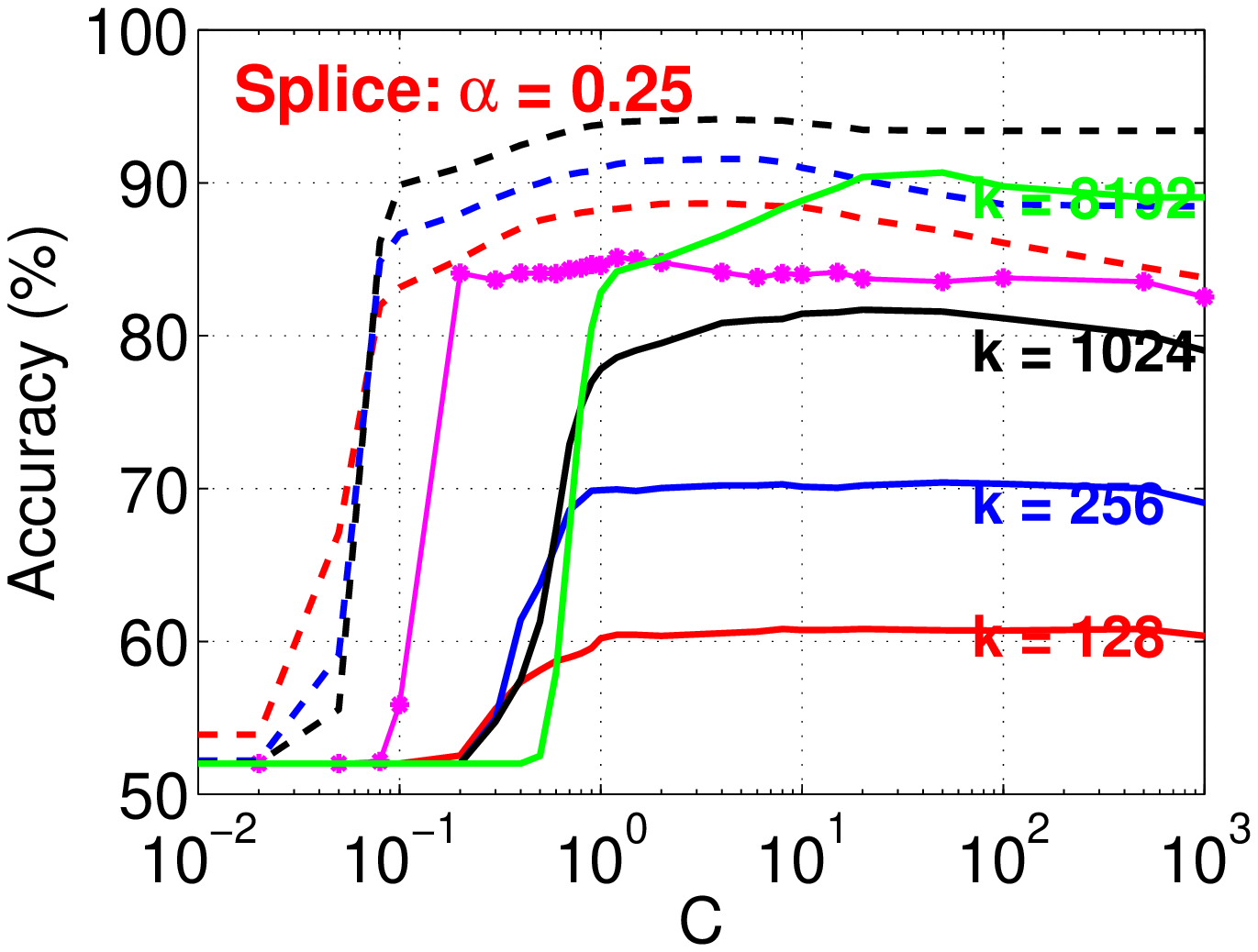}\hspace{-0.15in}
\includegraphics[width=2.3in]{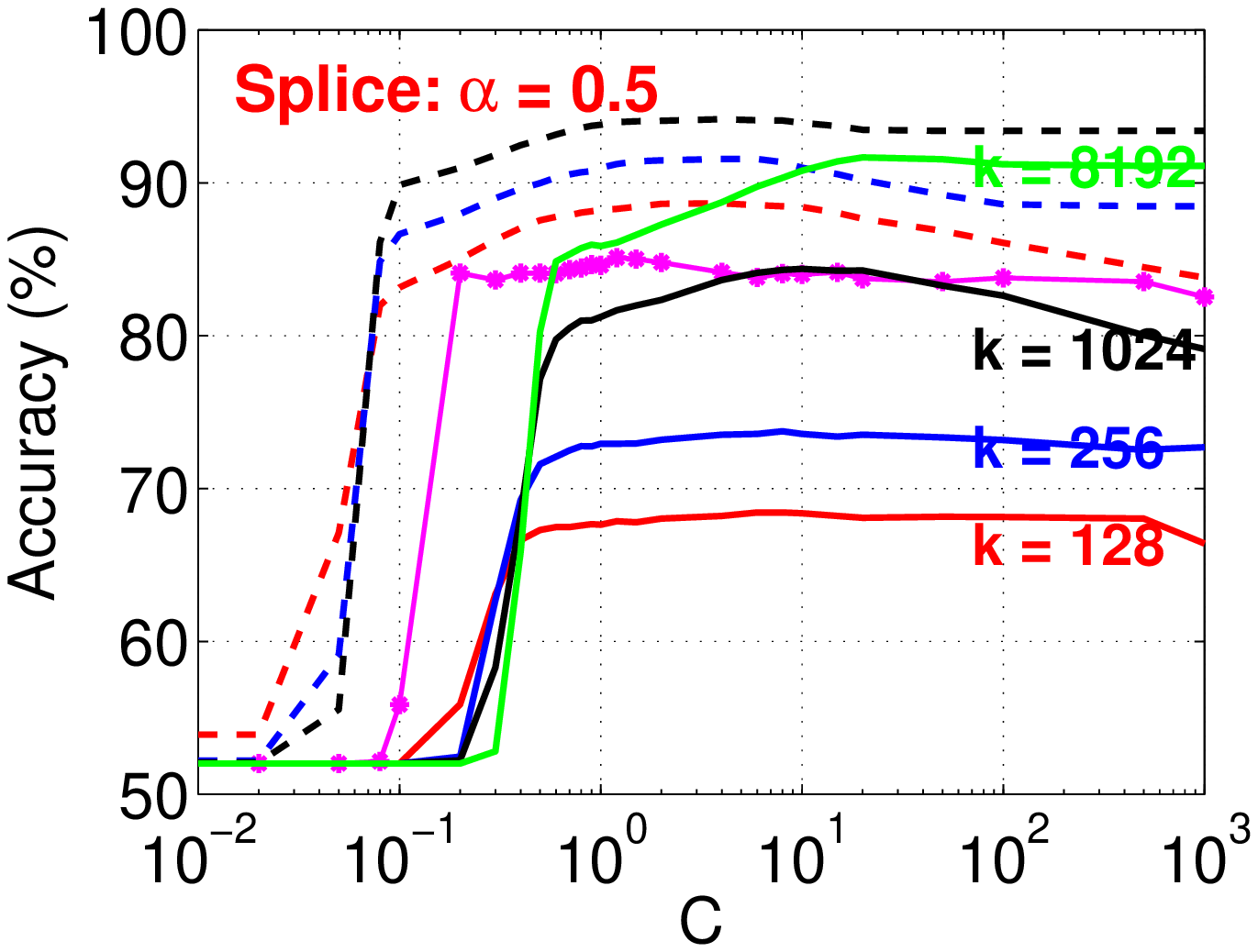}
}
\mbox{
\includegraphics[width=2.3in]{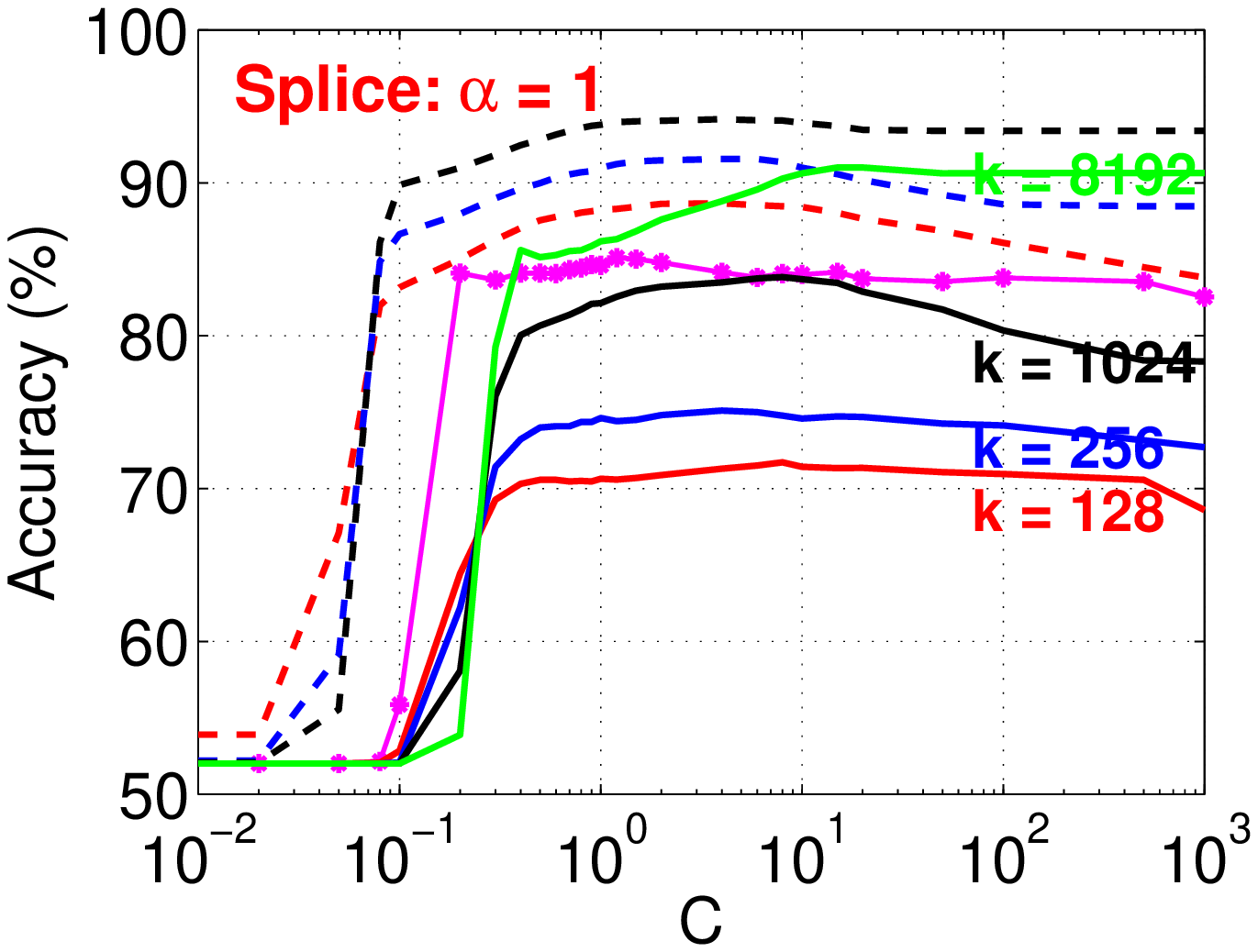}\hspace{-0.15in}
\includegraphics[width=2.3in]{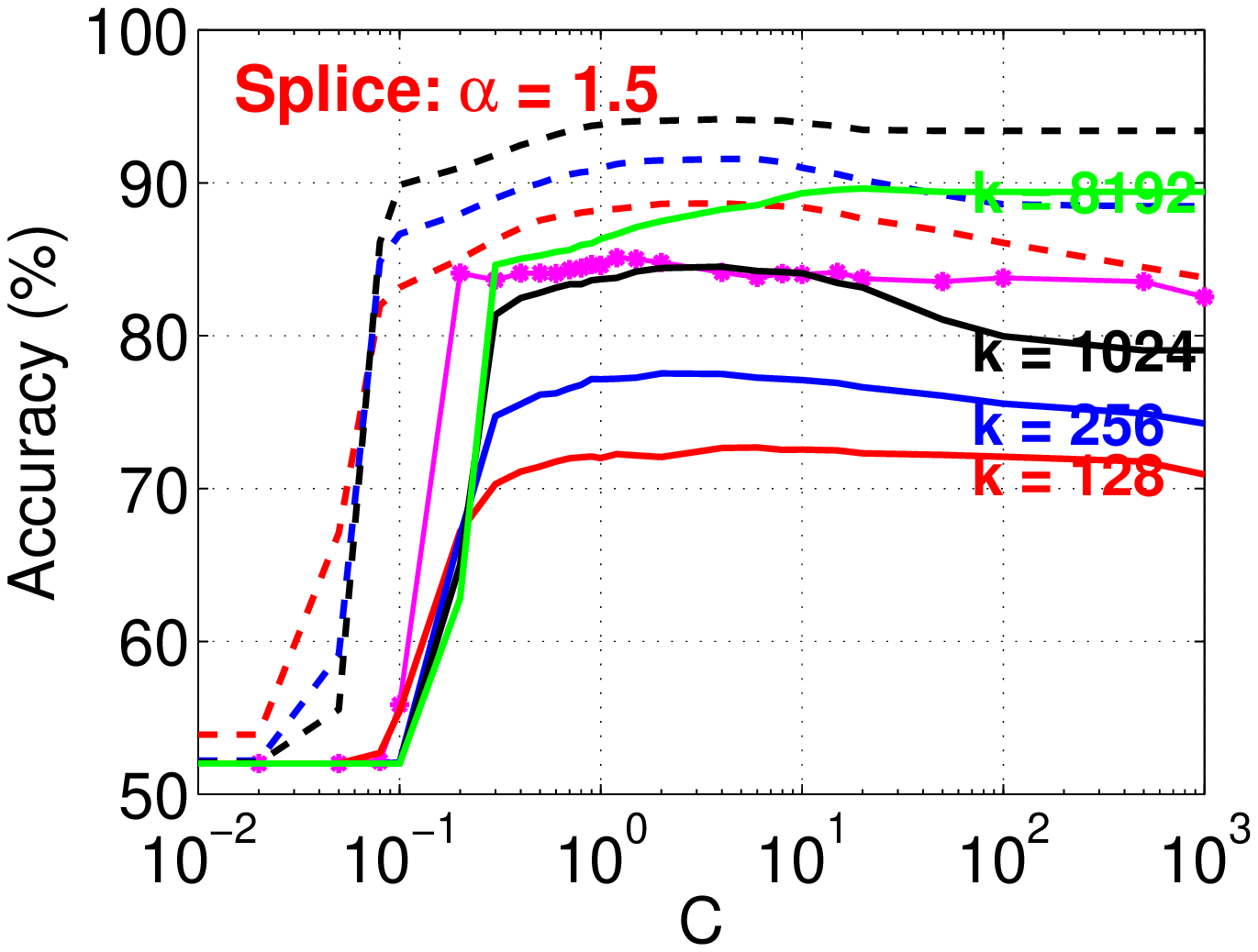}\hspace{-0.15in}
\includegraphics[width=2.3in]{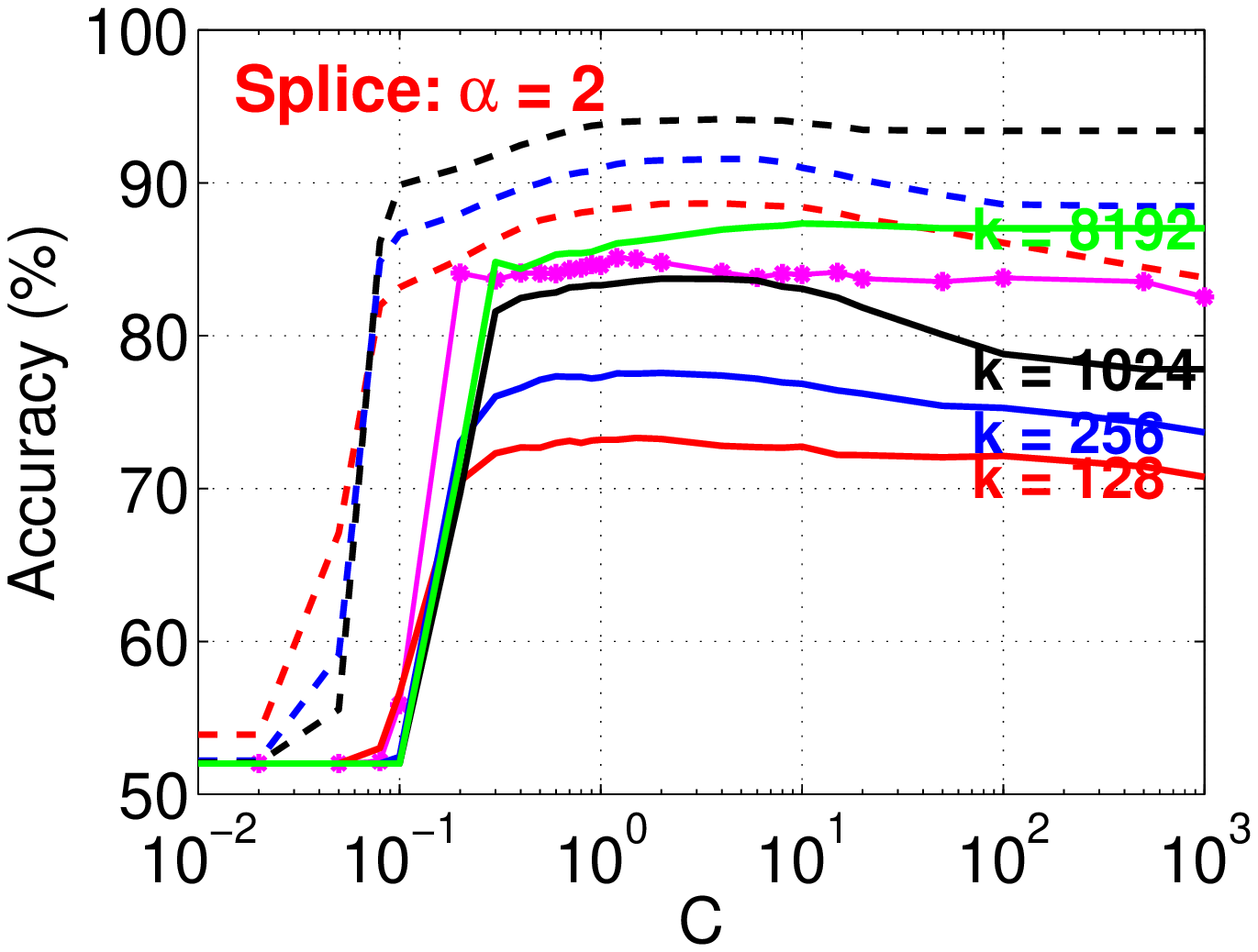}
}

\end{center}
\vspace{-0.3in}
\caption{\textbf{Shuttle1k} and {\bf Splice}. We compare sign $\alpha$-stable random projections with 0-bit consistent weighted sampling (CWS). Each panel (for each $\alpha$) consists of 8 curves. The solid (pink) curve marked by * represents the results of linear SVM. Four solid curves (labelled by $k=128$, $k=256$, $k=1024$, and $k=8192$, respectively) represent the results of sign $\alpha$-stable random projections for 4 different $k$ values.  The 3 dashed curves  correspond to the results of 0-bit CWS for $k=128, 256, 1024$ (a higher curve for a higher $k$ value). These experimental results, all conducted using LIBLINEAR, show that 0-bit CWS requires much fewer samples to achieve the sample accuracies.    }\label{fig_CWS3}
\end{figure}

\begin{figure}[h!]
\begin{center}

\mbox{
\includegraphics[width=2.3in]{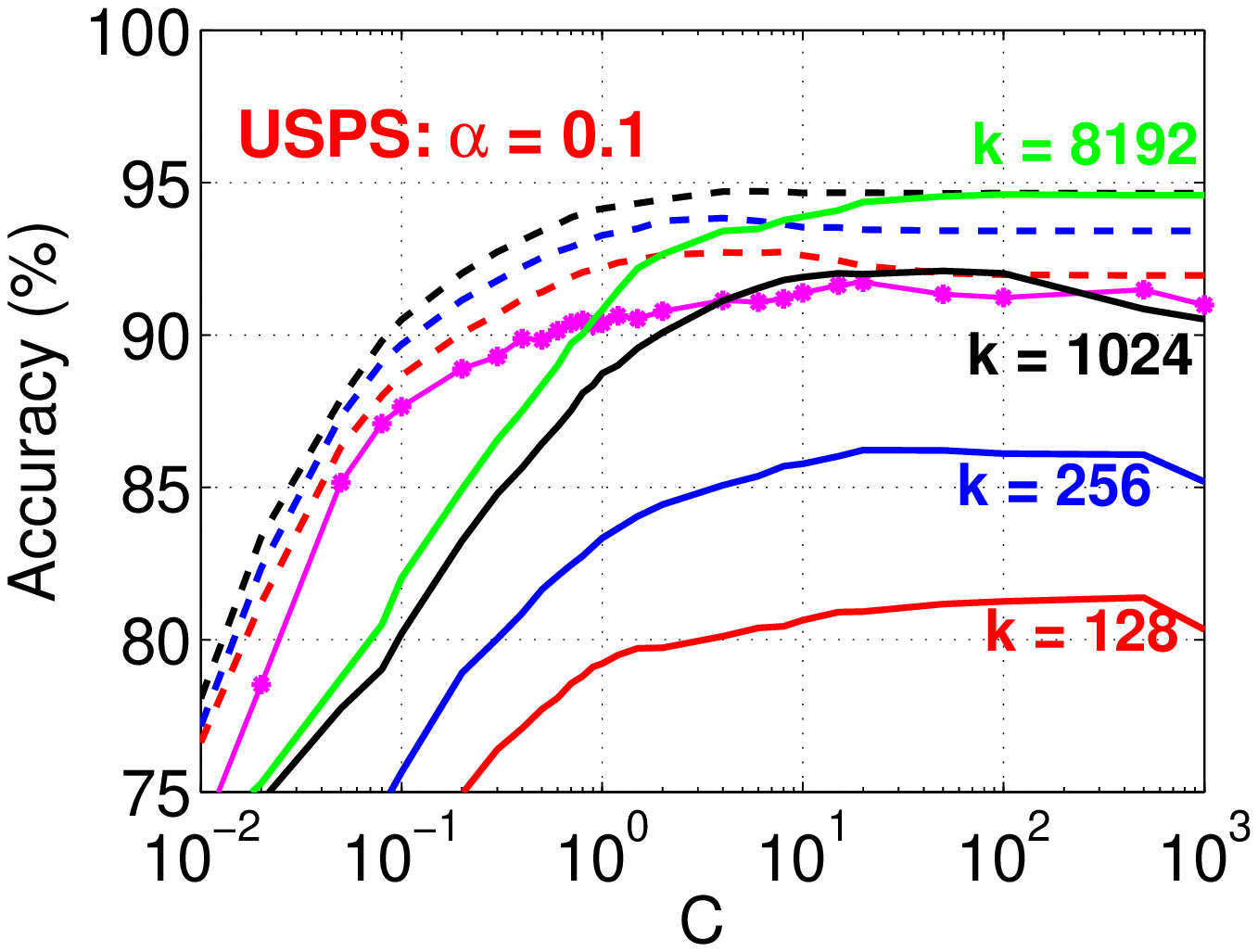}\hspace{-0.15in}
\includegraphics[width=2.3in]{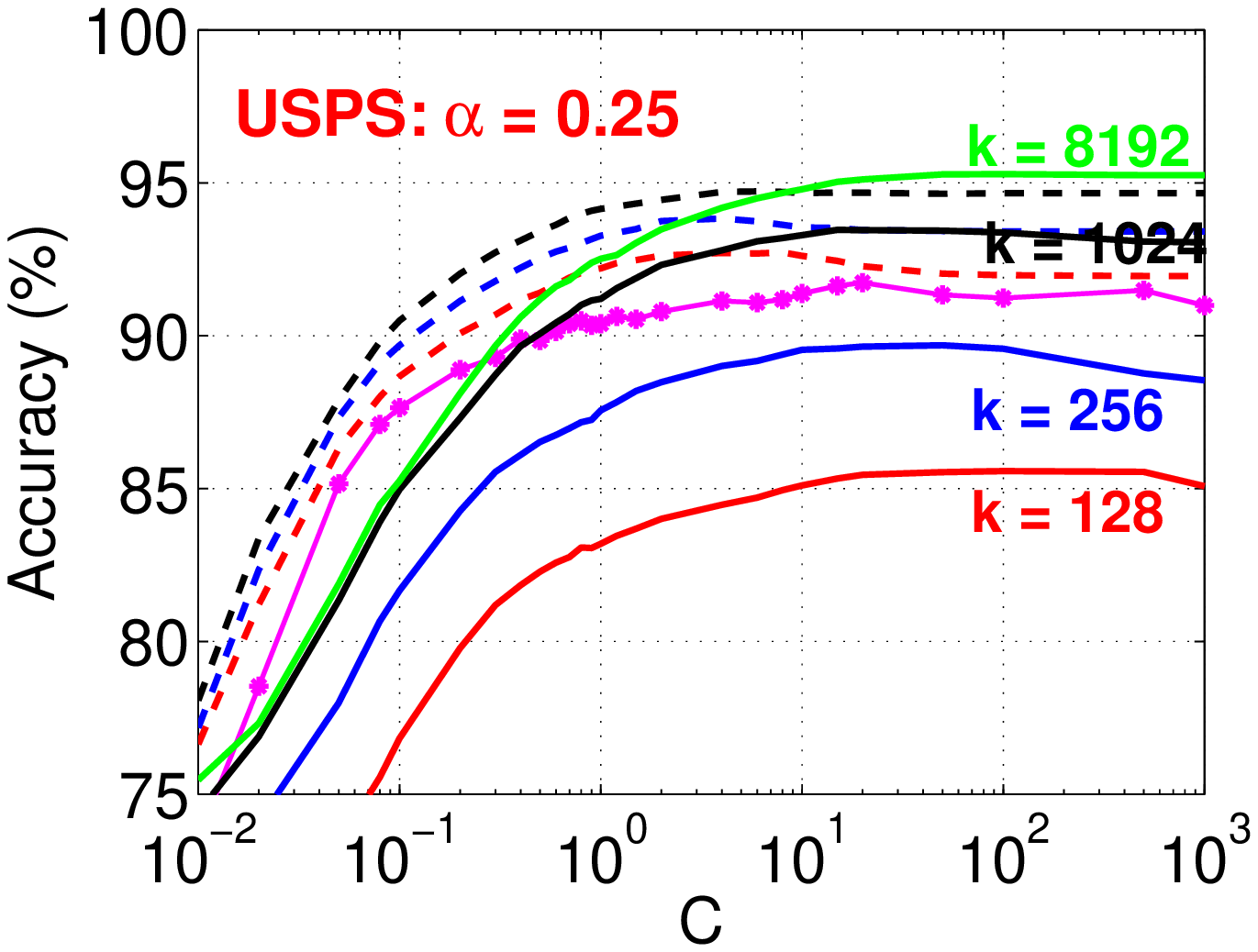}\hspace{-0.15in}
\includegraphics[width=2.3in]{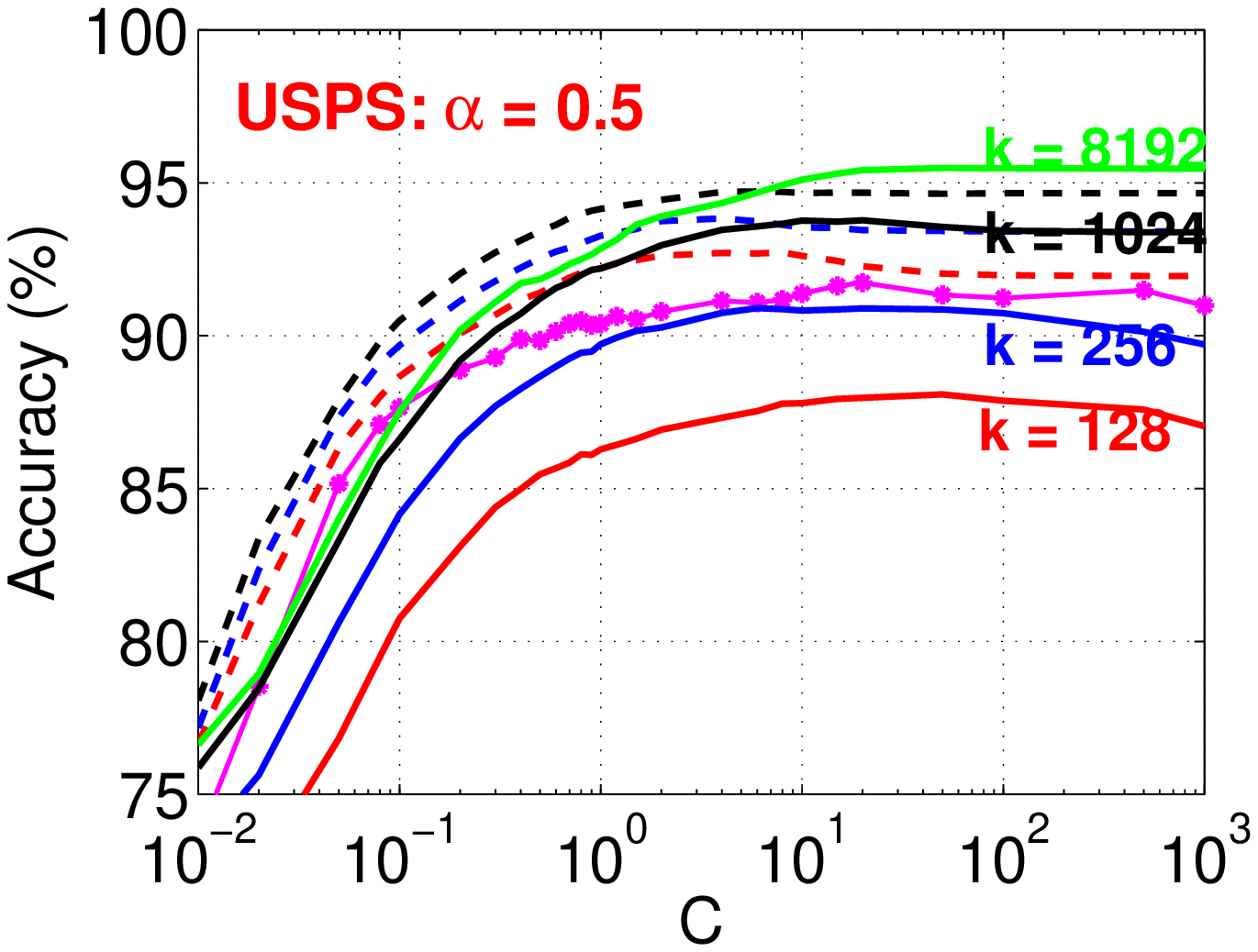}
}
\mbox{
\includegraphics[width=2.3in]{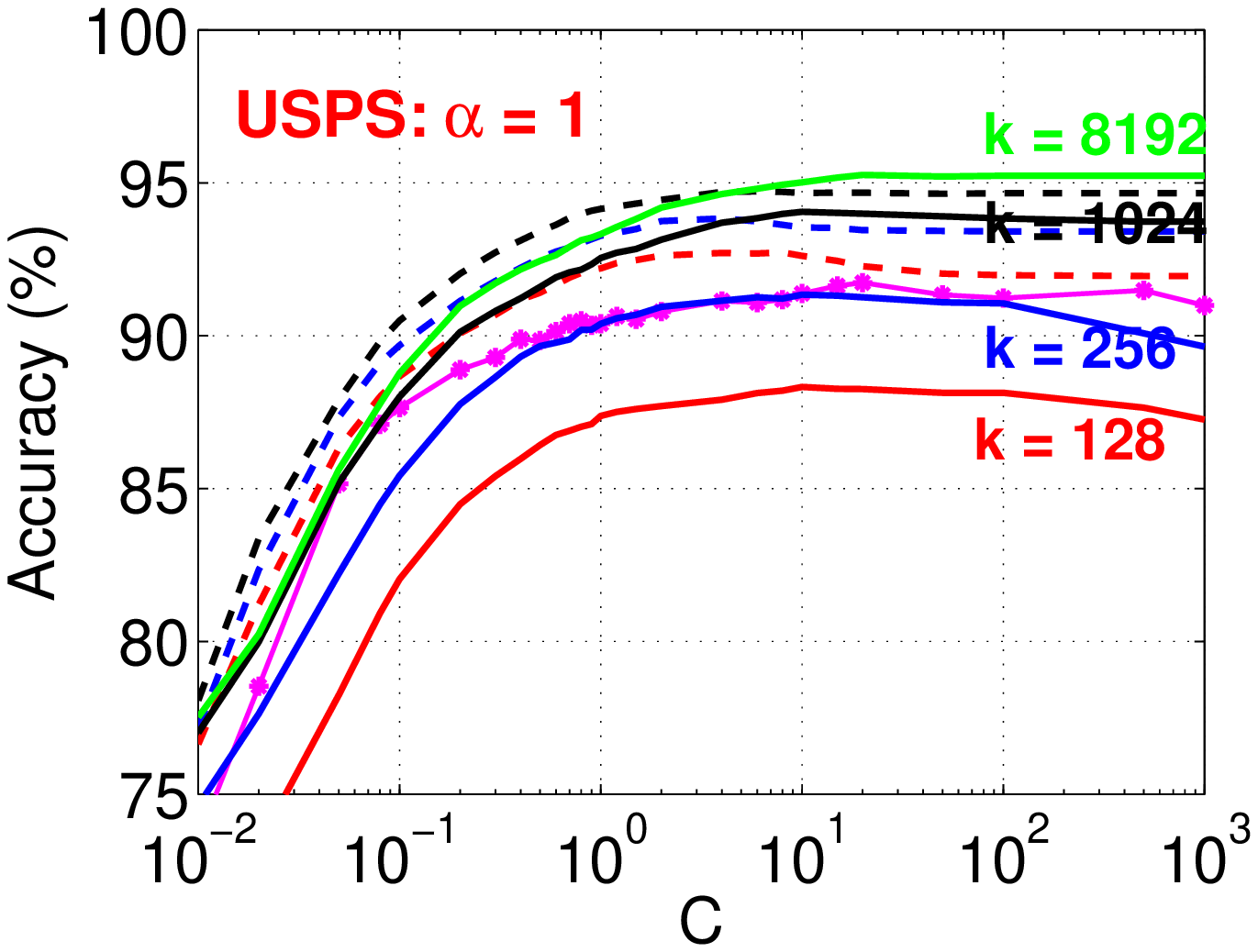}\hspace{-0.15in}
\includegraphics[width=2.3in]{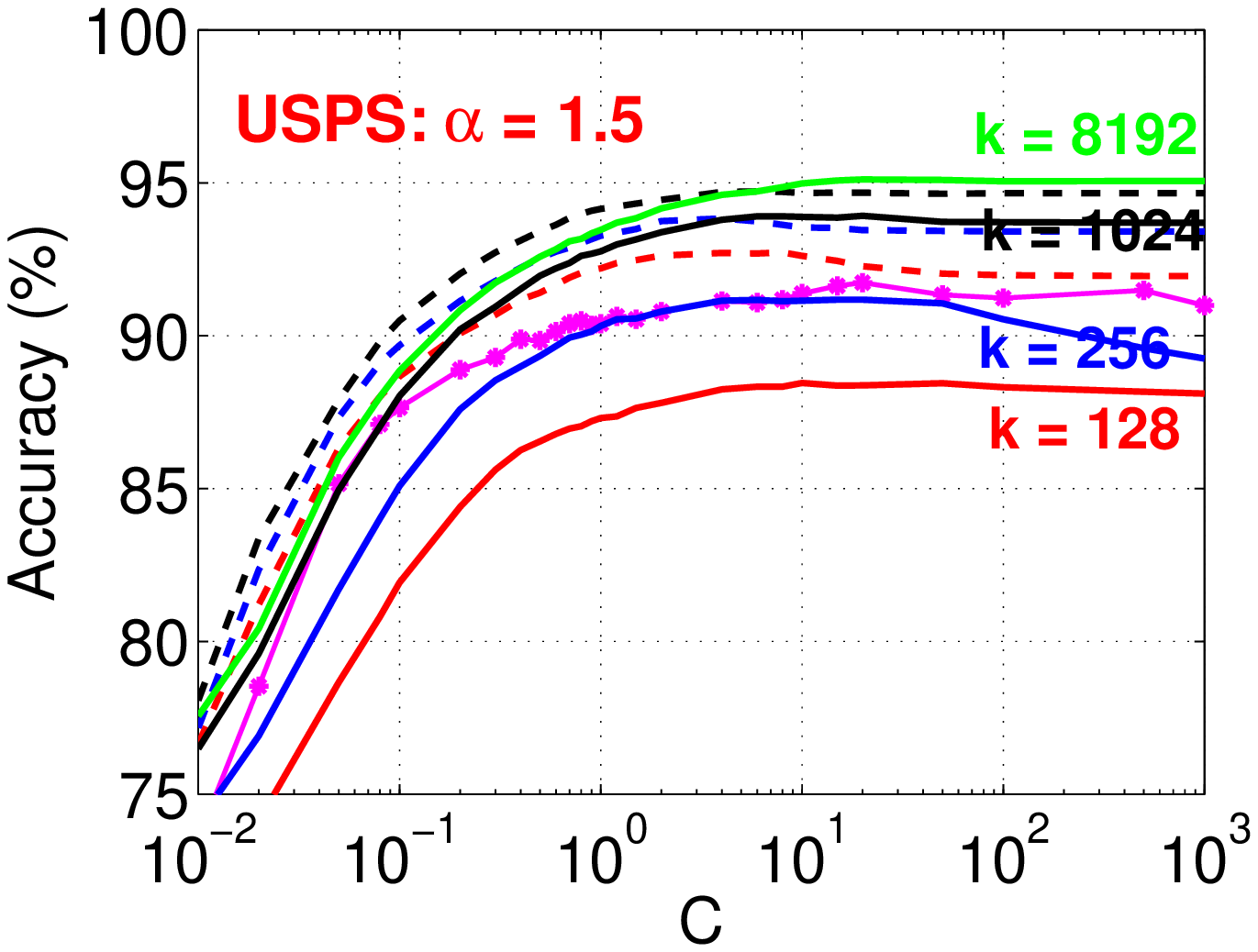}\hspace{-0.15in}
\includegraphics[width=2.3in]{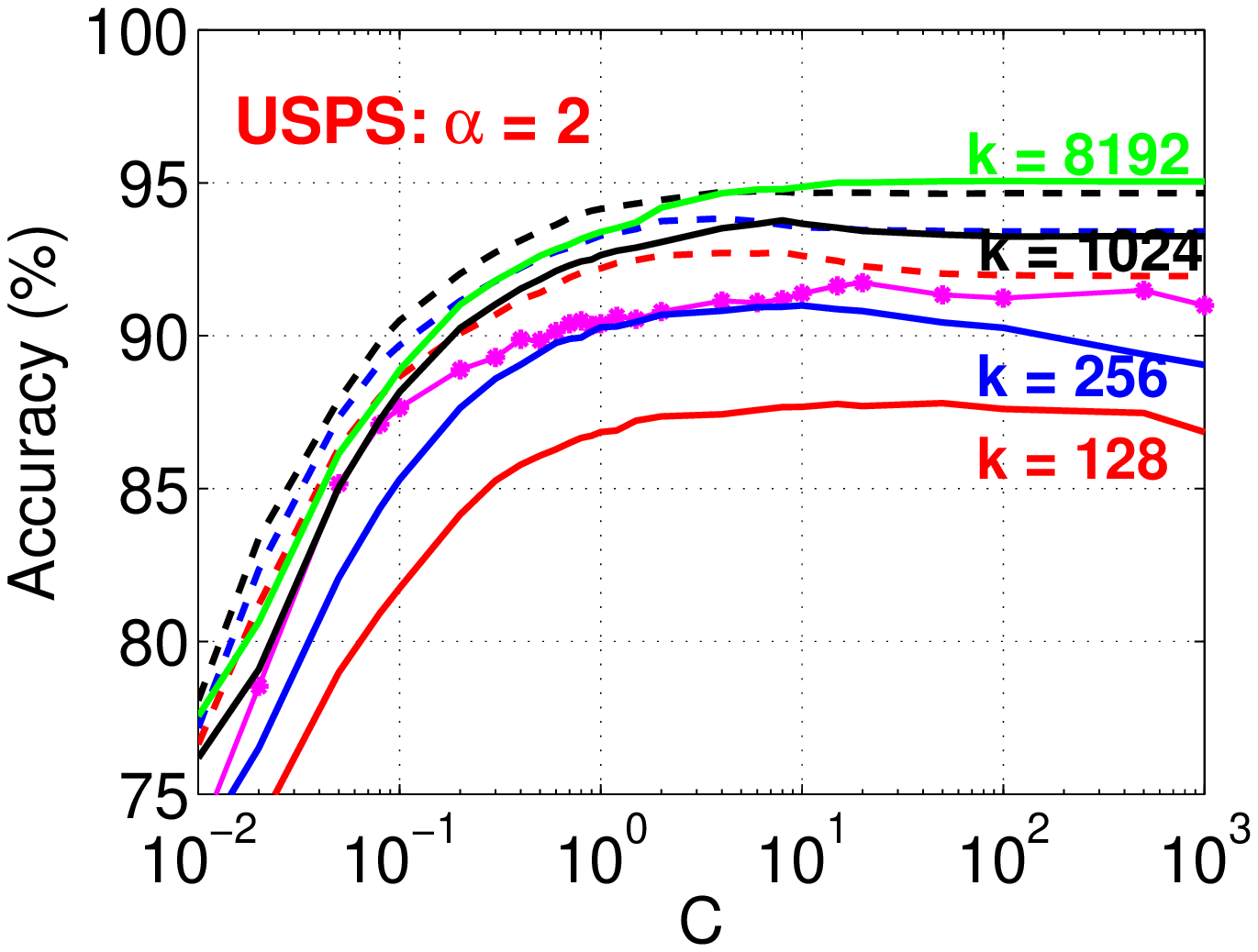}
}

\vspace{0.3in}

\mbox{
\includegraphics[width=2.3in]{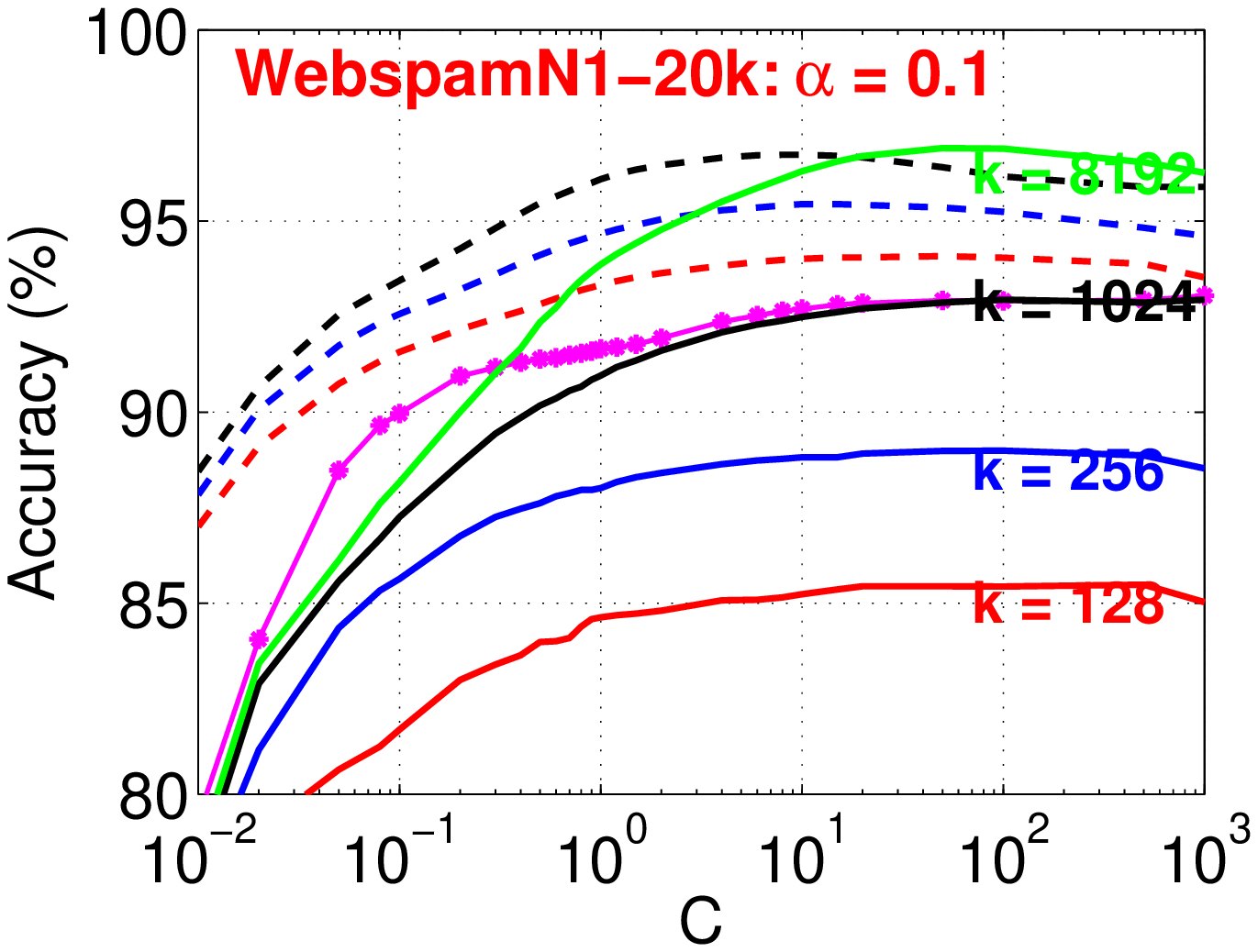}\hspace{-0.15in}
\includegraphics[width=2.3in]{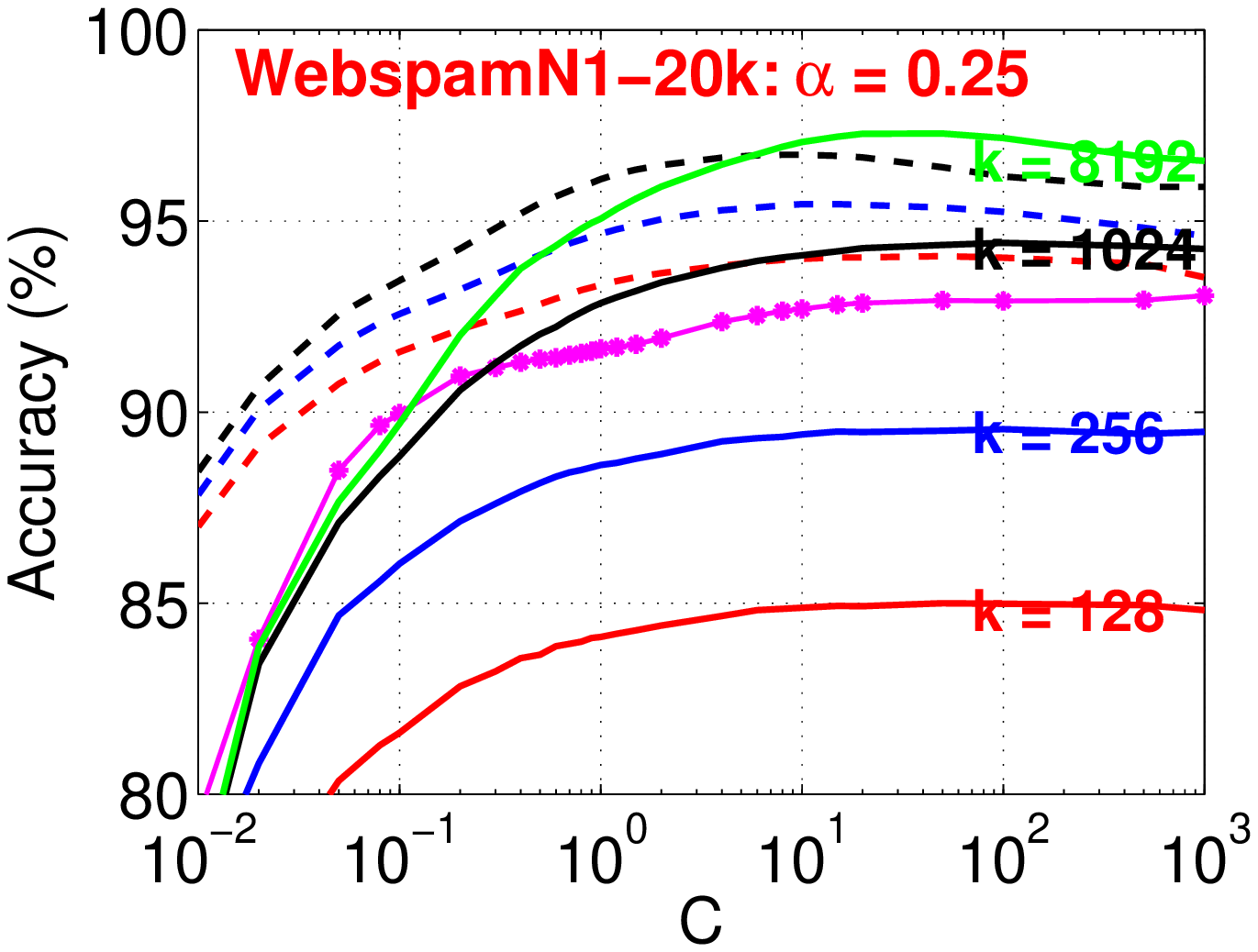}\hspace{-0.15in}
\includegraphics[width=2.3in]{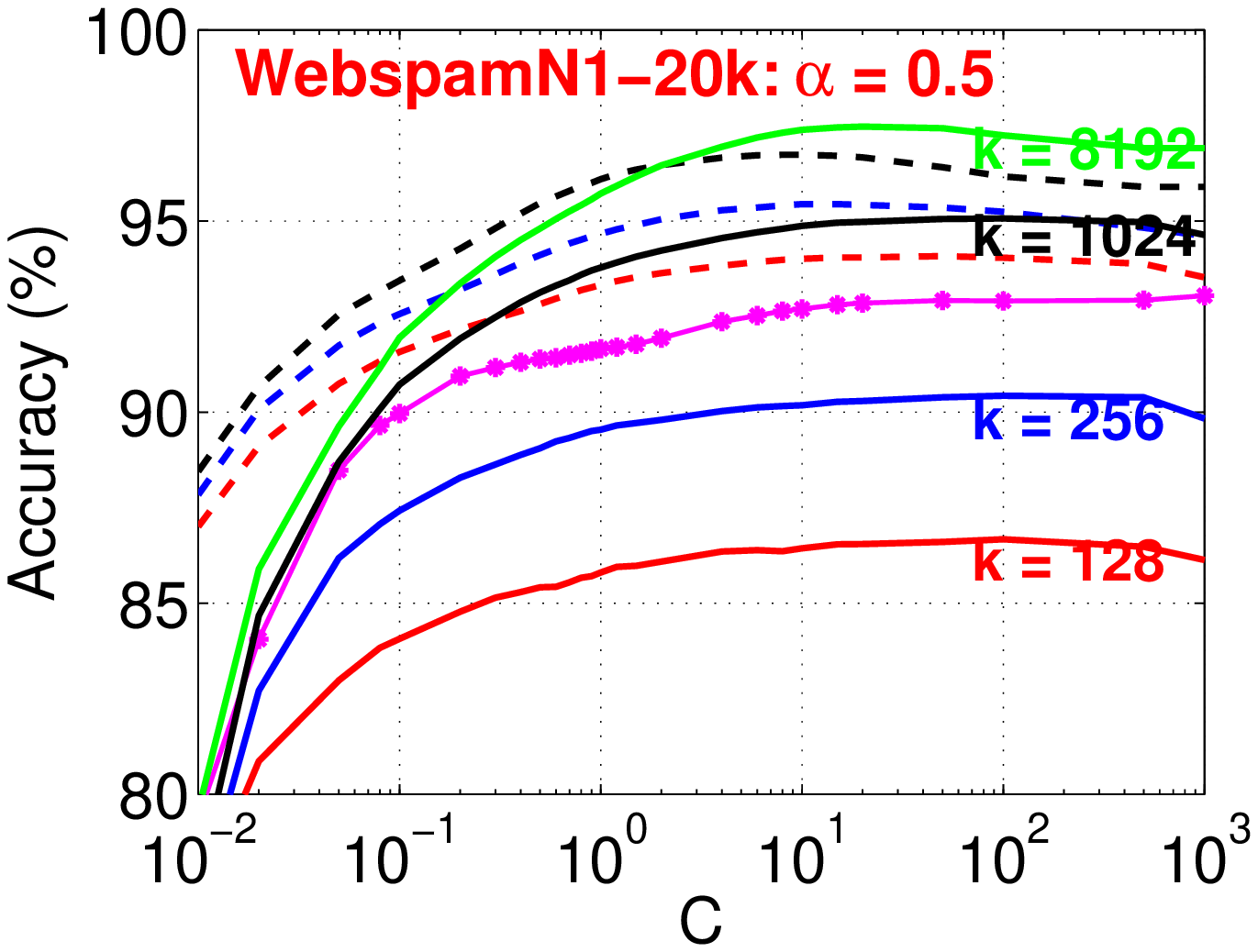}
}
\mbox{
\includegraphics[width=2.3in]{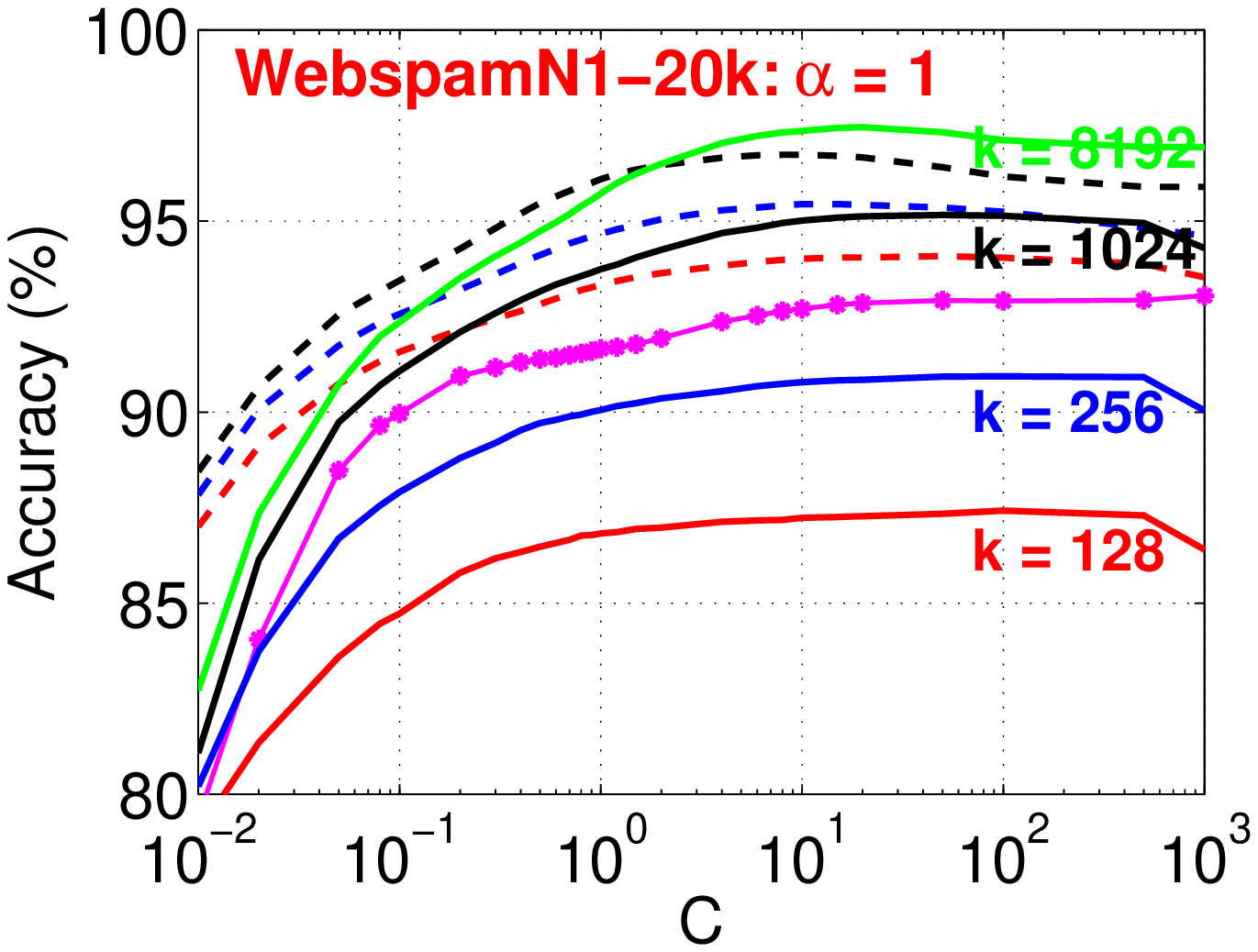}\hspace{-0.15in}
\includegraphics[width=2.3in]{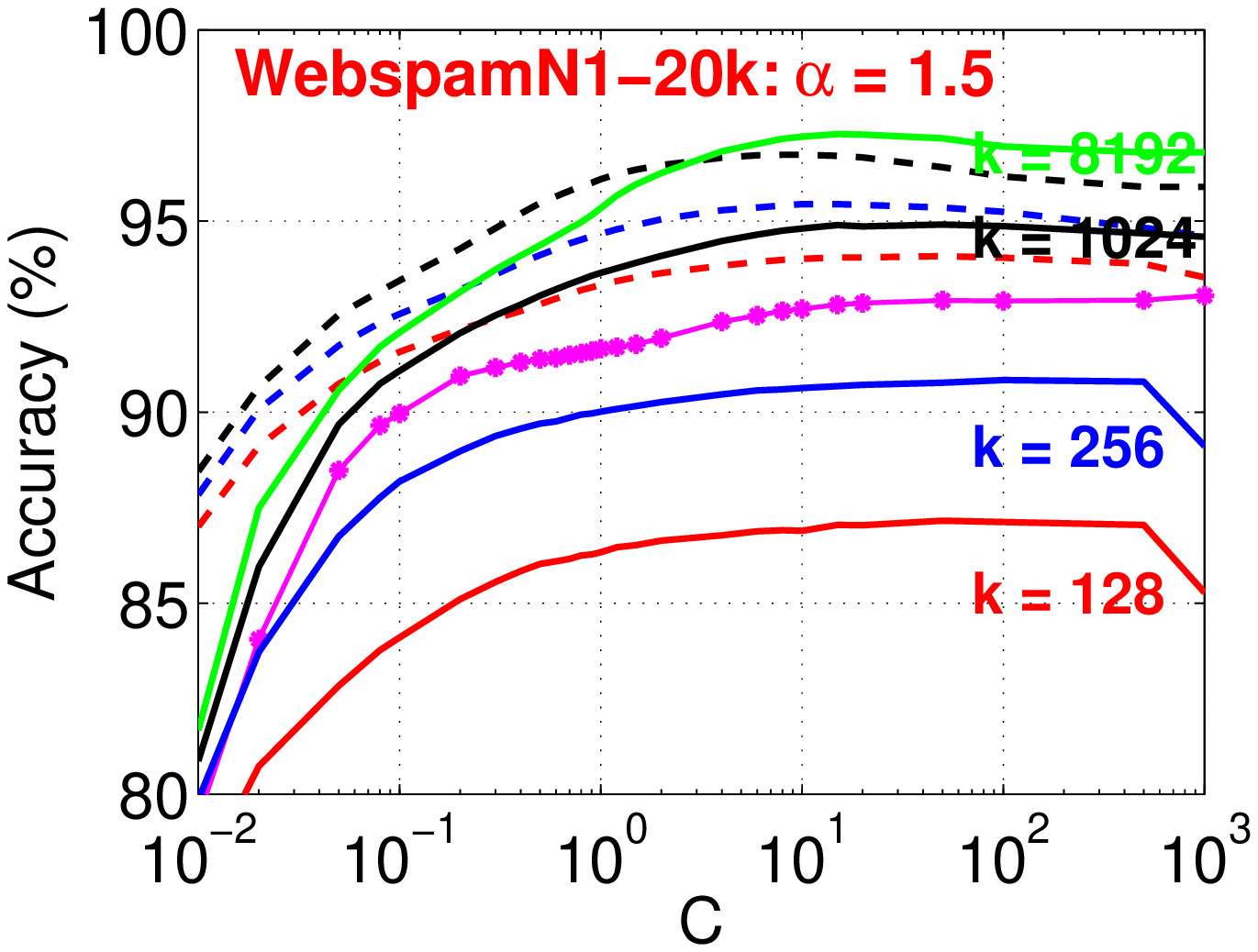}\hspace{-0.15in}
\includegraphics[width=2.3in]{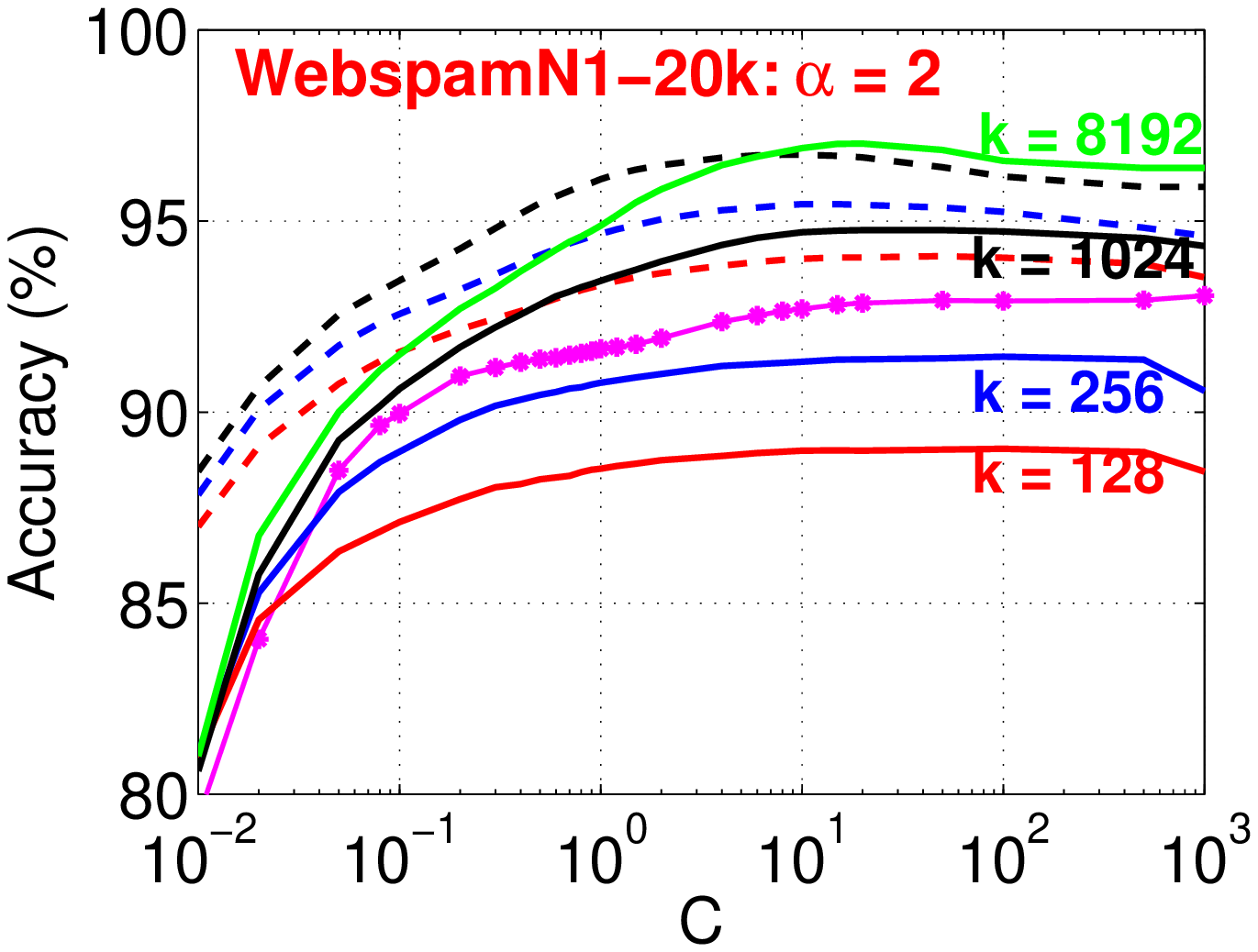}
}

\end{center}
\vspace{-0.3in}
\caption{ \textbf{USPS} and {\bf WebspamN1-20k}. We compare sign $\alpha$-stable random projections with 0-bit consistent weighted sampling (CWS). Each panel (for each $\alpha$) consists of 8 curves. The solid (pink) curve marked by * represents the results of linear SVM. Four solid curves (labelled by $k=128$, $k=256$, $k=1024$, and $k=8192$, respectively) represent the results of sign $\alpha$-stable random projections for 4 different $k$ values.  The 3 dashed curves  correspond to the results of 0-bit CWS for $k=128, 256, 1024$ (a higher curve for a higher $k$ value). These experimental results, all conducted using LIBLINEAR, show that 0-bit CWS requires much fewer samples to achieve the sample accuracies.   }\label{fig_CWS4}
\end{figure}

Figures~\ref{fig_CWS1} to~\ref{fig_CWS4} compare sign $\alpha$-stable random projections with 0-bit CWS~\cite{Report:Li_CWS15} on selected datasets. For clarity, we only show the results of sign stable random projections for $k=128, 256, 1024, 8192$ projections, and the results for 0-bit CWS with $k=128,256,1024$ samples.  These results demonstrate that 0-bit CWS  requires much fewer samples, although we should keep in mind that 0-bit CWS is only for nonnegative data.
\vspace{-0.3in}

\newpage\clearpage

\subsection{Experiment on a Larger Dataset}

The paper on 0-bit CWS~\cite{Report:Li_CWS15} only experimented with datasets of moderate sizes for an important reason. To prove the correctness, they need to show that the result of 0-bit CWS with enough samples could approach that of exact min-max kernel. A straightforward and faithful implementation of SVM with min-max kernel is to use the LIBSVM pre-computed kernel functionality by computing the kernel explicitly and feeding it to SVM from outside. This strategy, although most repeatable, is very expensive for datasets which are not even large~\cite{Book:Bottou_07}. On other hand, once we have proved the correctness of 0-bit CWS, applying the method to larger datasets is easy, except that we would not be able to compute the exact result of min-max kernel.

\begin{figure}[h!]
\begin{center}

\mbox{
\includegraphics[width=2.3in]{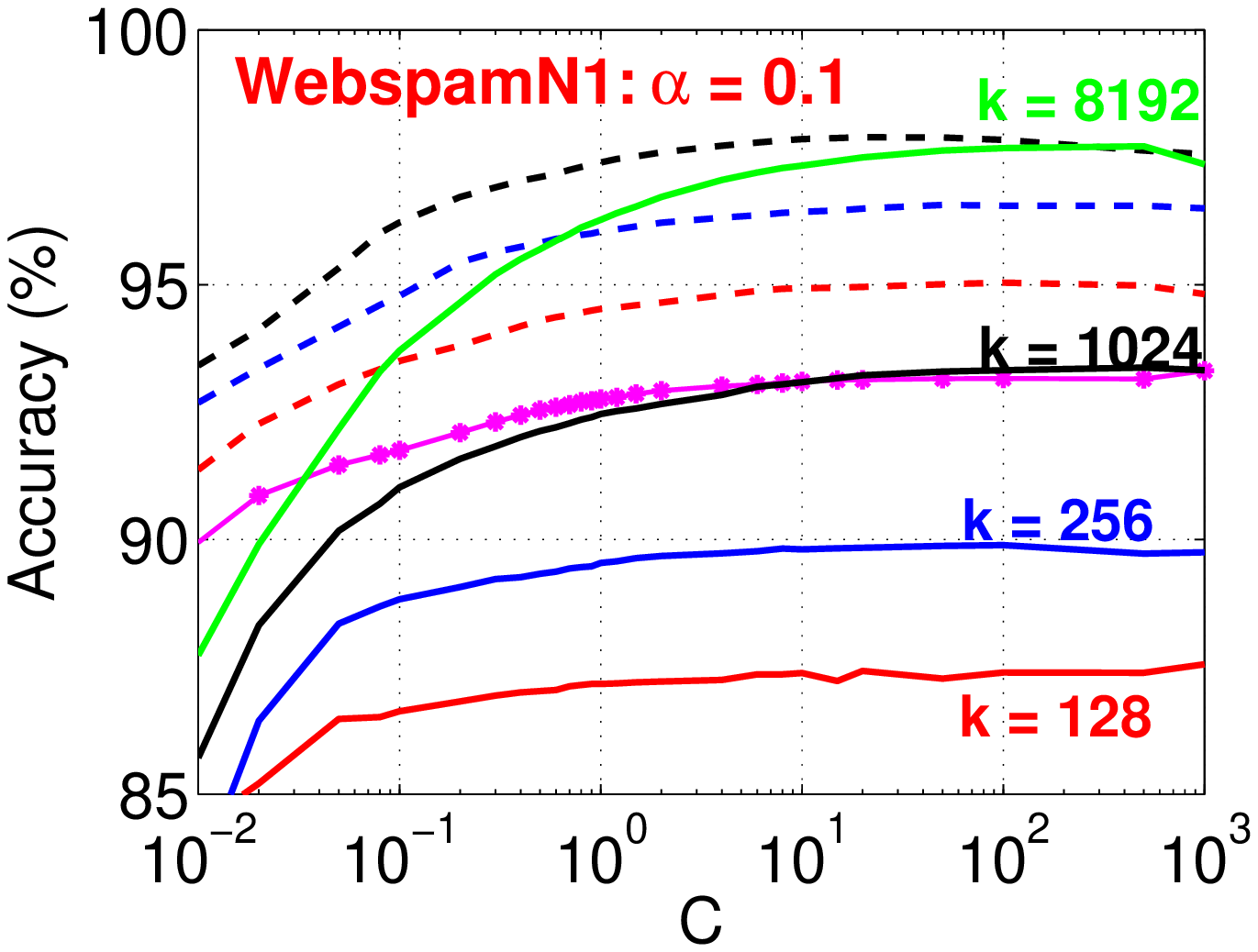}\hspace{-0.15in}
\includegraphics[width=2.3in]{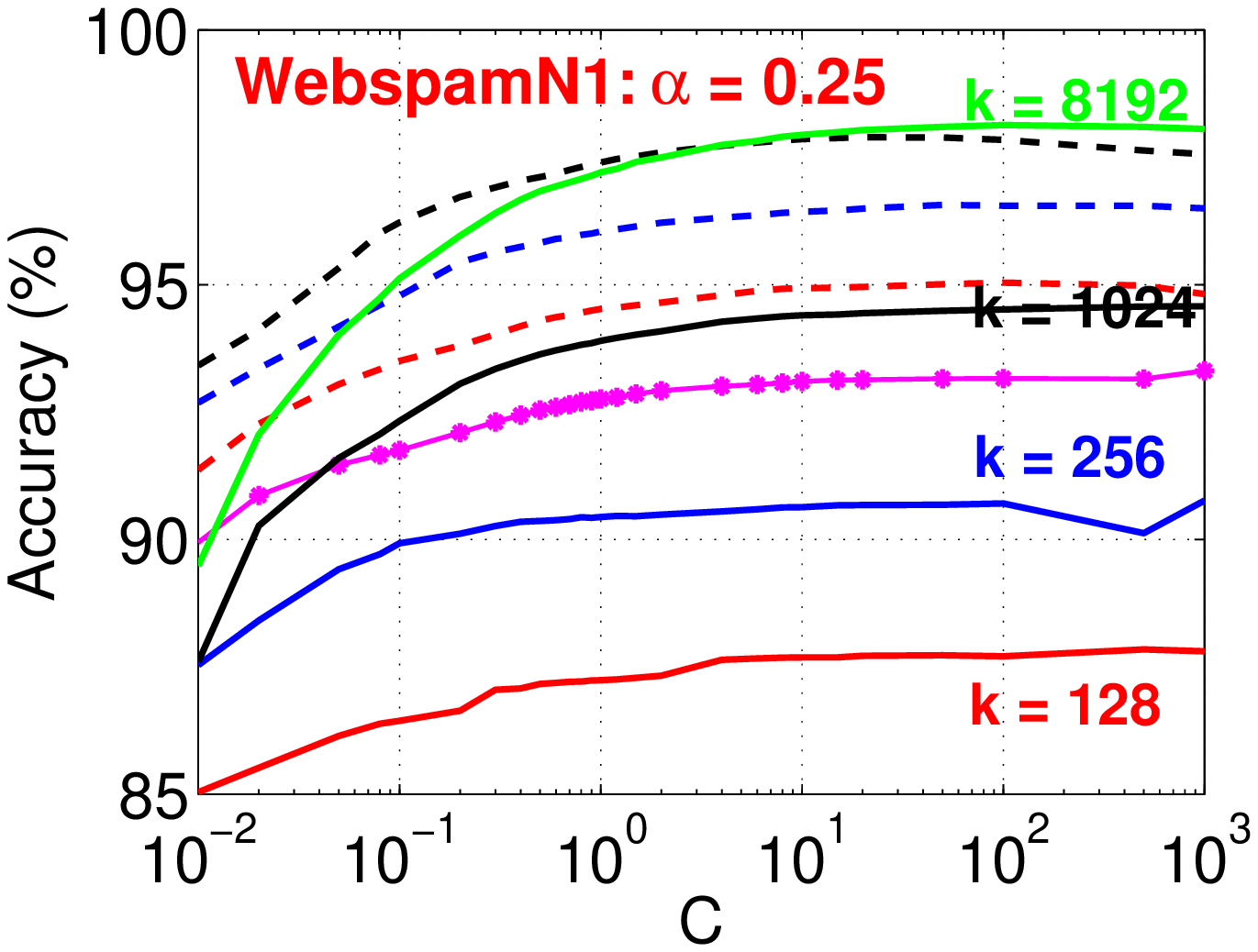}\hspace{-0.15in}
\includegraphics[width=2.3in]{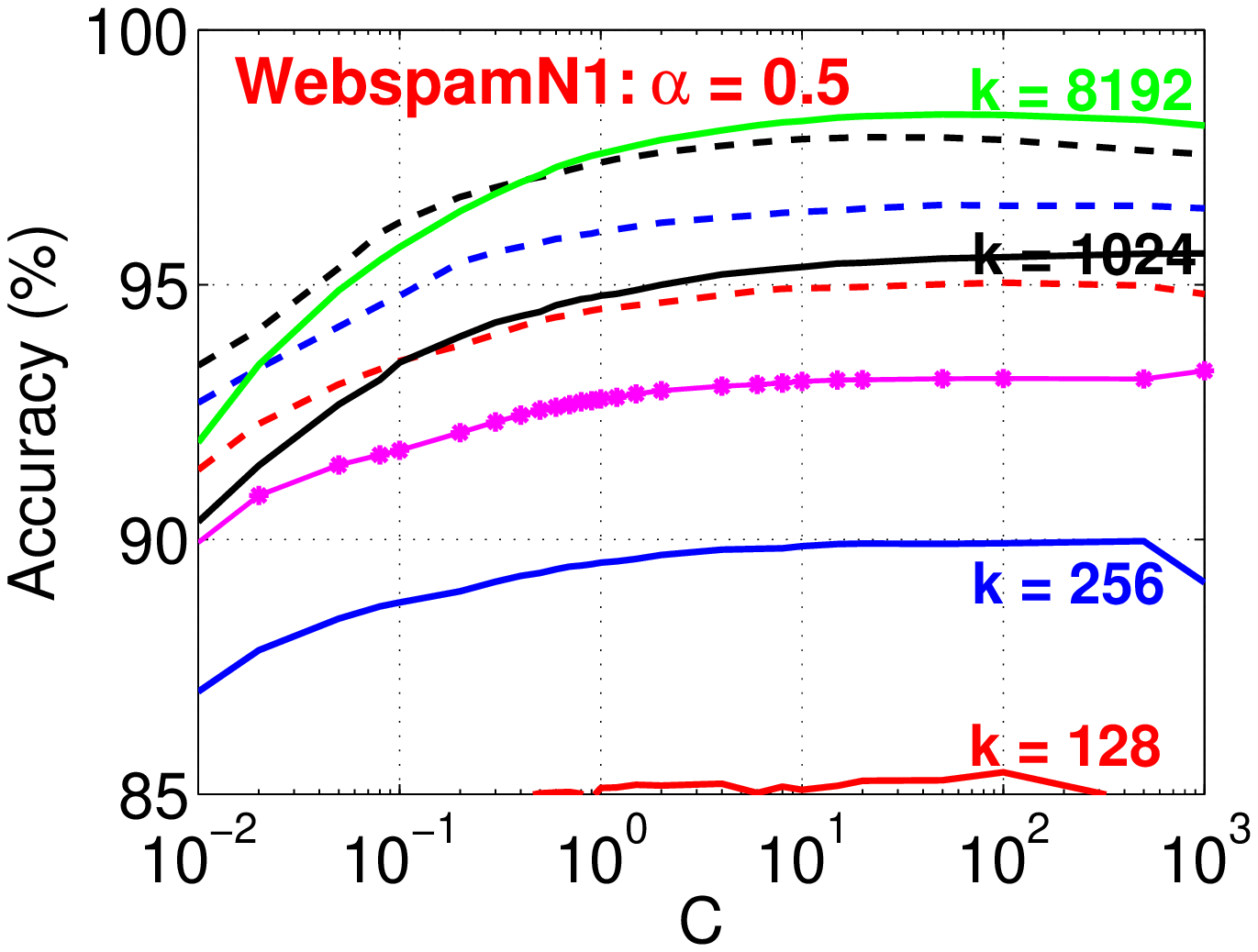}
}

\mbox{
\includegraphics[width=2.3in]{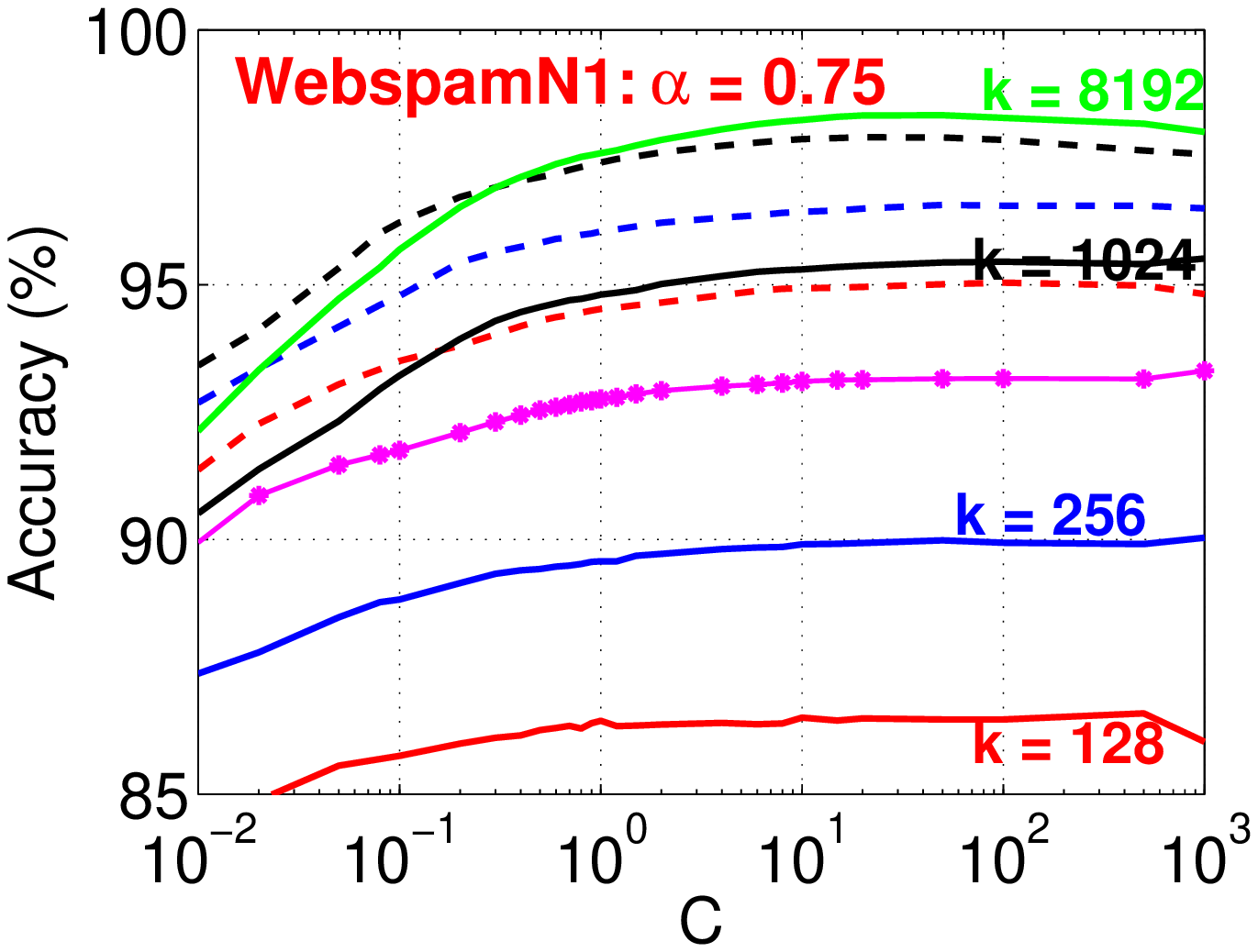}\hspace{-0.15in}
\includegraphics[width=2.3in]{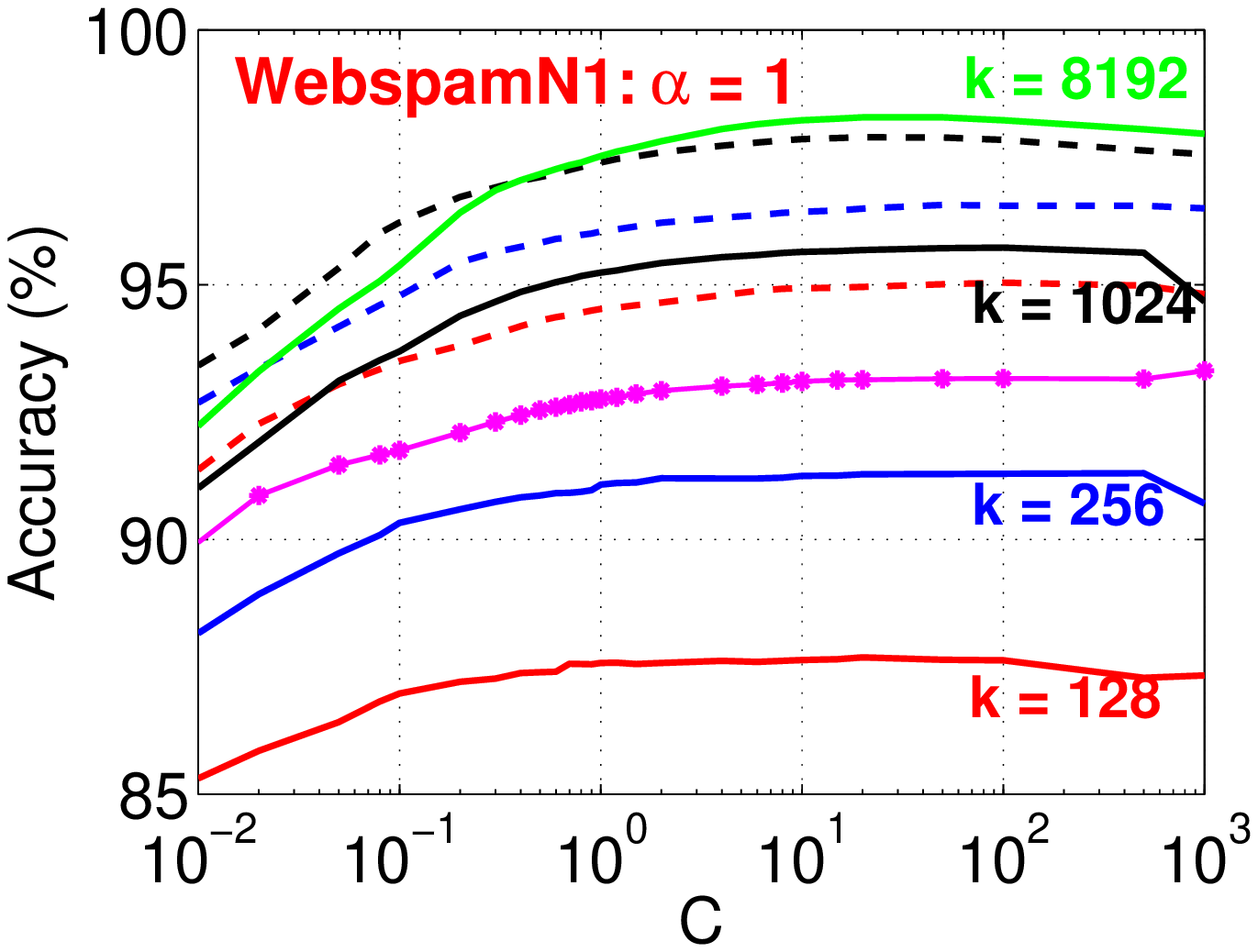}\hspace{-0.15in}
\includegraphics[width=2.3in]{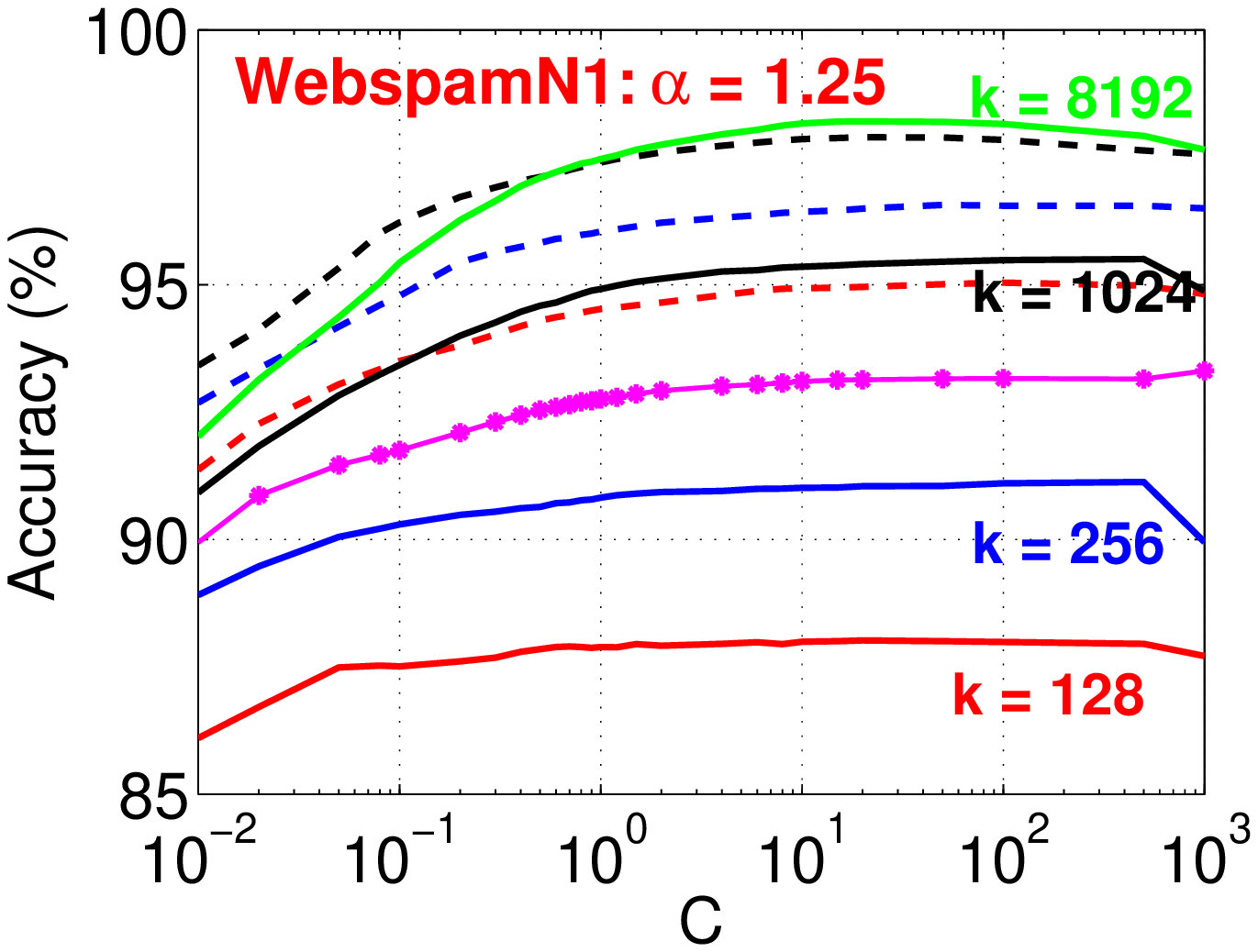}
}

\mbox{
\includegraphics[width=2.3in]{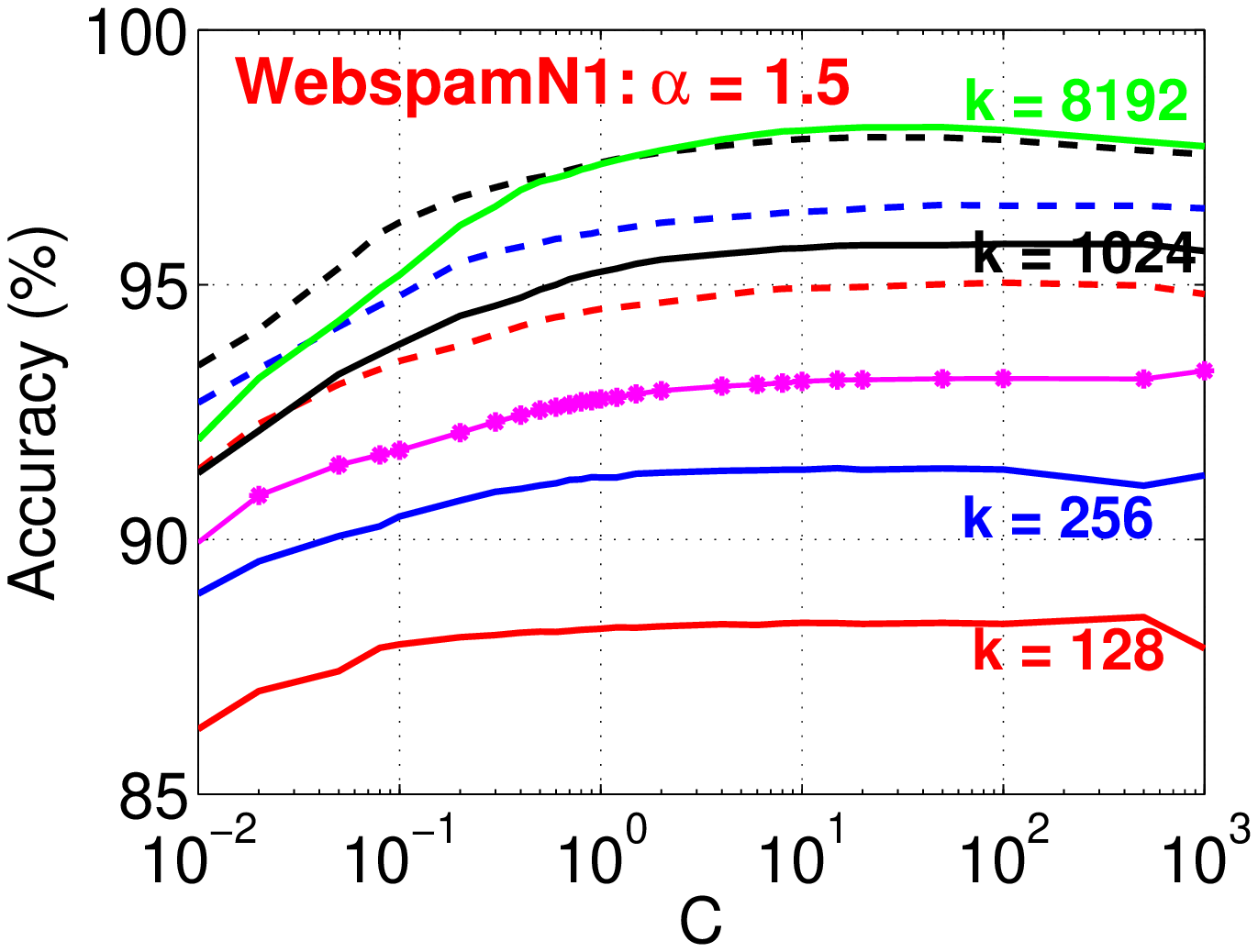}\hspace{-0.15in}
\includegraphics[width=2.3in]{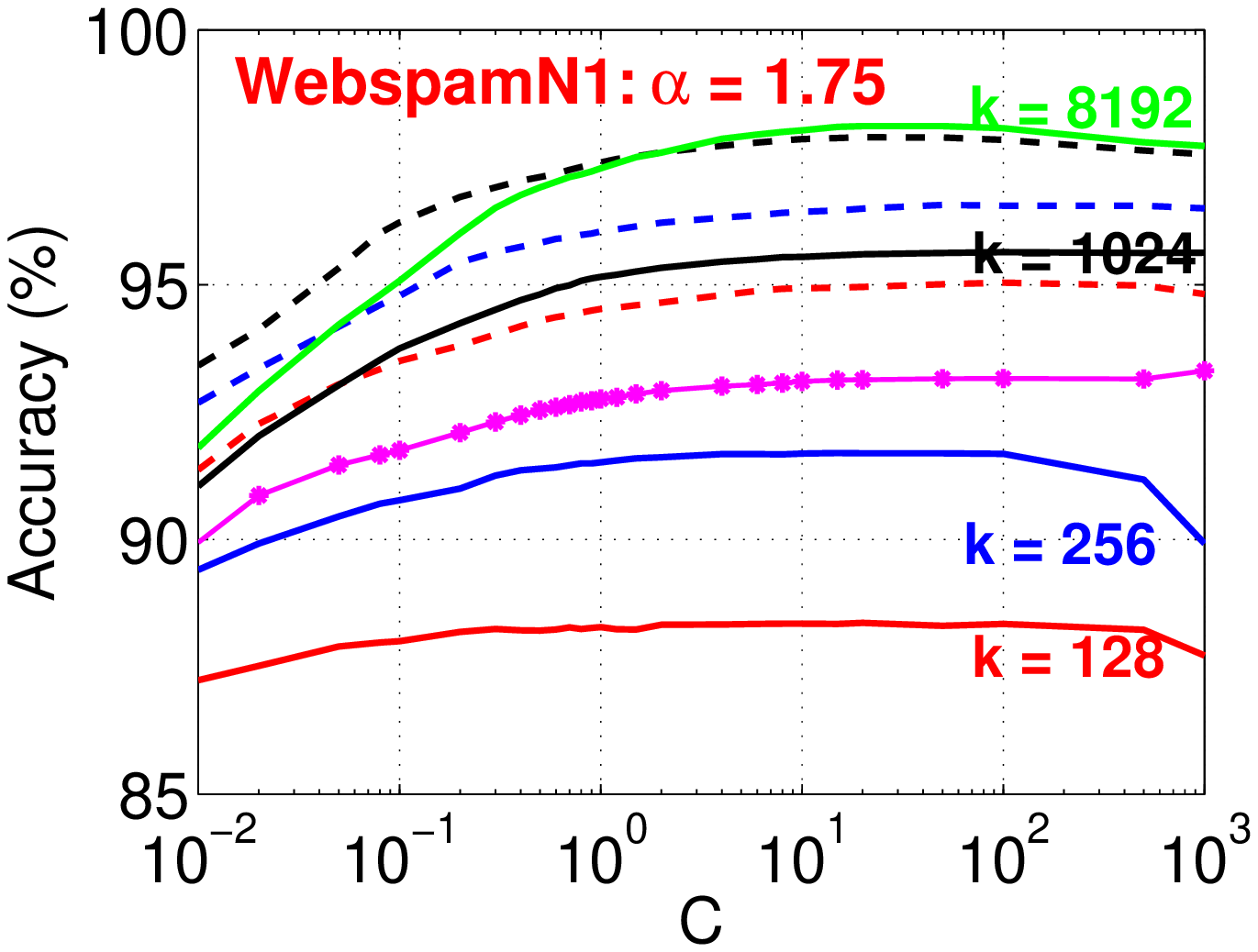}\hspace{-0.15in}
\includegraphics[width=2.3in]{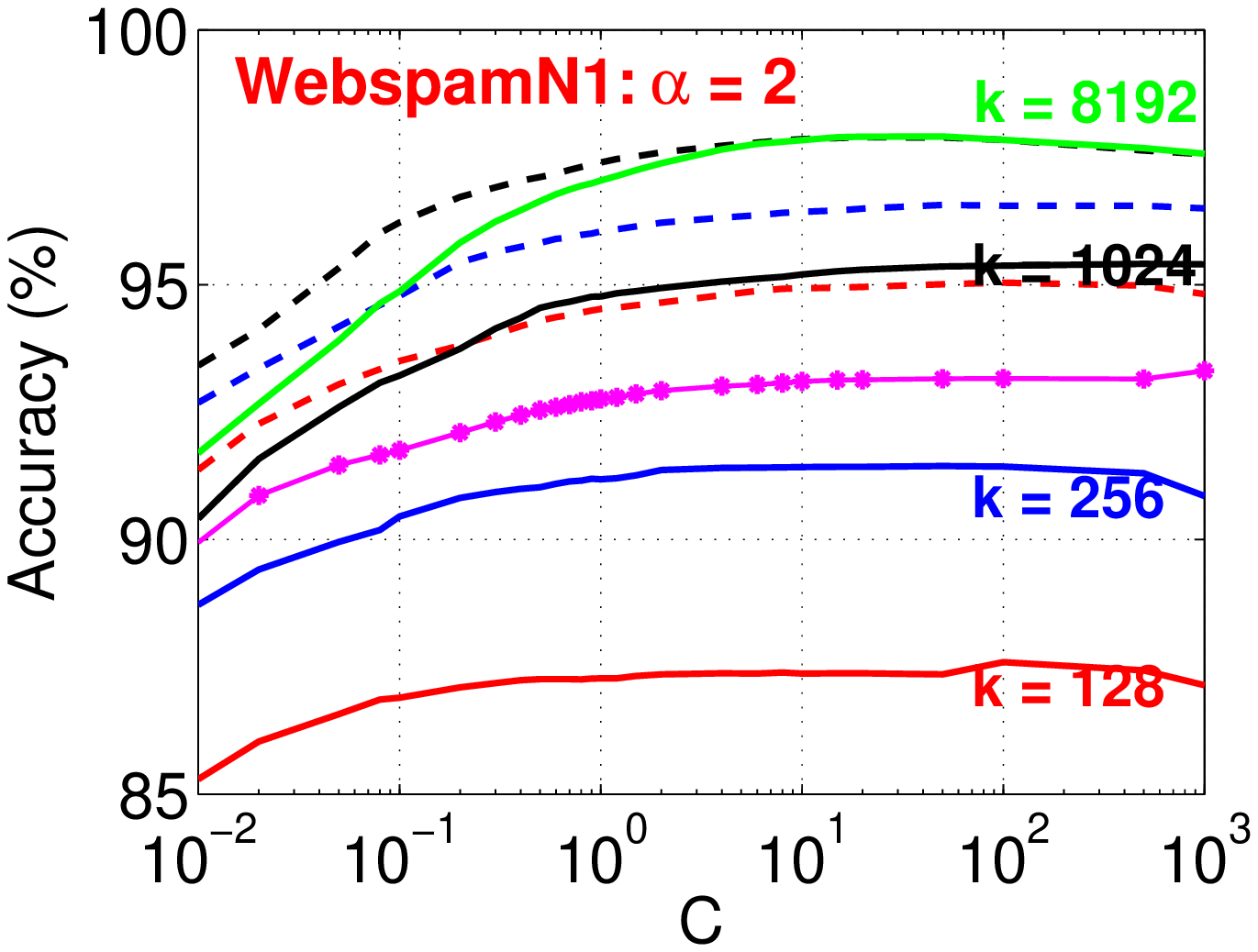}
}

\end{center}
\vspace{-0.3in}
\caption{\textbf{WebspamN1}. We compare sign $\alpha$-stable random projections with 0-bit consistent weighted sampling (CWS). Each panel (for each $\alpha$) consists of 8 curves. The solid (pink) curve marked by * represents the results of linear SVM. Four solid curves (labelled by $k=128$, $k=256$, $k=1024$, and $k=8192$, respectively) represent the results of sign $\alpha$-stable random projections for 4 different $k$ values.  The 3 dashed curves  correspond to the results of 0-bit CWS for $k=128, 256, 1024$ (a higher curve for a higher $k$ value). }\label{fig_WebspamN1}
\end{figure}

Figure~\ref{fig_WebspamN1} presents the detailed results on the {\em WebspamN1}  dataset, which has 350,000 examples. We use 50\% of the examples for training and the other 50\% for testing. With linear SVM, the test classification accuracy is about $93\%$. Both sign $\alpha$-stable random projections and 0-bit CWS can achieve $>98\%$ accuracies given enough samples. The figure also confirm that 0-bit CWS requires significantly fewer samples than the number of projections needed by sign stable random projections, to achieve comparable  accuracies.

\clearpage\newpage

\section{Conclusion}

This paper provides an extensive empirical study of sign $\alpha$-stable random projections for large-scale learning applications. Although the paper focuses on presenting the results on classification tasks, one should keep mind that the method is a general-purpose data processing tool which can be used for classification, regression, clustering, or near-neighbor search.  Given enough projections, the method can often achieve good performance. The comparison with 0-bit CWS should be also interesting to practitioners. \\

\noindent\textbf{Future work}: \ \  The processing cost of sign $\alpha$-stale random projections can be substantially improved by ``very sparse stable random projections''~\cite{Proc:Li_KDD07}. An empirical study is needed to confirm this claim. Another interesting line of research is to combine sign stable random projections with 0-bit CWS, for example, by a strategy similar to that in the recent work of ``CoRE kernels''~\cite{Proc:Li_UAI14}.


\begin{thebibliography}{10}

\bibitem{Book:Bottou_07}
L.~Bottou, O.~Chapelle, D.~DeCoste, and J.~Weston, editors.
\newblock {\em Large-Scale Kernel Machines}.
\newblock The MIT Press, Cambridge, MA, 2007.

\bibitem{Proc:Broder_WWW97}
A.~Z. Broder, S.~C. Glassman, M.~S. Manasse, and G.~Zweig.
\newblock Syntactic clustering of the web.
\newblock In {\em WWW}, pages 1157 -- 1166, Santa Clara, CA, 1997.

\bibitem{Article:Chambers_JASA76}
J.~M. Chambers, C.~L. Mallows, and B.~W. Stuck.
\newblock A method for simulating stable random variables.
\newblock {\em Journal of the American Statistical Association},
  71(354):340--344, 1976.

\bibitem{Proc:Charikar}
M.~S. Charikar.
\newblock Similarity estimation techniques from rounding algorithms.
\newblock In {\em STOC}, pages 380--388, Montreal, Quebec, Canada, 2002.

\bibitem{Article:Fan_JMLR08}
R.-E. Fan, K.-W. Chang, C.-J. Hsieh, X.-R. Wang, and C.-J. Lin.
\newblock Liblinear: A library for large linear classification.
\newblock {\em Journal of Machine Learning Research}, 9:1871--1874, 2008.

\bibitem{Article:Goemans_JACM95}
M.~X. Goemans and D.~P. Williamson.
\newblock Improved approximation algorithms for maximum cut and satisfiability
  problems using semidefinite programming.
\newblock {\em Journal of ACM}, 42(6):1115--1145, 1995.

\bibitem{Article:Indyk_JACM06}
P.~Indyk.
\newblock Stable distributions, pseudorandom generators, embeddings, and data
  stream computation.
\newblock {\em Journal of ACM}, 53(3):307--323, 2006.

\bibitem{Proc:Ioffe_ICDM10}
S.~Ioffe.
\newblock Improved consistent sampling, weighted minhash and \text{L1}
  sketching.
\newblock In {\em ICDM}, pages 246--255, Sydney, AU, 2010.

\bibitem{Proc:Li_KDD07}
P.~Li.
\newblock Very sparse stable random projections for dimension reduction in
  $l_\alpha$ ($0<\alpha\leq 2$) norm.
\newblock In {\em KDD}, San Jose, CA, 2007.

\bibitem{Proc:Li_SODA08}
P.~Li.
\newblock Estimators and tail bounds for dimension reduction in $l_\alpha$
  ($0<\alpha\leq 2$) using stable random projections.
\newblock In {\em SODA}, pages 10 -- 19, San Francisco, CA, 2008.

\bibitem{Proc:Li_UAI14}
P.~Li.
\newblock \text{CoRE} kernels.
\newblock In {\em UAI}, Quebec City, CA, 2014.

\bibitem{Report:Li_CWS15}
P.~Li.
\newblock Min-max kernels.
\newblock Technical report, arXiv:1503.0173, 2015.

\bibitem{Report:Li_1bitCS15}
P.~Li.
\newblock One scan 1-bit compressed sensing.
\newblock Technical report, arXiv:1503.02346, 2015.

\bibitem{Article:Li_Konig_CACM11}
P.~Li and A.~C. K\"onig.
\newblock Theory and applications b-bit minwise hashing.
\newblock {\em Commun. ACM}, 2011.

\bibitem{Proc:Li_ICML14}
P.~Li, M.~Mitzenmacher, and A.~Shrivastava.
\newblock Coding for random projections.
\newblock In {\em ICML}, 2014.

\bibitem{Proc:Li_NIPS13}
P.~Li, G.~Samorodnitsky, and J.~Hopcroft.
\newblock Sign cauchy projections and chi-square kernel.
\newblock In {\em NIPS}, Lake Tahoe, NV, 2013.

\bibitem{Report:Manasse_CWS10}
M.~Manasse, F.~McSherry, and K.~Talwar.
\newblock Consistent weighted sampling.
\newblock Technical Report MSR-TR-2010-73, Microsoft Research, 2010.

\bibitem{Article:Muthukrishnan_05}
S.~Muthukrishnan.
\newblock Data streams: Algorithms and applications.
\newblock {\em Foundations and Trends in Theoretical Computer Science},
  1:117--236, 2 2005.

\bibitem{Book:Samorodnitsky_94}
G.~Samorodnitsky and M.~S. Taqqu.
\newblock {\em Stable Non-Gaussian Random Processes}.
\newblock Chapman \& Hall, New York, 1994.

\bibitem{Book:Zolotarev_86}
V.~M. Zolotarev.
\newblock {\em One-dimensional Stable Distributions}.
\newblock American Mathematical Society, Providence, RI, 1986.

\end{thebibliography}

\end{document}